\documentclass[10pt]{article}
\usepackage[utf8]{inputenc}
\usepackage[T1]{fontenc}
\usepackage{amsmath}
\usepackage{amsfonts}
\usepackage{amssymb}
\usepackage[version=4]{mhchem}
\usepackage{stmaryrd}
\usepackage{bbold}
\usepackage{graphicx}
\usepackage[export]{adjustbox}
\usepackage{hyperref}
\graphicspath{ {./images/} }

\title{ActPC-Geom: \\ Towards Scalable Online \\ Neural-Symbolic Learning \\ via Accelerating Active Predictive Coding \\ with Information Geometry \\ \& Diverse Cognitive Mechanisms  }
\author{Ben Goertzel \footnote{SingularityNET, TrueAGI}}
\date{}

\begin{document}
\maketitle

\begin{abstract}
This paper introduces {\bf ActPC-Geom}, a novel, speculative approach for accelerating Active Predictive Coding (ActPC) based learning in artificial neural networks through the integration of information geometry, specifically leveraging Wasserstein-metric-based methods for measure-dependent gradient flows.

We suggest the possibility of replacing ActPC's use of KL-divergence for predictive error assessment with the Wasserstein metric as well, and provide some formal observations suggesting that may enable more robust integrated behavior on the part of the overall network. 

We propose several strategies aimed at making this approach computationally tractable at the large scale, including 

\begin{itemize} 
\item the use of neural approximators for the inverse measure-dependent Laplacians required for information geometry calculations
\item the application of approximate kernel PCA embeddings to construct low-rank approximations for these neural approximators to output
\item the creation of compositional hypervector embeddings from these kPCA embeddings, to serve as complements to the kPCA vectors
\item ... where the compositional algebra of the hypervectors is configured to work effectively for concepts derived from a fuzzy FCA lattice automatically learned via a neural architecture performing analysis of overall network states
\end{itemize}

The result is an ActPC-based neural architecture that, we suggest, will be capable of real-time online learning, and will support robust integration between continuous ActPC neural networks (including transformer-like architectures) and discrete symbolic ActPC networks (including eg. ActPC-Chem networks leveraging PC methods to guide the evolution of algorithmic chemistry networks; and construction of shared probabilistic, concept-lattice and compositional-hypervector models across symbolic and subsymbolic networks).  

We argue that in a transformer-like architecture of this nature,

\begin{itemize}
\item the compositional algebra of the hypervector embeddings can lead directly to effective compositional reasoning on the part of the overall network, e.g. when answering natural language questions or doing commonsense or scientific reasoning
\item Hopfield-net-type dynamics can be enabled in many of the layers, potentially enabling effective associative long-term memory and other cognitive features as a result of the attractor dynamics.
\end{itemize}

We describe the potential for the ActPC-Geom architecture to combine activation-space-based few-shot learning with incremental online weight updates, potentially enabling sophisticated deliberative thinking dynamics and seamless integration of symbolic and subsymbolic reasoning. 

We also explore how ideas derived from Galois connections can be used to enable efficient concurrent processing of hybrid ActPC/ActPC-Chem networks, and we give a rough sketch for a specialized custom HPC network and corresponding customized hardware design optimized for the demands of ActPC-Geom, with an aim of achieving efficient focused attention for real-time deliberative reasoning.
\end{abstract}

\tableofcontents

\section{Introduction}
Active Predictive Coding (ActPC) comprises a biologically inspired framework for learning and inference. By minimizing prediction errors between internal models and observed data, ActPC iteratively refines both latent representations and model parameters. Unlike backpropagation-based neural networks, ActPC emphasizes local error signals, making it inherently more suited for real-time, online learning and enabling integration of reinforcement learning and symbolic reasoning.

Traditional gradient-based optimization in neural networks struggles to support large-scale real-time learning dynamics due to the brittleness of the underlying backpropagation algorithm, which requires carefully coordinated and synchronized updating across a large network (leading to a reliance on large batch-based updates), and which suffers convergence problems when neural architectures are too recurrent or otherwise too complex. These shortcomings lead to an unfortunate sociotechnical dynamic in which neural architectures oriented toward robust AGI-oriented learning, reasoning and memory are insufficiently pursued because they tend to involve network topologies for which backpropagation will not easily converge.

ActPC resolves these issues on a conceptual and mathematical level, but can also suffer lengthy convergence times and undesirable transient dynamics.   Information geometrically enhanced ActPC (ActPC-Geom) provides a compelling potential alternative: by incorporating measure-dependent operators derived from the Wasserstein distance, one aligns parameter updates with the natural structure of the probability distributions underlying ActPC, thus accelerating learning and smoothing out digressive transient dynamics that might otherwise occur.  

Like most practical applications of information geometry, ActPC-Geom faces significant computational challenges.  However, we believe these can be addressed via appropriate deployment of machine learning and reasoning algorithms within the geometric modeling itself (using neural approximators and information-geometry-guded kernel-PCA-based embeddings).  Further, these ML and MR algorithms can be used to inject valuable cognitive properties into the neural learning process, alongside providing acceleration.

We propose that this novel architecture has potential to achieve sophisticated deliberative thinking dynamics, e.g. blending few-shot learning in activation space with incremental weight updates, and incorporating flexible associate long-term memory and sophisticated compositional reasoning.  We believe this architecture should also be extremely amenable as the subsymbolic half of hybrid neural-symbolic architectures, due to the generality of the principles of predictive coding and information geometry.

\subsection{Addressing Scalability Challenges with Cognitive Solutions}

Because of the fundamental locality of ActPC, it is relatively straightforward to modify it to use Wasserstein-metric based information geometry to accelerate its operations and optimize its intelligence (thus obtaining the overall approach we call here Act-Geom).  

We argue that, alongside this addition, it may be effective to replace the use of KL divergence for the assessment of prediction error in ActPC with an appropriate deployment of the Wasserstein metric.  This allows, as we show, elegant holistic properties of the system-wide dynamics.

While information-geometric machinery can be computationally costly,  we propose integrative-AI solutions to this, starting out with using neural approximators for the inverse of the measure dependent Laplacian  (the key quantity needed for working with the Wasserstein metric)

We argue that, to make this sort of neural approximation scalable, one probably needs to have the neural approximator output an embedding vector embodying a compressed version of the inverse Laplacian matrix.     For example, one might use an approximate kernel PCA based embedding (where the kernel comes from the Wasserstein metric itself) to construct LoRa approximations to these matrices for these neural approximators to output.   

Taking this approximation one step further leads in interesting cognitive directions.   One can expand kPCA based embeddings into hypervector embeddings with compositionality properties, so that the natural algebraic operations on the hypervectors reflect natural operations regarding combination of concepts with different properties.   One can find appropriate concepts to use for constructing these embeddings via using an appropriate neural architecture to learn fuzzy-FCA style concept lattices automatically from analysis of ActPC system states.   In this way the use of approximations to the inverse measure-dependent Laplacian serves a double role as a cognitive strategy: i.e., one is achieving compression via abstraction with a symbolic flavor.

A further critical point is that, due to its localized nature, ActPC (whether in basic or ActPC-Geom form) has strong potential to support online learning, which is critical for AGI-oriented applications.

\subsection{ActPC-Geom Transformers}

The application of these ActPC-Geom mechanisms to implement transformer-like architectures has a number of interesting potential advantages, including

\begin{itemize}
\item having pretraining and instruction tuning in the same framework
\item combining few-shot activation-space-based online learning with ActPC-based link-weight-modification online learning, potentially leading to sophisticated deliberative thinking dynamics
\item incorporating Hopfield-net-like lateral connections in each layer to enable associative long-term memory retrieval
\item leveraging hypervectors obeying compositional symbolic algebra (learned as models of fuzzy concept lattices derived from system states), for guiding both weight-modification and in-context activation-space-based query processing
\end{itemize}

\subsection{Neural-Symbolic Synergies}

There is a fairly clear route to coupling continuous-variable ActPC-Geom networks as described here with discrete ActPC-Chem networks as described in \cite{goertzel2024actpc}, in a way that incorporates both 1) activation spreading between the discrete and continuous networks, and 2) probabilistic modeling across both networks, to provide more accurate geometric information for use in information-geometric acceleration

For instance, the neural approximator used to estimate the measure-based Laplacian and its inverse may leverage the state and history of both the discrete and continuous networks as input data in order to make estimations regarding both networks.   This can be achieved for instance by doing common kPCA and compositional hypervector embedding across the continuous and discrete networks, and guiding these with a fuzzy-FCA-style concept lattice learned by joint neural analysis of both networks.

Potentially, this comprises a quite fine-grained version of the general ''cognitive synergy'' principle \cite{goertzel2023hyperon}, wherein the coupled discrete and continuous networks are providing one another with input to help each other better model their information-geometric contours, and in this way helping guide and accelerate each others' operations

\subsection{Potential for Aggressive Optimization}

Efficient parallel and distributed implementation should be feasible, leveraging the local nature of ActPC learning which should enable multiple processors and/or cores to concurrently process parts of the same neural network in a relatively independent way, without need for extremely precise coordination or synchronization.   We elaborate how the mathematics of Galois connections can be used to explicate certain routes to this sort of optimization, particularly regarding efficient concurrency across multiple cores.

We argue this sort of optimization is highly critical, not only to compete effectively with backpropagation-based neural architectures, but also because highly rapid  online weight updating will be needed to support real-time feedback between activation-space learning and weight-updating learning.   As a crude estimate, one might aim for say a  10-1 ratio of execution speed between weight updating and activation spreading in the focus of attention of a network, to obtain the needed emergent cognitive dynamics.

At the end of the paper, we give a rough outline of a possible specialized custom HPC architecture intended to support this sort of optimization, which serves also as a high-level review of the various components involved in the overall system proposed and how they relate to each other.

\part{Enhancing and Accelerating ActPC with Information Geometry}

\section{Background}

\subsection{ Active Predictive Coding (ActPC)}

Active Predictive Coding (ActPC) \cite{ororbia2022neural}  is a novel reinforcement learning (RL) paradigm grounded in predictive coding principles and, as one compelling option, implemented through Neural Generative Coding (NGC) circuits.   It is inspired by the general notion of Predictive Coding, a biologically inspired theoretical framework founded on the idea that the brain continuously predicts sensory inputs and updates internal representations by reducing prediction errors.  The overall Predictive Coding framework owes a great deal to the conceptual neuroscience models of Karl Friston \cite{friston2009free}, however the technical and mathematical underpinnings of contemporary ActPC work are significantly different from any of Friston's specific published proposals.  Some might perceive the directions pursued in this paper as divergent from some of the concrete ideas in Friston's works, however, we believe they are substantially in the same spirit.

\subsubsection{ActPC: Key Conceptual Aspects}

ActPC leverages the predictive-coding concept to guide action selection and learning without relying on backpropagation-based gradient computations. Instead, it uses local, Hebbian-like learning rules to update synaptic weights, making the approach both biologically plausible and computationally robust (yet far more efficient than simplistic Hebbian learning heuristics).

Key technical aspects of the ActPC approach include:

\begin{enumerate}
  \item Sparse Rewards: Many RL tasks, for example in robotics, provide rewards infrequently. Conventional backpropagation-based RL methods struggle here due to unstable gradient signals. ActPC mitigates this by incorporating both exploratory (epistemic) and goal-oriented (instrumental) signals, enabling effective learning in environments where explicit rewards are rare.
  \item Gradient-Free Optimization: By entirely sidestepping backpropagation, ActPC is not affected by issues like vanishing gradients and differentiability constraints. This is especially useful when employing complex, biologically realistic neuron models or when scaling to large networks.
\end{enumerate}

\paragraph{Neural Generative Coding (NGC) Framework:} Ororbia's NGC framework is a specific instantiation of the ActPC idea, with the following key aspects:

\begin{itemize}
  \item Core Idea: NGC layers predict the activity of subsequent layers. The difference between actual and predicted activity generates an error signal, which drives synaptic updates.
  \item State Updates: Neuron states are iteratively updated based on error signals and local connectivity. A layer's activity, $z^{\ell}$, is refined through a dynamic inference process where the error neurons $e^{\ell}=z^{\ell}-\hat{z}^{\ell}$ guide adjustments.
  \item Predictions: Predictions $\hat{z}^{\ell}$ are generated from forward synaptic connections and memory terms $m_{t}$, reflecting temporal context.
  \item Hebbian-Like Updates: Synaptic weights $W$ and error synapses $E$ are adjusted locally using Hebbian-like rules that strengthen connections contributing to accurate predictions.
\end{itemize}

ActPC, as manifested in NGC, integrates multiple predictive circuits to jointly solve the RL problem. Two key signals drive behavior:

\begin{itemize}
  \item Epistemic Signal (Exploration): Encourages the agent to maximize prediction errors, effectively probing uncertain states and learning from them. This forms an intrinsic reward proportional to the squared prediction errors.
  \item Instrumental Signal (Goal-Directed Behavior): Encourages the agent to minimize prediction errors related to goal achievement. This aligns with traditional RL's objective of maximizing cumulative reward.
  \end{itemize}
  
\noindent A combined reward function fuses these signals, balancing exploration and exploitation to achieve stable and robust policy learning.  

This reward function drives action and reinforcement dynamics in a manner roughly similar to classical RL based systems:

\begin{itemize}
  \item Action Model: Motor outputs $a_{t}$ are produced by an NGC-inspired motor-action circuit, with actions bounded and generated via tanh-based nonlinearities.
  \item Policy Estimation: A policy circuit estimates expected returns $q_{t}$. Target values incorporate immediate rewards and future value estimates, analogous to standard RL value functions.
\end{itemize}

Intended advantages of this framework over backpropagation-based RL include:

\begin{itemize}
  \item Biological Plausibility: Local error-driven synaptic updates align with how the brain might learn, eliminating the need for global gradient signals.
  \item Sparse Reward Robustness: The combined reward structure (epistemic + instrumental) ensures that learning progresses even when external rewards are rare.
  \item Gradient-Free: Without backpropagation, ActPC avoids vanishing gradients and is more compatible with diverse neuron and activation models.
  \item Dynamic Adaptation: The iterative inference and updating process allows continuous adaptation in nonstationary environments.
\end{itemize}

Experiments reported in publications from Ororbia's lab validate these advantages to a significant extent, but they have not yet been demonstrated in large-scale practical or commercial applications.  However it seems a promising direction to explore that, via blending exploration-driven epistemic signals with goal-oriented instrumental signals in ActPC, one can achieve stable, adaptive, and efficient learning at a level of capability beyond conventional RL based approaches. 

\subsubsection{ActPC: Basic Equations}

We'll now give a slightly more in-depth summary of Active Predictive Coding (ActPC) in a formal-neural-network context. The goal is to illuminate (1) how states and weights update iteratively, (2) how local error signals drive these updates, and (3) how ActPC can incorporate reward signals and be amenable to advanced (e.g. info-geometry) techniques. Note that various notations and variants exist in literature; the presentation here is a generic formulation combining aspects from multiple publications.

Consider a multi-layer (or multi-module) neural system with layers $l=1, \ldots, L$. For each layer $l$ :

\begin{itemize}
  \item Let $\mathbf{z}^{l}$ be the latent activations (or states) of that layer.
  \item Let $\hat{\mathbf{z}}^{l}$ denote the predicted version of $\mathbf{z}^{l}$, typically computed from layer $l+1$ or from some generative model for layer $l$.
  \item Let $\mathbf{W}^{l}$ be the weights (or parameters) that help form these predictions.
\end{itemize}

Predictive coding posits that each layer tries to minimize a local mismatch (the ''prediction error'') between $\mathbf{z}^{l}$ and $\hat{\mathbf{z}}^{l}$.   It generally involves doing so iteratively rather than a single forward/back pass.

\paragraph{Overall Objective}
We define a local or global prediction error measure. A simple version might be:

$$
\mathcal{L}_{\text {pred }}=\sum_{l=1}^{L}\left\|\mathbf{z}^{l}-\hat{\mathbf{z}}^{l}\right\|^{2}
$$

\noindent though advanced setups may use more sophisticated geometric measures, or partial sums of errors, etc.  If a reward (or utility) is also considered, we can define:

$$
\mathcal{L}_{\text {total }}=\mathcal{L}_{\text {pred }}-\alpha \mathcal{R}
$$

\noindent where $\mathcal{R}$ is a reward measure (or negative cost), and $\alpha$ is a weighting. Alternatively, as we'll suggest in Section \ref{sec:wasser_outer} below, one might incorporate a Wasserstein distance in $\mathcal{L}_{\text {pred }}$.

\paragraph{Iterative Updates in ActPC}. ActPC is often described as involving two primary updates: one for states $\mathbf{z}^{l}$, and one for weights $\mathbf{W}^{l}$

For latent state updates, for each layer $l$, we define an iterative rule:

$$
\mathbf{z}^{l} \leftarrow \mathbf{z}^{l}-\eta_{z} \nabla_{\mathbf{z}^{l}} \mathcal{L}_{\text {total }},
$$

\noindent where $\nabla_{\mathbf{z}^{l}} \mathcal{L}_{\text {total }}$ is the partial derivative (local mismatch plus any reward-driven gradient) w.r.t. $\mathbf{z}^{l}$.

In a simple ''prediction error only'' case, for instance:

$$
\nabla_{\mathbf{z}^{l}} \mathcal{L}_{\text {pred }} \approx\left(\mathbf{z}^{l}-\hat{\mathbf{z}}^{l}\right)-(\text { some feedback from layer } l-1 \text { or } l+1) .
$$

Generally speaking, each $\mathbf{z}^{l}$ is iteratively refined in short micro-steps, seeking to reduce local mismatch.

\paragraph{Weight Updates}  In the basic learning loop, each $\mathbf{W}^{l}$ is updated to reduce mismatch (plus incorporate reward if relevant). A straightforward expression might be:

$$
\mathbf{W}^{l} \leftarrow \mathbf{W}^{l}-\eta_{W} \nabla_{\mathbf{W}^{l}} \mathcal{L}_{\text {total }} .
$$

\noindent If we separate the notion of ''predictive error'' from ''reward,'' we might do:

$$
\nabla_{\mathbf{W}^{l}} \mathcal{L}_{\text {total }}=\nabla_{\mathbf{W}^{l}} \mathcal{L}_{\text {pred }}-\alpha \nabla_{\mathbf{W}^{l}} \mathcal{R} .
$$

\noindent In short, the system ''learns the generative model'' (the mapping from layer $l+1$ to layer $l$ ) or from environment to layer 1 , while also shaping weights to capture reward signals if present.

\paragraph{ActPC Flow in a Multi-Layer Setting: The Forward Process}   Let's step through how all these parts fit together in the context of a multi-layer ActPC neural network..

Consider the forward process: For each layer $l$, we define a generative function $\hat{\mathbf{z}}^{l}=$ $g^{l}\left(\mathbf{z}^{l+1}, \mathbf{W}^{l}\right)$. Some versions also incorporate $\mathbf{z}^{l-1}$. The local error is:

$$
\mathbf{e}^{l}=\mathbf{z}^{l}-\hat{\mathbf{z}}^{l}=\mathbf{z}^{l}-g^{l}\left(\mathbf{z}^{l+1}, \mathbf{W}^{l}\right)
$$

To effect the state update for $\mathbf{z}^{l}$, we do a small step:

$$
\mathbf{z}^{l} \leftarrow \mathbf{z}^{l}-\eta_{z}\left(\frac{\partial\left\|\mathbf{e}^{l}\right\|^{2}}{\partial \mathbf{z}^{l}}\right)=\mathbf{z}^{l}-\eta_{z}\left(2 \mathbf{e}^{l}\right)
$$

\noindent assuming a simple squared error. (We can also incorporate feedback from $\mathbf{e}^{l \pm 1}$.)

To effect the weight update for $\mathbf{W}^{l}$, focusing on $\mathbf{e}^{l}$ gives us:

$$
\mathbf{W}^{l} \leftarrow \mathbf{W}^{l}-\eta_{W}\left(\frac{\partial\left\|\mathbf{e}^{l}\right\|^{2}}{\partial \mathbf{W}^{l}}\right),
$$

\noindent where

$$
\frac{\partial\left\|\mathbf{e}^{l}\right\|^{2}}{\partial \mathbf{W}^{l}}=2 \mathbf{e}^{l} \cdot \frac{\partial \mathbf{z}^{l}}{\partial \mathbf{W}^{l}}-2 \mathbf{e}^{l} \cdot \frac{\partial g^{l}\left(\mathbf{z}^{l+1}, \mathbf{W}^{l}\right)}{\partial \mathbf{W}^{l}} .
$$

\paragraph{ActPC Flow in a Multi-Layer Setting: Micro-Iterations} 

Instead of a single forward+back pass, ActPC typically does multiple ''micro-iterations'' per data sample or environment state. Each iteration:

\begin{enumerate}
  \item Compute local mismatch $\mathbf{e}^{l}$.
  \item Update $\mathbf{z}^{l}$ and $\mathbf{W}^{l}$ slightly.
  \item Possibly incorporate reward signals if this is an RL-like setting.
  \item Proceed until partial convergence or a time limit.
\end{enumerate}

\subsubsection{Discrete Rewriting + Continuous States}

In a hybrid discrete/continous scenario (like the ActPC-Chem framework outlined in \cite{goertzel2024actpc}):

\begin{itemize}
  \item Discrete expansions are also proposed in each iteration.   E.g. in ActPC-Chem, there is a pool of rewrite rules (rewriting input data, output primitives and each other), and each rewrite rule tries to reduce mismatch or yield a better predicted outcome.
  \item The system merges expansions from discrete rewriting with continuous updates in an aggregator process
  \item The same local error principle applies: each rewriting candidate that does not reduce error is pruned.
\end{itemize}

\subsubsection{Summary of the Architectural/Algorithmic Flow}

To sum up how ActPC roughly works in a neural-network or hybrid neural-symbolic setting:

\begin{enumerate}
  \item Initialize: We have an environment or input. Each layer $l$ has states $\mathbf{z}^{l}$ and weights $\mathbf{W}^{l}$.
  \item Predict: Compute $\hat{\mathbf{z}}^{l}$ from the generative functions.
  \item Error: $\mathbf{e}^{l}=\mathbf{z}^{l}-\hat{\mathbf{z}}^{l}$.
  \item State/Weight Updates: Perform micro-steps:

$$
\mathbf{z}^{l} \leftarrow \mathbf{z}^{l}-\eta_{z} \nabla_{\mathbf{z}^{l}} \mathcal{L}_{\text {pred }}, \quad \mathbf{W}^{l} \leftarrow \mathbf{W}^{l}-\eta_{W} \nabla_{\mathbf{W}^{l}} \mathcal{L}_{\text {pred }}
$$

\noindent (here is where reward $\mathcal{R}$ would be incorporated, and also where ActPC-Geom inserts advanced measure-based geometry and its mathematical and cognitive approximations.)
\item In a neural-symbolic system, here discrete (e.g. ActPC-Chem) expansions also produce candidate states. If they yield lower mismatch, they replace the old discrete config or are accepted by the aggregator.
\item Repeat until partial convergence or next environment step.
\end{enumerate}

Thus, we see how ActPC's local mismatch loops are equation-based but iterative (rather than feed-forward + single backprop pass). This style is well-suited to real-time updates, concurrency, and info-geometry enhancements.

\subsection{Information Geometry and Wasserstein Distance}

Information geometry, which in the following section we apply to enhance the ActPC approach, refers to the general study of the structure of probability distributions as a curved manifold.   It has a long history \cite{amari2016information}, a number of practical applications, and also multiple technical variations.

One approach to information geometry is the class of Wasserstein-metric-based approaches, grounded in optimal transport theory.  The Wasserstein distance measures the cost of transporting probability mass between two distributions $p$ and $q$ on the space $\mathcal{X}$ :

$$
W_{2}(p, q)=\inf _{\pi \in \Gamma(p, q)}\left(\int_{\mathcal{X} \times \mathcal{Y}}\|x-y\|^{2} d \pi(x, y)\right)^{1 / 2} .
$$

\noindent where $\Gamma(p, q)$ is the set of all joint distributions with marginals $p$ and $q$. In the continuous setting (on $\mathcal{X} \subseteq \mathbb{R}^{n}$ ), one often views $W_{2}$ through the lens of optimal transport maps and the geometry they induce on the space of probability measures.

\subsubsection{Riemannian viewpoint and ''Otto calculus''}

When one regards the space of probability densities $\mathcal{P}(\mathcal{X})$ (with suitable regularity conditions) as an infinite-dimensional manifold, the Wasserstein metric endows it with a formal Riemannian structure. In broad strokes:

\begin{itemize}
  \item A ''tangent vector'' at a density $p$ can be identified with a scalar function $\phi$ such that the infinitesimal change from $p$ is $\delta p=-\nabla \cdot(p \nabla \phi)$.
  \item The corresponding ''Riemannian metric'' on this tangent space is typically written as

$$
\langle\phi, \psi\rangle_{p}=\int_{\mathcal{X}} p(x) \nabla \phi(x) \cdot \nabla \psi(x) d x
$$
\end{itemize}

\noindent From this perspective, geodesic equations in this Wasserstein geometry coincide with the PDE for optimal transport (the so-called ''continuity equation'' with a velocity field that is a gradient).

\subsubsection{Measure-dependent Laplacian}

\paragraph{Continuous viewpoint}  In the continuous, PDE-based formulation, one often sees operators like the ''weighted'' or ''measure-dependent'' Laplacian:

$$
\Delta_{p} \phi=\nabla \cdot(p \nabla \phi)
$$

\noindent which arises when linearizing the PDE flows in this Wasserstein geometry. Concretely, $\Delta_{p}$ depends on $p$ because the velocity field that deforms $p$ is ''weighted'' by $p$ itself (reflecting mass transport). This is one way the Wasserstein metric ''knows'' about the underlying measure it is transporting.

\paragraph{Discrete viewpoint} In a discrete setting -- say one approximates $\mathcal{X}$ by a graph with nodes $i, j$ -- the measure-dependent Laplacian can be written in a form such as

$$
L(p)_{i j}=\omega_{i j}\left(p_{i}+p_{j}\right), \quad i \neq j
$$

\noindent where $\omega_{i j}$ encodes a notion of adjacency or cost for transport between nodes $i$ and $j$. The diagonal terms of $L(p)$ are typically chosen so that row sums vanish (mimicking the continuous Laplacian property $\sum_{j} L_{i j}=0$ ). The factor $\left(p_{i}+p_{j}\right)$ or variants thus ensures that the graph Laplacian ''adapts'' to the mass distribution $p$.

\paragraph{Why the inverse of $L(p)$ is critical}  When one discretizes PDE-based formulations or tries to solve optimization problems (e.g., finding geodesics, implementing gradient flows, or computing barycenters in Wasserstein space), one encounters linear systems involving $L(p)$ or its variations. In particular, if one wants to linearize or invert transformations in the tangent space (for example, to perform Newton-type methods for geodesic shooting), one needs $(L(p))^{\dagger}$ (where $\dagger$ denotes pseudoinversion) or else efficient method to solve linear systems with $L(p)$.

Because these matrices can be very large (and depend on $p$ ), developing fast linear-algebra techniques is central to making Wasserstein-based methods practical in large-scale applications. This basically demands specialized algorithms that exploit the structure of $L(p)^{\dagger}$.

\paragraph{The inverse measure-dependent Laplacian in ActPC-Geom} The core idea of ActPC-Geom (to be elaborated just below) is that, by embedding Wasserstein geometry into ActPC, neural link weight updates become distribution-aware, aligning local corrections with the natural geometry of the underlying probability space and thus accelerating learning.   Leveraging this in practice requires geodesic shooting and related methods (to allow the ActPC dynamics to follow the geodesics in the Wasserstein space), which leads to a need to efficiently estimate the inverse of the (potentially very large) measure-dependent Laplacian matrix, which leads to some mathematical intricacies and -- in the route that we suggest here -- some cognitive subtleties as well.

\section{ActPC-Geom: Upgrading ActPC with Wasserstein Geometry} \label{sec:wasser}

A central challenge in practical implementations of continuous-variable Active Predictive Coding (ActPC) is achieving stable, efficient convergence for larger or more complex networks.  While simply scaling model size or data can help, a complementary approach is to introduce adaptive optimization into the underlying dynamics -- e.g. by using Wasserstein-based natural gradient steps to guide the continuous update dynamics. The idea is to align gradient directions with the intrinsic geometry of the probability distributions involved, potentially improving convergence speed and stability. 

The emphasis in ActPC on local iterative updates naturally aligns it with the incremental, geometry-aware steps afforded by Wasserstein-based optimization. Additional conceptual synergies between ActPC and information geometry include:

\begin{itemize}
  \item Activation Space Matching: Predictive coding focuses on reducing local distribution mismatches, a task that directly benefits from the transport-based perspective of Wasserstein geometry.
  \item Real-Time Feasibility: The smaller, iterative updates in ActPC fit well with the computational demands of measure-dependent operators, as opposed to the large batch-based gradients of backpropagation.
\end{itemize}

We now sketch out an outline of how one might integrate Wasserstein-geometry techniques into (continuous) ActPC.

\subsection{Setup}

We will assume a learning scenario where an agent tries to minimize an error measure or maximize a reward through local, gradient-free updates, as in standard ActPC.  However, we will leverage ideas from optimal transport (Wasserstein geometry) to shape the gradient steps in the underlying probability space of parameters or distributions.

Let $\boldsymbol{\xi} \in \mathbb{R}^{m}$ represent the agent's current parameters-- assuming a neural net based agent this might include neural states, weights, or a mixture of both.   (For an ActPC-Chem based system, it would comprise e.g. the rewrite rules in the system and any numerical parameters associated with them.)   These parameters induce a probability distribution

$$
p(\cdot \mid \boldsymbol{\xi})
$$

\noindent which captures how the agent's model (or policy) generates predictions about the environment or next states.  So, the distribution goes over future states, conditioned on present state, and captures the predictions about future states implicit in present state.

\paragraph{Error Functional} In RL-style tasks, one would typically define an error functional such as:

$$
F(\boldsymbol{\xi})=-r(\boldsymbol{\xi}) \quad \text { or } \quad F(\boldsymbol{\xi})=\mathcal{L}(\boldsymbol{\xi})
$$

\noindent where $r(\boldsymbol{\xi})$ is a reward function (the agent tries to maximize reward by minimizing $-r$ ), and $\mathcal{L}(\boldsymbol{\xi})$ might be an information-theoretic or predictive-coding loss (e.g., KL divergence or Wasserstein distance between predicted and observed distributions).  

 In Section \ref{sec:wasser_outer} we will make an argument for having ActPC-Geom systems measure the overall prediction error of their outputs using Wasserstein distance,
 
 $$
\mathcal{L}(\boldsymbol{\xi})=W_{2}(q(\mathbf{x}), p(\mathbf{x} \mid \boldsymbol{\xi}))
$$

\noindent for greater consilience with the internal use of Wasserstein distance to accelerate learning. The considerations in this section, though, are applicable whether one uses Wasserstein or KL in the "outer loop" for measuring overall systemic prediction error.  One could also, say, average together a prediction-error measure with a traditional external reward signal.

\subsection{Wasserstein Geometry}

Let $p(\mathbf{x} \mid \boldsymbol{\xi})$ be a distribution over states (or outputs) $\mathbf{x} \in \mathcal{X}$.  

As reviewed in \cite{li2018natural}, a measure-dependent Laplacian 

$$
L(p)_{i j}=\omega_{i j}\left(p_{i}+p_{j}\right), \quad i \neq j
$$

\noindent i.e.

$$
L(p)(\mathbf{x},\mathbf{y})=\omega(\mathbf{x}, \mathbf{y}) \left(p( \mathbf{x})+p(\mathbf{y})\right)
$$

\noindent can be defined on a graph (or continuous space) capturing the ground metric $\omega(\mathbf{x}, \mathbf{y})$ for transporting mass from $\mathbf{x}$ to $\mathbf{y}$. 

\subsubsection{The Ground Metric}

Where does the ground metric come from?   A simple syntactic distance between two states would be one option -- e.g. an edit distance.  Conceptually, one could even use a logical inference system like PLN to estimate the similarity between states.  In practice, though, one needs the ground metric to be rapidly evaluable.   A sensible idea would be to train a separate neural predictor to estimate the semantic (behavioral) distance between two neural states.  

To make this estimator passably efficient, one could create an appropriate mapping of the neural states into embedding vectors, and have this predictor act on these vectors.   As a simple example, the  embedding could use an efficient approximation to kernel PCA with a kernel based on a crude syntactic distance, along the lines discussed in Section \ref{sec:kpca} below.   Or one could use hypervectors manifesting a compositional algebra, harking forward to Section \ref{sec:hyper}. 

To train such an estimator, one would want an ensemble of cases where one knew both internal states and external behaviors of an ActPC network.  The goal of the estimator would be to predict the distance between the behaviors associated with two states, based on looking at the embedding vectors corresponding to the two states.

\subsubsection{The Metric Tensor}

Given a ground metric, one then obtains a metric tensor:

$$
\noindent G(\boldsymbol{\xi})=J_{\boldsymbol{\xi}}^{\top} L(p(\boldsymbol{\xi}))^{\dagger} J_{\boldsymbol{\xi}}
$$

\noindent where $J_{\boldsymbol{\xi}}$ is the Jacobian mapping from parameters $\boldsymbol{\xi}$ to the distribution $p(\cdot \mid \boldsymbol{\xi})$, and $\dagger$ denotes (pseudo)inversion.

\subsection{ Wasserstein Natural Gradient Update}

The agent seeks to minimize $F(\boldsymbol{\xi})$. In a Wasserstein geometry, the update in continuous time is

$$
\frac{d \boldsymbol{\xi}}{d t}=-G(\boldsymbol{\xi})^{-1} \nabla_{\xi} F(p(\boldsymbol{\xi}))
$$

\noindent where $\nabla_{\xi} F$ is the standard gradient in parameter space (assuming we can differentiate $F$ with respect to $\boldsymbol{\xi})$.

In practice, we use small discrete steps:

$$
\boldsymbol{\xi}_{k+1}=\boldsymbol{\xi}_{k}-\eta G\left(\boldsymbol{\xi}_{k}\right)^{-1} \nabla_{\xi} F\left(p\left(\boldsymbol{\xi}_{k}\right)\right),
$$

\noindent where $\eta$ is a step size or learning rate. This update rule can be seen as a generalization of standard gradient descent that respects the underlying geometry of the probability space.

\subsection{Predictive Coding Error}

Now where does the PC come into it?

If $\mathbf{z}$ are neural states (representations), we might measure mismatch between predicted $\hat{\mathbf{z}}$ and actual $\mathbf{z}$ with a cost such as

$$
\mathcal{L}_{P C}(\boldsymbol{\xi})=\sum_{\ell}\left\|\mathbf{z}^{\ell}-\hat{\mathbf{z}}^{\ell}(\boldsymbol{\xi})\right\|^{2} \quad \text { (or a KL divergence or Wasserstein form). }
$$

If a reward function $r(\boldsymbol{\xi})$ is given, one can define

$$
F(\boldsymbol{\xi})=-r(\boldsymbol{\xi})+\alpha \mathcal{L}_{P C}(\boldsymbol{\xi}),
$$

\noindent blending RL and ActPC error. 

Minimizing $F$ amounts to maximizing reward while minimizing predictive error.\\

The partial derivatives $\nabla_{\xi} F$ come from backprop-like expansions within the local PC or RL objectives, though the overall update is shaped by $G(\boldsymbol{\xi})$.

\subsection{Summary of Core Equations}

Collating the above, we have the process flow:

\begin{enumerate}
\item  Compute local gradient:
$$
\mathbf{g}_{k}=\nabla_{\xi} F\left(p\left(\boldsymbol{\xi}_{k}\right)\right)
$$
\item  Construct measure-dependent Laplacian:
$$
L\left(p\left(\boldsymbol{\xi}_{k}\right)\right) \left.\omega\right)
$$
\item  Form metric tensor: 
$$
G\left(\boldsymbol{\xi}_{k}\right)=J_{\boldsymbol{\xi}_{k}}^{\top} L\left(p\left(\boldsymbol{\xi}_{k}\right)\right)^{\dagger} J_{\boldsymbol{\xi}_{k}}
$$
\item  Update step:
$$
\boldsymbol{\xi}_{k+1}=\boldsymbol{\xi}_{k}-\eta G\left(\boldsymbol{\xi}_{k}\right)^{-1} \mathbf{g}_{k}
$$
\end{enumerate}

\subsection{Crude Pseudocode Sketch}

Corresponding crude pseudocode might look like:

\begin{verbatim}
Initialize parameters xi (e.g., random) // xi in R^m
Initialize step size eta
Initialize environment / data
Initialize ground metric omega, define measure-dependent Laplacian L()
function ACTPC_WASSERSTEIN_UPDATE(xi, environment):
    // 1. Evaluate cost and gradient
    cost_value = F(xi, environment) // e.g. combined RL + PC error
    gradient_g = compute_gradient(F, xi, environment)
     // partial derivatives wrt xi
    // 2. Construct measure-dependent Laplacian
    dist_p = compute_distribution(xi) // e.g. p(. | xi)
    L_p = build_measure_dependent_laplacian(dist_p, omega) 
    // graph-based or continuous space
    // 3. Compute metric tensor G(xi)
    J_xi = compute_jacobian_of_distribution(xi) 
    // partial p(.|xi) wrt xi
    L_p_inv = pseudo_inverse(L_p) 
    // or approximate inversion
    G_xi = J_xi^T * L_p_inv * J_xi
    // 4. Perform Wasserstein natural gradient update
    G_xi_inv = invert(G_xi)
     // or approximate this instead (as well be be discussed below)
    delta_xi = - eta * (G_xi_inv * gradient_g)
    xi_new = xi + delta_xi
    return xi_new
// Main Learning Loop
xi = initialize_xi()
for iteration in 1..MAX_ITERS:
    // Possibly interact with environment, gather reward and errors
    xi = ACTPC_WASSERSTEIN_UPDATE(xi, environment)
    // Evaluate progress or break early if converged
    if check_convergence(xi):
        break
\end{verbatim}

To run through the key steps here:

\paragraph{compute\_gradient( $F, x i$, environment)}  This function obtains the partial derivatives of the cost function $F$ w.r.t. $\boldsymbol{\xi}$. The cost can encode a mixture of RL reward terms and predictive-coding discrepancy terms. Implementation depends on whether $\boldsymbol{\xi}$ is pure weights, neuron states, or both.

\paragraph{ compute\_distribution(xi)} The agent's model or policy might define $p(\mathbf{x} \mid \boldsymbol{\xi})$. This step extracts the current distribution given $\boldsymbol{\xi}$.

\paragraph{build\_measure\_dependent\_laplacian(dist\_p, omega)}  Depending on whether we are in a discrete or continuous domain, this might build a graph-based Laplacian with edge weights $\omega_{i j}$, or a continuous operator approximating the local geometry of the distribution $p(\mathbf{x} \mid \boldsymbol{\xi})$. If we treat states $\mathbf{x}$ as nodes, $\omega\left(\mathbf{x}_{i}, \mathbf{x}_{j}\right)$ is the ground distance for transporting probability mass.

\paragraph{compute\_jacobian\_of\_distribution(xi)}  We need the derivative of $p(\mathbf{x} \mid \boldsymbol{\xi})$ or relevant summary statistics with respect to $\boldsymbol{\xi}$. In practice, this might be approximated or only partially computed.\\

\paragraph{invert(G\_xi)}  We invert  $G(\boldsymbol{\xi})$; or more realistically, approximate this inverse. If dimension is large, one might use low-rank approximations or diagonal blocks to keep it tractable.

\paragraph{ActPC RL} The cost $F$ might combine a negative reward term $-r(\boldsymbol{\xi})$ and a predictive-coding error $\mathcal{L}_{P C}(\boldsymbol{\xi})$. The approach then updates $\boldsymbol{\xi}$ in directions that reduce both cost terms.

\subsection{Conclusion}

The details are slightly intricate -- and will become significantly more so in following sections of this paper -- but the core idea here is simple: By incorporating $G(\boldsymbol{\xi})$ into each step, we move $\boldsymbol{\xi}$ in ways that respect the underlying geometry of the distribution. This can mitigate chaotic or inefficient updates that sometimes plague naive gradient steps, especially when data is limited or the environment's distribution is high-dimensional or complex.

Exact computation of measure-dependent Laplacians or their inverses can be expensive. Hence, approximate or stochastic versions may be necessary in real-world or large-scale problems. However, even partial use of Wasserstein geometry-for instance, using simpler or block-diagonal approximations—can stabilize and accelerate learning compared to purely local Euclidean steps.

Overall, an ''ActPC on Wasserstein'' approach to RL unifies standard active predictive coding (ActPC) with geometry-aware gradient flows, potentially enabling faster, more stable convergence in complex or data-scarce regimes -- an appealing direction for future theoretical and experimental research.

\section{Neural Approximators for Measure-Dependent Operators: Making ActPC-Geom Feasible}

While the use of information geometry to accelerate ActPC is conceptually appealing, the exact computation of $G(\mathbf{W})^{-1}$ or $\hat{L}^{\dagger}$ for large networks is computationally prohibitive. 

We push forward to explore how  to work around this issue and make ''ActPC on information geometry'' efficient and scalable in practice, via introducing appropriate approximations.  This turns out to be a deep dive, beginning with fairly traditional mathematical approximations and leading us toward more cognitive approaches.

Our first step is to introduce a neural approximator approach:  Train a neural network $f_{\theta}(\mathbf{f})$ to output low-rank approximations of  $\hat{L}^{\dagger}$, given features $\mathbf{f}$ extracted from the network's state distribution.

Next arises the question of how to train this neural approximator.   As a matter of temporary scaffolding one could use a traditional backpropagation methodology, or one could use other methods such as evolutionary learning.   What would be most natural here, though, would be to leverage predictive coding based learning here as well.   One could go full-on recursive and use information geometry to help guide the predictive coding methods used to train the approximator network.  At some level of recursion depth, though, one has to set the information geometry aside and use a simpler ML approach as a ''base case.''

The approximator network can be trained (by whatever method) to minimize some reconstruction error, e.g.,

$$
\mathcal{L}^{\dagger}_{\text {approx }}=\| L(\hat{p})^{\dagger}-f_{\theta}(\text { features }) \|^{2},
$$

\noindent where $L(\hat{p})^{\dagger}$ is an inverse Laplacian computed on a smaller reference set or a carefully sampled subset.

This approach can be executed in batch mode, but should also work fine in an online learning setting.   As the main ActPC network evolves, the distribution $p(\mathbf{x} \mid \boldsymbol{\xi})$ change. The ''Laplacian approximator'' network sees new samples or partial computations and refines its parameters $\theta$.

\subsection{Stochastic Low-Rank Decomposition of the Inverse Measure-Dependent Laplacian}

One approach to making this sort of approximator work at scale would be to factor the matrix $(L(p))^{\dagger}$ via low-rank decomposition or sketching methods, such as:

\begin{itemize}
\item Random Projection:
\begin{itemize}
  \item Project the Laplacian $L$ onto a small set of random vectors, building a compressed representation:
  \item Solve the smaller system in this reduced dimension.
  \end{itemize}
  \end{itemize}
  
  \noindent or (probably more promising for large-scale cases)
  
  \begin{itemize}
  \item Nystrom Method or Similar SVD Approximations:
    \begin{itemize}
  \item Sample a subset of states or representative points $\left\{\mathbf{x}_{i}\right\}$.
  \item Construct an approximate factorization of $(L(p(\boldsymbol{\xi})))^{\dagger}$.\\
$\mathcal{O}\left(r^{3}\right)$, for rank $r \ll n$.
\end{itemize}
  \end{itemize}

Given such a dimensionally reduced version, we can train a neural network $f_{\theta}(\cdot)$ that, given the current distribution parameters (or a representative sample set) and the ground metric, outputs an approximation the inverse Laplacian $(L(p(\boldsymbol{\xi})))^{\dagger}$.

To make this work tractably in practice, one might do partial ''ground-truth'' calculations for $L^{\dagger}$ on a small batch of states, then train $f_{\theta}$ to generalize to the broader parameter space.

I.e., at each time step, the system:

\begin{enumerate}
  \item Extracts relevant features describing $p\left(\cdot \mid \boldsymbol{\xi}_{t}\right)$. (e.g., statistics or sampled states.)
  \item Passes them to the neural approximator network $f_{\theta}$, which approximates a compressed representation of $\hat{L}$ or $\hat{L}^{\dagger}$
  \item Uses this approximated compressed $\hat{L}$ or $\hat{L}^{\dagger}$ to update parameters $\boldsymbol{\xi}$ via

$$
\boldsymbol{\xi}_{t+1}=\boldsymbol{\xi}_{t}-\eta \hat{G}\left(\boldsymbol{\xi}_{t}\right)^{-1} \nabla_{\xi} F
$$

\noindent where 

$$
\hat{G}\left(\boldsymbol{\xi}_{t}\right)=J_{\boldsymbol{\xi}_{t}}^{\top} \hat{L}^{\dagger} J_{\boldsymbol{\xi}_{t}}
$$

\end{enumerate}

This process can be tweaked and adjusted in various ways, and may in itself be effective at large scale.  However, it seems possible that to  make this work at the scales needed for AGI, it may be necessary to get even more clever and aggressive about how to build a low-dimensional representation of the inverse Laplacian $\hat{L}^{\dagger}$.  Much of this paper is an elaboration of this theme.   First, in the following subsection, we propose some fairly straightforward linear-algebra/numerical-analysis approaches.   Then we proceed to more AI-ish approaches, involving embedding vectors (Section \ref{sec:kpca}) and hypervectors obeying compositional algebra (Section \ref{sec:hyper})

 \subsection{Compressed Representations for the Inverse Measure-Dependent Laplacian}. 
 
 We now dig into a little more detail regarding how one might choose a compressed representation (or ''embedding vector'') for the ( $U_{k}, \Sigma_{k}, V_{k}$ ) factorization of the inverse measure-dependent Laplacian.  One point we wish to emphasize is why we might prefer embedding vectors to direct matrix factor outputs, and the variety of possible ways to construct such embeddings (e.g., kernel methods, autoencoders, or direct $\ell_{2}$-based projection). We will focus particular attention on approximate kernel PCA approaches, which we consider especially promising.
 
The basic motivation for looking at compressed representations is that,  even with a rank-r factorization, storing $(U, \Sigma, V)$ explicitly can be $\mathcal{O}(n r)$. For large $n$, this may still be huge.    A ''compressed embedding vector'' $\mathbf{z} \in \mathbb{R}^{d}$ (with $d \ll n r$ ) can be much more for the neural network to produce and manipulate in the core loop of a learning algorithm to be executed in real time across a large network.

Effectively constructed embeddings may also have other benefits as well, e.g.

\begin{itemize}
\item Stability
\begin{itemize}
  \item If the neural net must directly output matrix entries, small numerical errors can blow up after matrix multiplication or inversion.
  \item A stable, ''pre-orthogonalized'' or otherwise well semantically ordered embedded representation can be more robust.
\end{itemize}

  \item Generalization and Feature Sharing
\begin{itemize}
  \item Different distributions $p_{k}$ may yield closely related factor matrices. By embedding them in a learned manifold, the net can generalize across distributions more effectively.
  \item The ideas in Section \ref{sec:hyper} regarding hypervector embedding extend this point even more dramatically
\end{itemize}
\end{itemize}

\subsubsection{Direct $\ell_{2}$ Projection of Factor Matrices}

A simple approach  to creating a compressed representation would be:

\begin{enumerate}
  \item Compute $\left(U_{k}, \Sigma_{k}, V_{k}\right)$ from partial SVD or rank-r factorization.
  \item Flatten or vectorize: $\operatorname{vec}_{k} \in \mathbb{R}^{(n \cdot r)}$ or $\mathbb{R}^{(2 n \cdot r)}$ if you include $U_{k}$ and $V_{k}$ separately.
  \item Project $\mathrm{vec}_{k}$ into a lower dimension $d$ via a linear map or a small MLP. E.g.,

$$
\mathbf{z}_{k}=W_{\text {proj }} \operatorname{vec}_{k} \in \mathbb{R}^{d}
$$

\end{enumerate}

One could also enforce orthogonality in $U$ and $V$ before flattening, then the net deals with ''stable'' columns.

This simple approach gives dramatic dimension reduction, but doesn't necessarily incorporate the geometry of $\omega$ or the distributions themselves, beyond being ''whatever the partial SVD gave.''    One might say it's purely mathematical rather than semantic or cognitive.

\subsubsection{Learned Autoencoder for Factor Triples}

A slightly subtler approach would be: Instead of using a linear projection, treat $\left(U_{k}, \Sigma_{k}, V_{k}\right)$ as a ''signal'' and train a small autoencoder, along the lines:

\begin{itemize}
  \item Encoder: $\operatorname{Enc}_{\phi}\left(\operatorname{vec}_{k}\right) \rightarrow \mathbf{z}_{k} \in \mathbb{R}^{d}$.
  \item Decoder: $\operatorname{Dec}_{\phi}\left(\mathbf{z}_{k}\right) \rightarrow \widehat{\mathrm{vec}_{k}} \approx\left(U_{k}, \Sigma_{k}, V_{k}\right)$.
\end{itemize}

A separate reconstruction loss ensures

$$
\left\|\widehat{\mathrm{vec}_{k}}-\mathrm{vec}_{k}\right\|^{2}
$$

\noindent stays low.

This potentially captures nonlinear relationships among factor entries, which is a major plus, likely more than compensating for the complication of having more moving parts to train.   The net itself and the autoencoder might require careful regularization to ensure stable re-inversion, e.g., orthogonality constraints or conditioning on $\Sigma$.

\subsubsection{Kernel Methods}

Kernel PCA is often used to embed data into a lower-dimensional manifold via a chosen kernel $\kappa(\mathbf{a}, \mathbf{b})$. In this context:

\begin{itemize}
  \item Data: The objects are factor matrices $\left(U_{k}, \Sigma_{k}, V_{k}\right)$.
  \item Kernel:

\begin{itemize}
  \item We define a function $\kappa\left(\left(U_{i}, \Sigma_{i}, V_{i}\right),\left(U_{j}, \Sigma_{j}, V_{j}\right)\right)$ that measures similarity among factorizations.
  \item Or, more directly, we define a kernel on $\hat{L}_{i}, \hat{L}_{j}$ themselves: e.g. $\kappa\left(\hat{L}_{i}, \hat{L}_{j}\right)$.
  \item Wasserstein-Related Kernel: Potentially, we incorporate $\omega(\mathbf{x}, \mathbf{y})$ or approximate transport distances among distributions to define a kernel. For instance,

$$
\kappa\left(\hat{L}_{i}, \hat{L}_{j}\right)=\exp \left(-\alpha W_{2}\left(\hat{L}_{i}, \hat{L}_{j}\right)^{2}\right)
$$

\noindent a Gaussian kernel on the Wasserstein distance between the distributions $p_{i}, p_{j}$.
\end{itemize}
\end{itemize}

With a Wasserstein metric based kernel, this kernel PCA might cluster factor matrices that represent ''similar geometry of transport.'' This ensures the embedded vectors are aligned with the environment's geometry, hopefully giving better generalization for distributions that are close in Wasserstein sense.

To apply Kernel PCA, then:

\begin{itemize}
  \item We solve for principal components in kernel space, obtaining a set of eigenvectors $\left\{\mathbf{z}_{\ell}\right\}$.
  \item Project each factor matrix to these top $\ell$ or $d$ principal coords. This yields the embedding $\mathbf{z}_{k}$.
\end{itemize}

This is a conceptually powerful and straightforward approach.  The operational downside is that Kernel PCA requires storing or computing kernel matrix $\mathcal{O}\left(K^{2}\right)$ if we do it offline with a dataset of size $K$.   Extending to online or large-scale might require fancy extensions such as approximate/streaming kernel PCA or Nystrom methods.

\subsubsection{Approximating Kernel PCA} \label{sec:kpca}

Several methods are available to approximate the principal subspace of $\mathbf{K}$ using fewer computations and memory.   Just as with exact kPCA, once we have the approximate principal components, we can map each data point (factor matrix) to a low-dimensional embedding $\mathbf{z}_{k}$ that captures nonlinear relationships.

\paragraph{Nystrom Method} The Nystrom method approximates an $N \times N$ kernel matrix $\mathbf{K}$ using a smaller subset of columns/rows:

\begin{enumerate}
  \item Select a Subset of size $m \ll N$, say $\mathcal{S}=\left\{s_{1}, \ldots, s_{m}\right\}$.
  \item Partition $\mathbf{K}$ as
  
$$
\mathbf{K}=\left(\begin{array}{ll}
\mathbf{K}_{\mathcal{S}, \mathcal{S}} & \mathbf{K}_{\mathcal{S}, \mathcal{R}} \\
\mathbf{K}_{\mathcal{R}, \mathcal{S}} & \mathbf{K}_{\mathcal{R}, \mathcal{R}}
\end{array}\right),
$$

\noindent where $\mathcal{R}$ is the set of the remaining $N-m$ indices.\\
\item Approximate $\mathbf{K}$ using only $\mathbf{K}_{\mathcal{S}, \mathcal{S}}$ and the cross-terms. For example,

$$
\mathbf{K}_{\text {approx }}=\mathbf{K}_{\mathcal{R}, \mathcal{S}} \mathbf{K}_{\mathcal{S}, \mathcal{S}}^{\dagger} \mathbf{K}_{\mathcal{S}, \mathcal{R}}
$$
 \item Eigen Decomposition: Factor $\mathbf{K}_{\mathcal{S}, \mathcal{S}}$ (an $m \times m$ matrix). The eigenvectors/eigenvalues provide a partial expansion that extends to the rest of $\mathbf{K}$.
\end{enumerate}

One can then do embedding of the multiple factor-matrix triples, i.e.:

\begin{itemize}
  \item Data Points: each factor matrix $\left(U_{k}, \Sigma_{k}, V_{k}\right)$ is treated as one ''data item.''
  \item Kernel: $\kappa\left(\left(U_{i}, \Sigma_{i}, V_{i}\right),\left(U_{j}, \Sigma_{j}, V_{j}\right)\right)$. We sample a subset $\mathcal{S}$ of size $m$, compute the kernel submatrix, factor it fully, then approximate $\mathbf{K}$.
  \item Result: We can embed all $N$ factor matrices into the top $d$-dimensional principal subspace with complexity $\mathcal{O}\left(m^{2} N\right)$ or $\mathcal{O}\left(m^{3}+m^{2} N\right)$ instead of $\mathcal{O}\left(N^{3}\right)$.
\end{itemize}

The overall Nystrom process looks like:

\begin{enumerate}
  \item In an offline step, gather factor matrices for some portion of the data (e.g., partial coverage or random episodes).
  \item Use Nystrom with a subset of size $m$.
  \item The top $d$ eigenvectors yield a partial principal components basis.
  \item Each data item $\left(U_{k}, \Sigma_{k}, V_{k}\right)$ is quickly projected onto these basis vectors to yield an embedding $\mathbf{z}_{k} \in \mathbb{R}^{d}$.
  \item The neural net can learn to predict $\mathbf{z}_{k}$ from distribution features $\mathbf{f}_{k}$.
\end{enumerate}

\paragraph{Random Feature Expansions}

Another approach to approximate a shift-invariant kernel , alternative to the Nystrom approach, is Random Feature Expansions \cite{rahimi2007random}.   One uses an approximant like $\kappa(\mathbf{x}, \mathbf{y})=\kappa(\mathbf{x}-\mathbf{y})$, e.g. Gaussian RBF. One draws random samples $\boldsymbol{\omega}_{\ell}$ from the Fourier transform of $\kappa$. Then one defines:

$$
\phi(\mathbf{x})=\sqrt{\frac{1}{D}}\left(\begin{array}{c}
\cos \left(\boldsymbol{\omega}_{1}^{\top} \mathbf{x}+b_{1}\right) \\
\vdots \\
\cos \left(\boldsymbol{\omega}_{D}^{\top} \mathbf{x}+b_{D}\right)
\end{array}\right),
$$

\noindent and approximates

$$
\kappa(\mathbf{x}, \mathbf{y}) \approx \phi(\mathbf{x})^{\top} \phi(\mathbf{y})
$$

To adapt this process to factor matrices, we would need a way to define $\phi\left(\left(U_{k}, \Sigma_{k}, V_{k}\right)\right)$ or $\phi\left(L_{k}\right)$.   E.g. we could potentially embed each factor matrix into a vector representation, then apply random features to replicate the kernel structure.   If $\kappa$ is a standard RBF in the ''flattened'' factor space, random features then yield $\phi\left(\mathrm{vec}_{k}\right) \approx$ embedding. Then standard PCA or direct linear dimension reduction might follow.

Typically, computing $\phi(\mathbf{x})$ is $\mathcal{O}(D)$, with a chosen $D \ll N$. This approach can handle large-scale problems by controlling $D$. The embedding $\phi(\mathbf{x})$ then gets used in a standard linear PCA or linear modeling, bypassing expensive kernel matrix builds.

\paragraph{Incremental/Streaming Kernel PCA}

These methods, or others, could be used together with an incremental Kernel PCA approach, roughly conceived as:

\begin{itemize}
  \item Start with an empty or small kernel model.
  \item For each new ''data item'' (factor matrix), incorporate it by updating an existing partial decomposition.
  \item Can incorporate approximate orthogonalization or rank trimming to keep the dimension bounded.
\end{itemize}

\noindent As new factor matrices $\left\{\left(U_{k}, \Sigma_{k}, V_{k}\right)\right\}$ come in from new distributions encountered by ActPC, we incrementally update the kernel subspace.  This prevents re-computing from scratch while allowing the system to adapt as the environment or policy evolves.

\paragraph{Overall Flow} Leveraging any of these kPCA approximation schemes within an information-geometry approach to ActPC, the overall process then becomes:

\begin{enumerate}
  \item Collect Factor Matrices: From partial SVD or rank- $r$ decompositions of the measure-dependent Laplacian at different episodes/time steps.
\item Approximate Kernel PCA:

\begin{itemize}
  \item Nystrom: If we want a fairly global perspective, choose a subset $\mathcal{S}$ of size $m$; compute $\mathbf{K}_{\mathcal{S}, \mathcal{S}}$ fully, then approximate the rest of $\mathbf{K}$.
  \item Random Features: If the kernel is RBF-like on the flattened factor vectors, build a random feature map $\phi$. Then do a simpler linear dimension reduction on $\phi\left(\mathrm{vec}_{k}\right)$.
  \item Incremental: As new factor matrices appear, keep updating the partial decomposition or random feature representation.
\end{itemize}

  \item Train a Regression Net: Let the learned principal components map each factor matrix to a vector $\mathbf{z}_{k} \in \mathbb{R}^{d}$. We then have pairs $\left(\mathbf{f}_{k}, \mathbf{z}_{k}\right)$. A neural net $f_{\theta}: \mathbf{f}_{k} \mapsto \mathbf{z}_{k}$ is trained, so at runtime we skip the heavy kernel step.

\item Runtime: For a new distribution $p\left(\mathbf{x} \mid \boldsymbol{\xi}_{t}\right)$ :

\begin{enumerate}
  \item Compute features $\mathbf{f}_{t}$.
  \item Predict $\hat{\mathbf{z}}_{t}=f_{\theta}\left(\mathbf{f}_{t}\right)$.
  \item Decode $\hat{\mathbf{z}}_{t}$ back to a rank- $r$ factor or approximate operator $\hat{L}$.
  \item Use $\hat{L}$ to form the geometry-aware update in the ActPC loop.
\end{enumerate}
\end{enumerate}

\paragraph{Recap: Why Approximate Kernel PCA Helps}  In any of these Approximate kPCA approaches, we avoid constructing an $\mathcal{O}\left(N^{2}\right)$ kernel matrix for a large dataset or long-run experience.   This is critical both for scalability reasons, and because in a large-scale or real-time RL or AGI scenario, the agent's distribution or factor matrices might shift significantly over time.
Approximate kernel methods allow efficient continuous integration of new data points, enabling the embedding to remain relevant as the environment changes.  The key step needed for the neural approximator of the inverse measure-dependent Laplacian needed for the Wasserstein metric -- embedding each factor matrix $\hat{L}_{k}$ or $\left(U_{k}, \Sigma_{k}, V_{k}\right)$ into $\mathbf{z}_{k} \in \mathbb{R}^{d}$ -- becomes computationally feasible.

That is, we:

\begin{itemize}
  \item Preserve the main advantage of kernel PCA: discovering low-dimensional embeddings that reflect the environment's geometry.
  \item Control memory/computational cost, suitable for real-time or large-scale ActPC with big neural networks.
  \item Enable a ''once learned'' principal subspace to handle new factor matrices on-the-fly by mapping them to the subspace quickly.
\end{itemize}

\subsection{ Predicting the Embedding Vector}

We return now to the main flow of leveraging compressed representations for making the neural approximator of the inverse measure-dependent Laplacian scalable.   Regardless of how we define the embedding $\mathbf{z}_{k}$ -- via linear projection, autoencoder, kernel PCA or something fancier -- the final step is to train a function:

$$
\mathbf{z}_{k}=g\left(\operatorname{vec}_{k}\right) 
$$

\noindent and then a neural net $f_{\theta}$ that, given the distribution features $\mathbf{f}_{k}$, predicts

$$
\widehat{\mathbf{z}_{k}}=f_{\theta}\left(\mathbf{f}_{k}\right)
$$

At inference time:

\begin{enumerate}
  \item We compute $\mathbf{f}_{k}$ from the current distribution.
  \item $f_{\theta}\left(\mathbf{f}_{k}\right)=$ embedding $\widehat{\mathbf{z}_{k}}$.
  \item We decode it (via a known transformation or an invertible mapping) to approximate ( $U_{k}, \Sigma_{k}, V_{k}$ ). If we used a kernel PCA approach, we might keep a ''decoder'' system or store the principal components for expansion.
\end{enumerate}

\paragraph{ Low-Frequency Recalibration:}  To make sure all this works effectively, it may be helpful if the agent occasionally does a partial factorization of a newly visited distribution's inverse Laplacian, obtains a ''ground-truth'' $\mathbf{z}_{\text {true }}$, and updates the network:

$$
\left\|\mathbf{z}_{\text {true }}-f_{\theta}\left(\mathbf{f}_{t}\right)\right\|^{2} \rightarrow \min _{\theta}
$$

\paragraph{Fast Inference:}. To speed up inference one can break things up int little bits, i.e.: On each step or small batch, the system only calls $f_{\theta}$. The re-expansion to $(U, \Sigma, V)$ or the approximate matrix $\hat{L}$ is thus $\mathcal{O}(n r)$ at worst, or $\mathcal{O}(d)$ to produce the embedding itself.

\subsection{Overall Methodology for Leveraging Compressed Representation to Guide Neural Approximators}

Summing up, the methodology we have presented is as follows:

\paragraph{Offline / Periodic Phase}
\begin{itemize}
  \item Gather $\left\{p_{k}\right\}$ from sampling policy or from offline demonstrations.
  \item For each $p_{k}$, do partial factorization of $L\left(p_{k}\right)$. 
  \item Convert to some embedding $\mathbf{z}_{k}$ (via linear map, autoencoder, or kernel PCA).
  \item Train $f_{\theta}: \mathbf{f}_{k} \mapsto \mathbf{z}_{k}$. Train an associated decoder if needed.
\end{itemize}

\paragraph{Online Real-time Phase}

\begin{itemize}
  \item At each step, compute distribution features $\mathbf{f}_{t}$.
  \item Evaluate $\widehat{\mathbf{z}_{t}}=f_{\theta}\left(\mathbf{f}_{t}\right)$.
  \item Decode or re-expand $\widehat{(U, \Sigma, V)}$.
  \item Form or invert $\hat{L}^{\dagger}$, use in Wasserstein-based ActPC update.
  \item Periodically refine training with newly observed partial factorizations or new environment conditions.
\end{itemize}

 It may be worthwhile to experiment with many variations of this methodology, however, the particular approach that seems most promising to us is to:  

\begin{itemize}
  \item Use approximate kernel PCA to embed measure-dependent Laplacians into a low-dimensional latent space.
  \item Define the kernel $\kappa\left(L_{i}, L_{j}\right)$ based on the Wasserstein metric, capturing structural relationships between Laplacians.
\end{itemize}

\noindent The suggested methodology then trains the neural approximator to map distribution features $\mathbf{f}$ directly to the kernel-PCA embedding, allowing rapid inference of low-rank operators.

\section{Replacing KL with Wasserstein for ActPC Prediction Error Assessment} \label{sec:wasser_outer} 

ActPC is traditionally formulated using KL divergence to measure the prediction error of a neural network system relative to its targets.  This is a sensible enough choice, and in Friston's framework it emerges naturally as a reorganization of the ''free energy principle.''

From an AI view, however, there may be significant advantages to replacing KL divergence with Wasserstein distance in this role.   These advantages are especially clear in an ActPC-Geom context, because one gets benefits of synergy with the use of Wasserstein distance in the information geometry being leveraged to accelerate and focus internal ActPC learning dynamics.   However, once the possibility is opened up, it appears there are some advantages to such a switch even aside from the information-geometric context.

We consider in this section how one might replace KL divergence with a Wasserstein-based divergence measure in an Active Predictive Coding (ActPC) framework, and how that substitution could bring the ''reward-driven information divergence cost'' more closely into line with the system's internal local transport geometry. We focus on the why and how of using a Wasserstein-based cost function instead of KL, and on the potential benefits in aligning external reward signals with internal geometry-based updates.

Before digging into this, however, we briefly consider how one might conceptually justify this shift in terms of the broader principles underlying ActPC.

\subsection{Optimal Transport as a Cognitively Relevant Least Action Principle}

In Karl Friston's framework, which is one of the core motivations for the ActPC approach to neural learning, ''minimizing free energy'' is taken as a core principle and is shown to be equivalent (under certain assumptions) to minimizing a KL divergence between the organism's ''recognition density'' and the environment's ''true'' density.  This underpins a sort of ''thermodynamic'' or ''variational'' interpretation of the system's self-organization.

If we replace KL with a Wasserstein (optimal transport) cost in ActPC, we lose the immediate link to that free-energy argument, because the free-energy principle relies on a KL-based evidence bound.  However, the Wasserstein approach corresponds to a loosely analogous conceptual rationale, which sees a learning system's evolution as ''optimal transport''-like or ''minimal cost'' flow in probability space.  

In this section we briefly elaborate this perspective, in the hopes that it provides some philosophical and physical motivation for a shift from KL to Wasserstein in the context of prediction error assessment for ActPC.   Although ultimately, the reason for this shift in an AI context would simply be the improved effectiveness of ActPC based AI systems which we hypothesize may result.

\paragraph{Physics Metaphors for the Wasserstein Approach} In mathematical physics and PDE theory, it is common to interpret gradient flows in the Wasserstein metric in an intuitive/visceral way along the lines:

\begin{itemize}
  \item In ''Otto's geometry'' or the ''Benamou-Brenier formulation,'' we can interpret the evolution of probability distributions as a gradient flow in the space of distributions endowed with the Wasserstein metric.
  \item The physical analogy is: ''mass'' in the distribution tries to flow to a new state in a way that minimizes total transport cost (like minimal Earth-mover's distance) subject to certain forces or potential energies.
\end{itemize}

\noindent In this spirit, one might motivate a system that continuously evolves its ''internal representation'' or ''recognized distribution'' via Wasserstein gradient flows to align better with the environment's distribution. This is basically a ''principle of least action in the space of probability distributions,'' stating that an intelligent system spontaneously chooses paths that minimize the transport (Earth-mover) cost from prior beliefs to updated beliefs given new evidence.

That is: Wasserstein-based ActPC can be seen as the system's ''flow in probability space'' following a path of minimal mass transport cost to reduce mismatch with the environment's actual distribution. Instead of free-energy (KL-based) minimization, we have optimal mass transport minimization or ''least transport path.''

\paragraph{JKO Scheme and a ''Physical'' Interpretation} One can flesh this idea out further, and connect it with a variant of the ''free energy principle'', using the Jordan-Kinderlehrer-Otto (JKO) Scheme:

\begin{itemize}
  \item In PDE or measure theory, the JKO scheme shows how the Fokker-Planck (or diffusion) equation in probability space can be discretized as repeated steps of ''minimize functional plus Wasserstein distance.''
  \item Each step picks the distribution that is closest (in Wasserstein) to the old distribution but also reduces free energy or some potential.
\end{itemize}

Analogizing from the physics domain to the AI domain, one might say

\begin{itemize}
  \item If one adopts a ''predictive coding in distribution space,'' each micro-iteration might do: ''Pick the internal distribution that is a small Wasserstein step from the old distribution but also best matches incoming data.''
  \item We  can then interpret the system as spontaneously traveling along a ''gradient flow'' in the Wasserstein geometry.
\end{itemize}

\noindent I.e. one could say the system obeys a ''Wasserstein gradient flow principle,'' reminiscent of least-action or minimum free energy but in an optimal transport sense.

\paragraph{Physically/Biologically Inspired Terminology}  Pursuing this conceptual direction a little more impressionistically: If the free-energy principle is couched as ''the system tries to avoid surprise in a thermodynamic sense,'' a Wasserstein variant might be phrased as ''The system ''minimizes mass transport in updating its representation'' -- meaning it changes its internal distribution in the ''smoothest,'' ''least cost'' manner consistent with the new data or environment, akin to ''low friction'' or ''least physical work'' in probability space.

One might even call this a ''Minimum Probability Transport Principle,'' or ''Optimal Transport Action'' principle -- a conceptual anchor akin to how thermodynamic free energy frames the KL divergence approach.

In sum,  choosing Wasserstein for ActPC can be motivated by ''the system's distribution evolves by minimal transport cost to reduce predictive error,'' paralleling how the free energy principle sees ''the system's distribution evolves by minimal KL-based surprise.'' This is not a standard principle in neuroscience or biology at the present time, but it is mathematically coherent and physically reminiscent of ''optimal transport as the path of least overall cost in measure space.''

\subsubsection{Conceptual Advantages of the Wasserstein Approach for Explaining Intelligent Action}

Taking the above ideas a step further, one can make a reasonable conceptual argument that Wasserstein-inspired approaches (i.e., evolving distributions via least transport action in probability space) can feel more natural for action dynamics in complex AI systems, relative to Friston's free-energy principle (which is more obviously suited to perceptual inference).   The following points are not meant to dismiss the free-energy principle's broad scope, but merely to highlight how optimal transport or Wasserstein frameworks may resonate more directly with the action side of intelligent systems.

\paragraph{The Friston Principle's Emphasis on Perception}  In Friston's work, ''minimize variational free energy'' is effectively ''minimize KL divergence between your internal recognition density and the environment's true distribution.''   The standard interpretation is that: The system reduces surprise, guiding perceptual inference to match actual sensory data.

The principle does attempt to unify action and perception by saying ''action ensures sensory states remain within predictable bounds.''
But this aspect can feel somewhat awkward: the system tries to suppress surprising inputs. One might see it as ''avoid states that produce large KL divergences,'' but why that specifically guides active behavior can be tricky to interpret in real tasks.

Hence, many find it straightforward that free-energy / KL minimization conceptually fits perceptual or cognitive modeling, whereas the action part can feel forced: ''The system picks motor commands to minimize sensory surprise,'' which is not always the most intuitive formulation for complex goal-directed tasks.

\paragraph{Why a Wasserstein or ''Least-Action in Distribution Space'' Perspective Can Aid in Explaining Action}  Complex AI systems, especially those controlling robots or planning intricate actions, often view action as moving from one state distribution to another in an optimal or cost-minimizing manner:

\begin{itemize}
  \item Path or trajectory optimization is frequently about ''Which sequence of control signals yields minimal cost to get from distribution $p_{0}$ to distribution $p_{\text {goal }}$ ?''
  \item If the cost is transport-based (like how many resources or how much energy/time it takes to reconfigure the environment or the agent's state), a mass transport/cost-of-movement perspective can be extremely direct.
\end{itemize}

With this in mind: If an intelligent system represents action outcomes as probability distributions (e.g. ''If I do action $a$, the environment distribution changes from $p$ to $p^{\prime \prime \prime}$ ), it becomes natural to talk about: ''Minimizing the transport cost of going from the old distribution to the new distribution that matches my goals or observations.''

Or to phrase it slightly differently: The system's ''actions'' shift the environment's distribution or the agent's internal state distribution in a manner that is geometrically minimal in an optimal transport sense.

\paragraph{ Overcoming the ''Hide From Surprise'' Critique} Friston's free-energy principle, when applied to learning and behavior, has been questioned via a ''hide from surprise'' critique -- which argues that the principle's focus on minimizing surprise (or prediction error) may lead to avoidant or overly conservative strategies, which can limit learning, exploration, and goal-directed action.  Using a Wasserstein approach for action bypasses this in a clear and simple way:

\begin{itemize}
  \item The system tries to directly move from the old distribution to a new one that aligns with desired outcomes, with minimal ''mass transport.''
  \item This is conceptually akin to classical ''least action'' in physics, or ''lowest energy path'' but applied in the space of probability measures.
  \item Instead of ''avoiding surprising inputs,'' we have ''achieving a target distribution in the most efficient movement sense.''
  \item In many tasks, that target distribution might correspond to a desired environmental configuration or an internal policy for future states.
\end{itemize}

\noindent The Wasserstein approach seems to resonate more naturally with goal-directed or manipulative actions -- ''I want to reconfigure the environment's state distribution'' -- than the purely ''avoid surprise in your sensory predictions'' framing.

The basic dichotomy here is transport vs. surprise:

\begin{itemize}
  \item Friston's KL approach: ''Don't let your sensory data deviate from your predictions.''  This feels somewhat avoidant regarding new states, even if they might be beneficial.
  \item Wasserstein approach: ''Move your distribution from state A to state B along a minimal transport cost path.''  This feels proactive and action-oriented -- the system invests the ''least work'' in bridging from current distribution to a new distribution consistent with new goals or new evidence.
\end{itemize}

\paragraph{Intuitive for Planning \& Control:}   Optimal transport in classical control theory or path planning is a well-known approach to find a ''lowest cost route.'' In high-dimensional probability space, this metaphor extends nicely to define action strategies in advanced AI systems.

In real-world mobile robotics, for example, agents typically do not simply ''hide from surprises,'' but often seek beneficial novelty or states that give advantage.  A ''least mass transport cost'' lens allows beneficial transitions to still be minimal cost in a geometry sense, but does not default to avoiding novelty; it just ensures the path from old distribution to new is minimal. This can incorporate exploration if it lowers the overall cost of distribution shift needed to achieve goals.

Overall, it seems Wasserstein or ''optimal transport'' principles might provide a more natural conceptual foundation for how a system chooses its actions.  The KL-based free-energy principle remains elegant for explaining perceptual inference; although, one can also frame an argument why the Wasserstein approach can provide additional explanatory value in the perception case as well.

\subsubsection{Advantages of the Wasserstein Approach for Explaining Perception}

While the classic Friston KL approach feels quite natural in the context of perception, one can also question whether it has the power to fully explain the high level of inventiveness that human perception sometimes displays.  One can frame an argument that a Wasserstein metric approach is conceptually preferable in the context of perception as well, though the argument is weaker here than in the case of action.

As a first point, Wasserstein can have superior mathematical properties in the context of partially-overlapping or multimodal distributions.  KL can blow up if distributions don't overlap, or behave strangely if they're multi-modal. The agent might produce big or abrupt updates to avoid infinite divergence.   On the other hand, Wasserstein remains well-defined even if there's little overlap. The transport cost is finite if each distribution has finite mass. This is smooth for partially disjoint or multi-modal distributions.

The Wasserstein distance's use of a ground metric may be an asset in this context.   For complex perceptual input (e.g. images, audio), defining a ground metric that respects local features, shapes, transformations can let Wasserstein capture a more holistic or structured distance. If small local shifts in image space are fairly cheap, then the agent's mismatch is small if only small pixel movements are needed. By contrast, KL might treat a shift of mass from region A to region B as massive if the distributions do not overlap, ignoring the fact they're only 5 pixels apart (of course this would not happen if the underlying distributional model were complete enough, but in real life it basically never will be).

Framing this in a more explicit AGI-theory context, one could argue that AGI perception requires both

\begin{itemize}
\item Integrating wholly or partially unknown phenomena into known structure,
\item Possibly relabeling, recombining, or synthesizing existing categories to handle new data.
\end{itemize}

\noindent Wasserstein fosters a ''transform or extend your distribution to meet the new environment state'' in a natural geometric sense. Instead of ''reject or label surprising data,'' the system sees ''where in the old representation can we shift mass to fit x without large global disruptions?''

Hence potentially one obtains a perceptual system that is:

\begin{itemize}
\item Less prone to ignoring or discarding partially novel data as ''too surprising.''
\item More apt to see how new data can be mapped or transported from existing manifold structure.
\end{itemize}

Framed in terms of the cognitive-science concept of compositionality, one might say that an AGI perceptual faculty should

\begin{itemize}
\item Combine known object parts or known transformations to interpret partially unseen objects.
\item If only a small ''transport'' in shape or feature space is needed, then the mismatch is small in a Wasserstein sense, even though the data might be quite ''surprising'' in log-likelihood terms if that shape was never exactly encountered.
\end{itemize}

\noindent So the system re-labors or extends old categories in a compositional manner to unify new and old data. ''We just shift part of the distribution from shape A to shape B,'' which is a direct step in the transport manifold.

This compositionality framing resonates well with the ideas of Section \ref{sec:hyper} which explicitly leverages hypervector embeddings with compositional algebra to approximate Wasserstein information geometry.

\subsubsection{Potential Advantages of the Wasserstein Approach for Explaining Logical Cognition}

Extending the above notions even more speculatively, one could pose an argument that the Wasserstein metric lends itself more naturally to explaining more abstract aspects of intelligence such as logical reasoning.   Impressionistically one might think about:

\begin{itemize}
\item Physical ''Movement'' of Concepts

\begin{itemize}
\item Wasserstein measures how ''far'' one distribution of ideas or states must ''move'' in logical or feature space to match another.
\item KL only measures how ''unexpected'' new information is under one's prior.
\end{itemize}

\item Smooth Handling of Novelty

\begin{itemize}
\item KL can blow up if new ideas don't overlap sufficiently with old concepts, leading to abrupt changes.
\item Wasserstein provides finite, continuous ''transport'' costs even for disjoint sets, allowing gentler, more structured transitions when new logical elements arise.
\end{itemize}

\item Compositional Extension

\begin{itemize}
\item Logical reasoning often requires combining or rearranging known properties.
\item Wasserstein sees this as shifting ''mass'' from old concepts to new combinations, which aligns well with building new logical constructs step-by-step, rather than simply labeling them ''surprising.''
\end{itemize}

\item Aligning With Structured Domains

\begin{itemize}
\item Logic is about rearranging discrete statements with well-defined relationships.
\item Wasserstein can incorporate domain-specific metrics (e.g. edit distances for symbolic transformations), making the ''cost'' of changing or adding logical elements more intuitive than abstract log-likelihoods.
\end{itemize}

\end{itemize}

Overall, for conceptual reasons along these lines, it seems the Wasserstein metric is comparatively more naturally interpretable for logical or compositional reasoning as it reflects gradual, structured reconfiguration of ideas in a conceptual space, instead of forcing a ''surprise minimization'' lens that can rigidly penalize novel or disjoint logical constructs.

In the end, of course, these conceptual arguments will matter only inasmuch as they pan out in terms of detailed mathematical formulations and corresponding highly functional AI systems.  Our aim with these physical/biological metaphors and analogical explanations is mainly to provide heuristic motivation for the following more technical developments.

\subsubsection{Wasserstein \& Neuroscience}

Finally, without venturing to do theoretical neuroscience in any serious way, one can extend the above points into a high level argument for the Wasserstein/optimal transport principle as a conceptual framework for understanding many aspects of brain function, complementing Friston's free-energy principle. While the free-energy principle focuses on minimizing surprise -- interpreted as reducing the KL divergence between an internal model and external inputs -- Wasserstein emphasizes the cost of transporting probability mass in a state or feature space. This perspective aligns with the brain's need to reorganize its distributed activity patterns to adapt to novel sensory inputs or environmental demands. Instead of simply rejecting surprising inputs or confining itself to predictable states, a Wasserstein-inspired brain seeks to move its internal representations along minimal-cost paths, actively reshaping and extending its knowledge to integrate new information. This dynamic transformation, grounded in geometry and cost minimization, captures a proactive and compositional process more aligned with learning and adaptive intelligence.

From a biological standpoint, the brain's transitions between activity states can be seen as ''transporting mass'' across neural assemblies, with metabolic and organizational costs akin to Wasserstein transport costs. Neural and glial systems coordinate these transitions, managing resource allocation and maintaining homeostasis. The brain's modular, multi-layered structure, with interconnections between cortical, subcortical, and glial systems, naturally supports a framework where local neural populations manage distributions of activity that flow to match environmental demands. Minimizing transport cost reflects not only the reorganization of neural firing patterns but also the metabolic efficiency of adjusting vascular and glial support for these transitions. This view ties the brain's energy constraints and distributed computations directly to a principle of optimal transport.

Wasserstein's emphasis on incremental, structured changes to internal representations also resonates with the brain's capacity for generalization and learning. Unlike the free-energy approach, which can appear reactive?focused on minimizing mismatch?the Wasserstein framework enables smooth transitions from prior knowledge to new representations, even when inputs are novel or disjoint. This proactive reconfiguration fosters generalizable learning and robust adaptation, resonating well with the brain's ability to build compositional, multi-scale models of its environment. While the brain undoubtedly employs many overlapping principles, the Wasserstein approach at least somewhat captures the brain's ability to reorganize and extend its internal representations with efficiency and flexibility.

\subsection{ KL vs. Wasserstein as the Divergence Measure in ActPC}

Now let us proceed through some technical details.   As we've noted, ActPC conventionally uses KL-like divergence for the mismatch between predicted and actual (or target) distributions, e.g. formulations like

$$
D_{\mathrm{KL}}(q(\mathbf{x}), p(\mathbf{x} \mid \theta)),
$$

\noindent where $q(\mathbf{x})$ is the ''true'' or observed distribution (including reward distribution) and $p$ is the agent's predicted distribution parameterized by $\theta$. This provides a measure of how the agent's prediction fails to match reality (or desired outcome).

KL-based geometry would harmonize well with information geometry based on a Fisher-Rao metric.   However, we have chosen a Wasserstein metric instead for our work here, for various reasons including its natural foundation in a ground metric, which makes it conceptually straightforward to apply in the context of an AI system with its own native ways of measuring distance between concepts, percepts and actions.

Suppose instead we define a Wasserstein-based divergence as the agent's cost measure, i.e.

$$
F_{\mathrm{Wass}}(\theta)=W_{2}(q(\mathbf{x}), p(\mathbf{x} \mid \theta)),
$$

\noindent where $W_{2}(\cdot, \cdot)$ is the Wasserstein distance.  This implies the agent tries to minimize the ''earth mover's distance'' between the distribution it predicts for outcomes (including reward) and the actual distribution $q$.

We argue this has clear technical advantages, along with and aligned with the more conceptual advantages described above; it

\begin{itemize}
  \item Aligns the external mismatch measure (reward cost) with the internal local geometry (transport cost) that shapes parameter updates.
  \item Is more stable when supports do not overlap.
  \item Potentially leads to more consistent local gradient directions between the ''outer loop'' cost and the ''inner loop'' geometry.
  \end{itemize}
  
  On the other hand, it also incurs some costs in a practical sense:

  \begin{itemize}
  \item Computing or approximating the Wasserstein distance on an external measure can be more expensive than computing KL.
  \item Implementation details may require specialized approximations (e.g. neural approximators, etc.).
\end{itemize}

Our   main task in the remaining parts of this section is to argue that these advantages are significant and these costs are manageable.

\subsection{Closer Connection Between External Cost \& Internal Geometry}

If the external cost is literally the Wasserstein distance between the agent's predicted reward distribution and the environment's actual reward distribution, then the agent's global objective (minimizing that distance) is more directly aligned with how the measure-dependent Laplacian $\hat{L}(p)$ shapes local parameter updates.   The system's movements in parameter space can be seen as physically ''moving mass'' from the agent's predicted distribution to the environment's distribution, matching the conceptual story from both the agent's viewpoint (local updates) and environment viewpoint (reward mismatch).   One obtains a more coherent alignment between ''how we measure mismatch with the environment's distribution'' and ''how we transport the agent's distribution internally.''

I.e.: In effect, ''outer loop'' and ''inner loop'' are describing the same transport geometry in distribution space.  Because the local gradient directions from the external cost and the local geometry are not in conflict, the agent's parameter updates can proceed in a more ''consistent'' direction -- potentially yielding smoother and more direct convergence to solutions that actually match the environment's reward distribution.   For instance, if both external cost and internal geometry share the same notion of ''transport,'' the system might avoid contradictory gradient directions, making learning more stable or convergent.

Further synergy may be provided via use of a common ground metric for the Wasserstein metrics used for system-level error assessment and for internal information-geometric guidance.   This should make sense given that the natural distance/cost structure on the environment should be similar to the natural distance/cost structure on the portions of a cognitive system concerned with directly modeling the environment.

In the next sections we articulate some formal propositions illustrating aspects of the synergy obtained by using the same measure in the inner and outer loop.   These relatively simple propositions don't fully capture the cognitive or mathematical aspects of this synergy, there are merely some of the simpler aspects to articulate -- dipping a foot tentatively in the deep waters.

\subsection{Transfer of Continuity and Scale Properties from Outer to Inner Loop} \label{sec:scale-prop}

We explore in this section a specific sense in which using Wasserstein for ''both external reward divergence and internal geometry'' can systematically reduce distortions between the agent's ''outer'' objective and ''inner'' distribution transformations, offering more coherent learning dynamics.   We conjecture this may lead to greater learning efficiency, though showing this formally  becomes more complex and is left for later theoretical works or practical experiments.

Specifically: Suppose the environment's reward distribution $q(\mathbf{r})$ and the agent's predicted distribution $p(\mathbf{r} \mid \theta)$ are both Lipschitz continuous with respect to the same ground metric $\omega\left(\mathbf{r}, \mathbf{r}^{\prime}\right)$. For instance,

$$
\left|W_{2}\left(q, p\left(\theta_{1}\right)\right)-W_{2}\left(q, p\left(\theta_{2}\right)\right)\right| \leq L\left\|\theta_{1}-\theta_{2}\right\|
$$

If we also define the internal measure-dependent operator for $\mathbf{x}$-space (or $\mathbf{r}$-space) using the same underlying metric $\omega$, then we argue that the following properties will hold:

\begin{itemize}
  \item Local Continuity: The cost function w.r.t. parameter shifts $\left\|\theta_{1}-\theta_{2}\right\|$ might reflect the same scale of mass transport that the internal geometry uses.
  \item Scale Matching: If the environment distribution's shape changes at a certain scale, the internal transport operator can also exhibit changes at a consistent scale, letting the system ''react'' in proportion to the environment's domain distances.
\end{itemize}

To frame this more precisely:

\paragraph{Proposition (Wasserstein Lipschitz ActPC Property)} {\it  Assume there is a constant $L>0$ such that for all $\theta_{1}, \theta_{2} \in \Theta$,

$$
\left|W_{2}\left(q, p\left(r \mid \theta_{1}\right)\right)-W_{2}\left(q, p\left(r \mid \theta_{2}\right)\right)\right| \leq L\left\|\theta_{1}-\theta_{2}\right\| .
$$

\noindent (Hence, the environment's cost function $\Delta_{\text {env }}(\theta)=W_{2}(q, p(r \mid \theta))$ is $L$-Lipschitz in $\theta$.)

Assume the agent's local measure-dependent operator (used for parameter updates in an ActPC or gradient-flow sense) references the same ground metric $\omega$. That is, small shifts in $\theta$ produce small Wasserstein displacements of $p(\mathbf{x} \mid \theta)$ or $p(r \mid \theta)$ in the same $\omega$-based geometry.

Then we obtain desirable properties including:

\begin{enumerate}
  \item Local Continuity: The external cost $\Delta_{\text {env }}(\theta)$ changes at most $L\left\|\theta_{1}-\theta_{2}\right\|$ over small moves in parameter space.
  \item Scale Matching: If the environment's distribution changes at a certain scale in $\omega$ distance, the agent's internal geometry will reflect comparable transport costs at that scale, aligning the environment's notion of ''distance'' with the agent's local distribution shifts.
\end{enumerate}
} %% end italics

\subsubsection{More Rigorous Statement of Scale Matching Property}

The precise best way to formulate the ''scale matching'' property here is not entirely clear.   A more rigorous, though perhaps somewhat overcomplex approach is as follows.  Let:

\begin{enumerate}
  \item $\mathcal{R}$ be a space of reward outcomes (or states) equipped with a ground metric $\omega: \mathcal{R} \times$ $\mathcal{R} \rightarrow \mathbb{R}_{\geq 0}$.
  \item $q(r)$ be the environment's (true) reward distribution over $\mathcal{R}$.
  \item $p(r \mid \theta)$ be the agent's parametric family of predicted reward distributions, $\theta \in \Theta \subset \mathbb{R}^{m}$
  \item $W_{2}(q, p(r \mid \theta))$ be the Wasserstein distance between $q$ and $p$.
  \item $\hat{L}(p)$ be the internal measure-dependent operator used in the agent's local geometry, constructed from $\omega$.
\end{enumerate}

Assume as above that: There exists a Lipschitz constant $L>0$ such that:

$$
\left|W_{2}\left(q, p\left(r \mid \theta_{1}\right)\right)-W_{2}\left(q, p\left(r \mid \theta_{2}\right)\right)\right| \leq L\left\|\theta_{1}-\theta_{2}\right\| \quad \forall \theta_{1}, \theta_{2} \in \Theta .
$$

Then, assume a scale-matching condition formulated as follows:  For any scale $\delta>0$ in the environment's distribution domain $\mathcal{R}$, suppose that shifting the agent's parameter $\theta$ by an amount $\|\delta \theta\|$ results in a distribution shift of magnitude $\delta$ in $\omega$ -distance-i.e.,

$$
W_{2}(p(r \mid \theta), p(r \mid \theta+\delta \theta))=\delta
$$

\noindent where $\delta \theta$ is small enough that linearization or local geometry is valid. 

Then, finally, given all these assumptions, what we want is: If the environment's distribution changes or exhibits a variation at scale $\delta$ in $\omega$-distance, the agent's internal geometry also attributes a cost of approximately $\delta$ for that distributional shift, ensuring the agent's local measure-dependent operator can ''see'' and react proportionately to that variation.

Even more concretely, what we want is: For each scale $\delta$ relevant to the environment's distribution, if $\Delta_{\text {env }}$ changes by some amount (bounded by $L \cdot\|\delta \theta\|$ ) when we move $\theta$ by $\delta \theta$, then the internal measuredependent operator $\hat{L}(p)$ or $\hat{L}(p)^{\dagger}$ will assign a transport cost of the same order (i.e., also $\approx \delta$ ) to the shift $p(r \mid \theta) \rightarrow p(r \mid \theta+\delta \theta)$. Thus, external scale $\delta$ in $\omega$-space is mirrored by an internal scale $\delta$ in the agent's measure-dependent geometry.

Given all these assumptions, we should be able to derive a conclusion of the form: {\it The environment's notion of ''distance'' or scale $\delta$ in reward distribution space is matched by the agent's local measure of distribution shift -- i.e. both rely on the same $\omega$.   That is: small changes in $\theta$ that produce a $\delta$-scale shift externally are likewise recognized as a $\delta$-scale shift internally.}

Proof sketches for these local continuity and scale matching properties are given in Appendix \ref{app:scale-match}.

\subsubsection{Exploring Weakening of the Lipschitz Condition}. One may wonder whether a Lipschitz condition is the best assumption for the ''scale matching'' property of the sort addressed in the Wasserstein Lipschitz ActPC Property explored above, or if a weaker continuity condition might suffice.

The Lipschitz assumption made above says that:

$$
\left|W_{2}\left(q, p\left(r \mid \theta_{1}\right)\right)-W_{2}\left(q, p\left(r \mid \theta_{2}\right)\right)\right| \leq L\left\|\theta_{1}-\theta_{2}\right\|
$$

\noindent for all $\theta_{1}, \theta_{2}$. This form of bounding is:

\begin{enumerate}
  \item Linear in Parameter Distance: We get a direct ratio $\Delta_{\text {env }}\left(\theta_{2}\right)-\Delta_{\text {env }}\left(\theta_{1}\right) \leq L \| \theta_{2}-$ $\theta_{1} \|$.
  \item Global (or at least for all relevant $\theta$ in $\Theta$ ): If it holds for all parameter pairs, we can ensure uniform local continuity across the domain.
\end{enumerate}

\noindent This Lipschitz property is convenient because it directly yields the statement about local continuity and scale matching: a small param shift $\left\|\theta_{2}-\theta_{1}\right\|$ translates into a proportionally bounded difference in cost $\left|W_{2}\left(q, p\left(\theta_{1}\right)\right)-W_{2}\left(q, p\left(\theta_{2}\right)\right)\right|$. In turn, we can say that if the agent's measure-dependent operator is built on the same ground metric, it recognizes distribution shifts in the same scale.

However, we could weaken the Lipschitz requirement in various ways, preserving the same sort of qualitative conclusion.

For instance, we could make the Lipschitz assumption local instead of global.  In this case we only require

$$
\left|W_{2}\left(q, p\left(r \mid \theta_{1}\right)\right)-W_{2}\left(q, p\left(r \mid \theta_{2}\right)\right)\right| \leq L\left(\theta_{0}\right)\left\|\theta_{1}-\theta_{2}\right\|
$$

\noindent for $\theta_{1}, \theta_{2}$ in some neighborhood of a reference $\theta_{0}$.

It's not hard to see that this still ensures ''scale matching'' in that local region near $\theta_{0}$.  The proof of local continuity and scale alignment given above then holds locally around each point.

An even weaker condition would be to assume  Holder Continuity (or $\alpha$-Holder) instead of Lipschitz, e.g. assume

$$
\left|W_{2}\left(q, p\left(r \mid \theta_{1}\right)\right)-W_{2}\left(q, p\left(r \mid \theta_{2}\right)\right)\right| \leq C\left\|\theta_{1}-\theta_{2}\right\|^{\alpha},
$$

\noindent for some $0<\alpha \leq 1$.  

 If $\alpha=1$, that's Lipschitz, but if $\alpha<1$, the cost can grow ''sub-linearly'' in parameter distance.

This might still yield a mild form of ''scale matching'': bigger parameter moves produce bigger changes in cost, though potentially at a slower rate.  The main consequence of this sort of ''diluted'' scale matching is that the agent's updates remain stable as long as the environment cost does not blow up unpredictably ... but the direct linear ratio is lost. This seems a potentially worthwhile direction for investigation, but we will leave it to explore later if it turns out to be useful.

A yet weaker approach would be to assume only that $W_{2}(q, p(\theta))$ is continuous in $\theta$.   In this case, we still get a minimal guarantee that arbitrarily small param changes do not produce big leaps in cost.   However, continuity alone does not provide an explicit bounding ratio -- there is no direct numerical guarantee that a $\delta \theta$ yields at most some factor times $\|\delta \theta\|$ in cost difference.   The ''scale matching'' statement becomes extremely weak: we can say small parameter shifts yield small changes in cost, but we cannot quantify a uniform scaling factor across the domain, or even a local neighborhood.

Overall it seems clear that: Lipschitz is not the only possible condition that can yield this kind of result, nor necessarily the most useful condition to assume.   It is just a convenient, stronger assumption that makes the statement simpler and clearer. In practical or more general contexts, local Lipschitz or Holder continuity might do just as well for ensuring the environment's ''scale'' changes get recognized proportionally by the agent's internal measure-based geometry.

Which variant of these formal results (if any) is most useful will likely become clear only after more practical experimentation with ActPC-Geom techniques has been performed.

\subsubsection{Making the Lipschitz Condition Probabilistic}  \label{sec:prob-lip} 

A different sort of variation of the Wasserstein Lipschitz ActPC Property is one in which the Lipschitz condition is made probabilistic.

I.e., what if we make a ''Partial Lipschitz'' assumption that the Lipschitz inequality holds for $\theta_{1}, \theta_{2}$ in a large subset $\Theta^{\prime} \subset \Theta$, which covers at least (say) $98 \%$ of the measure of $\Theta$. Outside $\Theta^{\prime}$, we do not guarantee a uniform Lipschitz bound.

The interpretation here is that:

\begin{itemize}
  \item With probability at least 0.98 , parameter pairs $\left(\theta_{1}, \theta_{2}\right)$ satisfy the Lipschitz property.
  \item There might exist ''rare'' or ''exceptional'' parameter sets (2\% measure) where the local cost function changes more abruptly or is less well-behaved.
\end{itemize}

Our Scale Matching argument above, under the full Lipschitz assumption, averred that:

\begin{itemize}
  \item If the environment's cost function is L-Lipschitz in parameter space (under $\|\cdot\|$ ) for all $\theta$, then small parameter shifts never produce large jumps in cost, and the measure-dependent operator in the same ground metric $\omega$ ensures local distribution shifts reflect that scale consistently.
\end{itemize}

Now, under the Partial Lipschitz assumption:

\begin{itemize}
  \item For $98 \%$ of the parameter pairs $\left(\theta_{1}, \theta_{2}\right) \in \Theta^{\prime}$, we still get

$$
\left|W_{2}\left(q, p\left(\theta_{1}\right)\right)-W_{2}\left(q, p\left(\theta_{2}\right)\right)\right| \leq L\left\|\theta_{1}-\theta_{2}\right\| .
$$

  \item For $2 \%$ of $\left(\theta_{1}, \theta_{2}\right)$, we do not have that guarantee. Possibly large or abrupt changes in cost can occur.
\end{itemize}

Within $\Theta^{\prime}$ (the $98 \%$ region), all the scale matching arguments from the proof sketch for the Wasserstein Lipschitz ActPC Property hold.   Specifically, small parameter shifts $\left\|\theta_{1}-\theta_{2}\right\|$ produce cost changes $\leq L\left\|\theta_{1}-\theta_{2}\right\|$. The measure-dependent geometry in $\omega$-space is still consistent with these scaled changes.

Outside $\Theta^{\prime}$ (the $2 \%$ region), we cannot guarantee that cost changes are bounded linearly by $\left\|\theta_{1}-\theta_{2}\right\|$. Therefore, the scale matching property might fail or be drastically weaker there. In practice:

\begin{itemize}
  \item Most of the agent's local steps happen in regions of parameter space it visits. If it rarely visits the $2 \%$ exceptional region, it rarely experiences abrupt or ''non-Lipschitz'' cost changes.
\item However, if the agent does wander into the $2 \%$ region, then scale matching can break, and large cost changes might happen from small param moves (or vice versa).
\end{itemize}

Hence, we can say: the Wasserstein Lipschitz ActPC Property holds ''almost everywhere'' (i.e. with probability at least 0.98 ) in parameter space. For the majority of local updates in $\Theta^{\prime}$, the environment's distribution mismatch changes at scale $\delta$ implies the same scale $\delta$ in the agent's measure-dependent operator. In the small leftover region $\Theta \backslash \Theta^{\prime}$, the argument no longer strictly applies.

Basically, the rare exceptions to Lipschitz continuity might cause spikes or anomalies in cost changes, but they do not destroy the overall synergy between external cost and internal geometry.

If the agent's measure-based operator is stable, even if it hits the $2 \%$ region, it might detect a large mismatch or gradient and quickly move back into the well-behaved region $\Theta^{\prime}$.  Over time, the agent might ''learn'' to stay out of the poorly behaved region if it typically yields poor cost.

In Appendix \ref{app:prob-lip} we explore a more rigorous formulation of the ''failing Lipschitz condition 2\% of the time'' situation.   This is interesting mostly because it's instructive how one can model the situation using Markov chains.

\subsection{Transfer of Convex-Like Properties from Outer to Inner Loop} \label{sec:convex}

Moving on, we next argue that, if the Wasserstein metric is used both for the outer loop (assessing system-level prediction error) and the inner loop (assessing internal errors for the purpose of information-geometric guidance of system learning), then: if the outer loop present a convex or almost-convex optimization problem, the inner loop should be similarly faced with a convex or almost-convex optimization problem.

To formalize this, consider as above parametric family of reward-distributions $\{p(r \mid \theta)\}_{\theta \in \Theta}$ and an environment's distribution $q(r)$.   Assume the agent's global objective is

$$
\min _{\theta \in \Theta} W_{2}(q, p(r \mid \theta))
$$

\noindent rather than a KL-based mismatch.   And assume that for internal parameter updates, the agent employs a measure-dependent Laplacian $\hat{L}(p)$ or $\hat{L}(p)^{\dagger}$ derived from the same ground metric $\omega(\mathbf{x}, \mathbf{y})$, shaping local gradient flow in distribution space.

Consider then the situation where the environment's mismatch function in distribution space (i.e., $W_{2}(q, p(r \mid \theta))$ ) is  unimodal or ''convex-like'' over $\theta$ (in the sense that small local moves in distribution space yield monotonic cost decrease toward a single global optimum).

Then, we argue for a Wasserstein Quasi-Convex ActPC Property, of the rough form: {\it The agent's local gradient flow in parameter space, guided by the internal Wasserstein-based geometry, will under these assumptions exhibit quasi-convex or ''straight-line'' geodesics for distribution shifts.}

More formally:

\begin{enumerate}
  \item If the environment's mismatch $\Delta_{\text {env }}(\theta)=W_{2}(q, p(r \mid \theta))$ has no local minima apart from the global optimum, then an iterative transport-based gradient method in parameter space also has no spurious local minima for small steps, provided local distribution changes reflect the same metric $\omega$.
  \item Under unimodality or ''smoothness'' conditions, local updates in the agent's measure-dependent operator approximate geodesics in distribution space that lead ''directly'' to the environment optimum. This yields ''convex-like'' or quasi-convex convergence in practice.
\end{enumerate}

Of course, these propositions are somewhat idealized; in real ActPC-Geom systems, approximate inverses, partial rank factorization, or domain complexities might prevent perfect alignment or strict convexity. Still, the theorems highlight that using the same ground metric for external cost and internal geometry can yield improved local alignment, potentially leading to more stable or efficient gradient-based (or ActPC-based) solutions.

In Appendix \ref{app:convex} we give a proof sketch for the Wasserstein Quasi-Convex ActPC Property, and also for a variation which holds if the convexity is imperfect and marred by small bumps.   The basic conclusion appears to still hold in the latter case: if the system's overall reward landscape is almost convex but has small local minima, then the same is true of the internal information-geometric landscape ... if one uses Wasserstein in both the inner and outer loops.

\subsection{Practical Challenges of Scalable Wasserstein Predictive Error Computation}

The conceptual and mathematical benefits of using the Wasserstein metric in both the outer and inner loops of an ActPC-Geom system seem clear.   What is less obvious is the computational tractability.   Using the KL divergence in the outer loop, as done in classic ActPC implementations, at least has the benefit of computational simplicity.

In terms of naive computational complexity, one may compare that

\begin{itemize}
  \item KL Divergence between distributions $p$ and $q$ can often be computed in $\mathcal{O}(n)$ if both distributions are represented discretely with size $n$ (simply summing $p(i) \log [p(i) / q(i)]$ over $i=1 . . n)$.
  \item Wasserstein Distance $\left(W_{2}\right)$ or the optimal transport problem typically requires $\mathcal{O}\left(n^{3}\right)$ in naive linear programming or $\mathcal{O}\left(n^{2} \log n\right)$ for sophisticated entropically-regularized versions, or partial-Sinkhorn methods. Even sample-based methods can be fairly heavy if high accuracy is desired.
\end{itemize}

\noindent That is: out of the box, computing $W_{2}(p, q)$ is significantly more expensive than computing KL for the same distribution representations.

We believe, however, that with a bit of effort, approximations to the Wasserstein metric can be put into place in a way that makes its use in the outer loop of ActPC-Geom entirely feasible.   Indeed we have already done most of the work needed to establish this in Section \ref{sec:wasser} above in the context of the use of Wasserstein in the inner loop.

The key point we focused on in the earlier discussion was the use of neural approximators acting on embedding spaces (produced e.g. by approximations to kPCA with appropriate kernel) to accelerate the core distributional calculations underlying the Wasserstein metric.  These mechanisms apply equally well to the outer as to the inner loop.   And we will extend and upgrade these mechanisms further in Section \ref{sec:hyper} below, using hypervectors.

There are also a few other important conceptual points one may make in this context:

\begin{enumerate}
  \item {\bf Incremental Updating}: It will often be desirable to reuse partial solutions across small changes in $\theta$.
  \begin{itemize}
  \item This exploit the fact that in almost all real situations, distributions $\theta \mapsto p$ do not shift drastically each micro-step.
  \item In an apparatus involving neural approximators and embedding vectors, this means doing continual, rapid online updates of the kPCA approximations underlying the embedding vectors, and of the neural approximators themselves
\end{itemize}
\item {\bf Attentional Focusing}: In real systems, the agent might only care about a subset of reward or environment states at a time (the ''hot'' region, or the Wasserstein-relevant ''attentional focus'')
\begin{itemize}
\item This may mean, when doing any particular prediction error calculation, ignoring some portions of the state that appear not relevant according to a relevant {\it attention mechanism}
\item This may leverage attention mechanisms associated with the non-information-geometry aspects of the ActPC system in question, e.g.  ECAN for an ActPC-Chem system running in a Hyperon instance, or the intrinsic attention mechanism of a transformer-like neural network
\item Aspects of the reward or state outside the attentional focus may be outright ignored in making the Wasserstein calculations, or instead they may simply be used in somewhat ''stale'' form and not updated as aggressively via online learning
\end{itemize}
\end{enumerate}

While we are surfacing these strategies here in the context of the outer loop, they are applicable and potentially highly relevant to the inner loop as well.

Of course, even with all these clever strategies, the approximations needed to meaningfully approximate the Wasserstein metric will still incur non-trivial overhead.   However, this must be balanced against the faster learning that can be achieved via  the synergy between W-based cost externally and measure-based transport internally, which has potentially to drastically reduce the number of evaluations of the metric that has to be done, compared to more naive and less semantically satisfactory approaches to predictive error assessment.

Further, software and hardware technologies for rapidly evaluating multilayer neural nets are now extremely sophisticated and evolving rapidly.   If implemented with GPU kernels using modern libraries, speeds for neural approximators acting on well designed embeddings may come out quite close to KL-level speeds in practice.

\part{Cognitive Enhancements: Compositionality, Associative Long-Term Memory and Neural-Symbolic Integration}

\section{Compositional Hypervector Embedding}  \label{sec:hyper} 

In this section we present a series of explorations aimed at synergizing the ''basic'' ActPC-Geom methods described above with other more ''cognitively'' oriented AI structures and techniques.   These integrations are not modular connections between black boxes, but rather tighter integrations in which other AI methods are used to make internal ActPC-Geom methods work better, or ActPC-Geom and other methods are used together to shape networks that reflect both of their principles.

We begin our adventures in cognitive ActPC-Geom integrations here with an exploration of potential use of hypervectors with compositional algebra to increase both efficiency and inferential generalization capability of ActPC-Geom networks.

In the above we have discussed using kernel PCA embedding of states in a discrete ActPC network, so as to give a lower dimensional target for a neural approximator to the inverse measure-dependent Laplacian.   A natural extension of this idea is to look at hypervector embeddings that combine  kernel PCA like properties with other desirable characteristics.  For instance, one may wish to do hypervector embeddings that act like kernel PCA but also embody a compositional algebra, i.e. representing entities as algebraic combinations of their properties.   It seems this could potentially help a purely neural or hybrid ActPC network develop more robust, compositional internal representations. 

We start by outlining the general idea of hypervector embeddings and then connect it to (1) kernel PCA-like metric embedding, (2) compositional representation, and (3) the synergy with hybrid (discrete/continuous) ActPC and information-geometry-driven learning.

\subsection{Some Basic Concepts of Hypervector Embedding}

Hyperdimensional computing or hypervector embedding typically involves representing entities as very high-dimensional vectors (e.g. thousands of dimensions) over $\{ \pm 1\}$ or integer or real values. Crucially, these vectors can be combined with well-defined algebraic operations such as:

\begin{enumerate}
  \item Binding: Often a componentwise multiplication or XOR.
  \item Bundling: Often a componentwise addition or majority vote.
  \item Permutation or rotation: Shifts or permutes vector indices to represent certain structural transformations.
\end{enumerate}

Placing these operations on hypervectors yields a compositional algebra, which can be used for multiple purposes, including e.g. flexible property/value representations, e.g. , we can represent ''an entity that has property A and property B'' by binding a vector for ''entity'' with vectors for each property and bundling them. Then we can do partial matches or unbindings to retrieve sub-parts of the representation.  Javier Snaider's work on Modular Compositional Representation \cite{snaider2014modular}, which we'll review below, comprises one interesting way to do this, which was integrated in a version of Stan Franklin's LIDA cognitive architecture \cite{snaider2014vector}.

We will argue below that it's possible to make hypervector embeddings that incorporate kernel-like similarity measures in high-dimensional spaces, giving them the effect of approximate isometries or ''random feature expansions.''    Putting this together appropriately with methods drawn from Snaider's work, we have a potential route to constructing an embedding that is:

\begin{enumerate}
  \item High dimensional enough to store compositional structure.
  \item Capable of capturing approximate ''kernel-based'' distances for discrete or continuous objects.
  \item Able to manifest interesting synergies between these ''kernel based'' distances and other semantics attached to operations like binding, bundling and permutation/rotation
\end{enumerate}

This sort of embedding, we argue, has potential to serve as a highly effective bridge between neural and symbolic.  By using this sort of embedding vector to approximate the state of a neural network or a neural-symbolic system, and using these embeddings to guide neural networks approximating the geometry of neural or neural-symbolic systems with a goal of guiding their dynamics ... one is using the symbolic to accelerate and enhance the neural at the ''internal plumbing'' level rather than via having a separate symbolic system interact with a separate neural system.   Of course one can have the latter as well, but it seems there is some special value to be obtained by weaving the symbolic and the neural together more tightly.

\subsection{Snaider's Compositional Hypervector Algebra}

We now give a brief conceptual and mathematical summary of Javier Snaider's concept of ''modular composite representation'' within hyperdimensional computing, focusing on how binding, bundling, and permutation operations on hypervectors are defined and used to represent composite meanings. While Snaider's work overlaps with broader hyperdimensional computing (HDC) traditions (e.g., holographic reduced representations, HRR; vector symbolic architectures, VSA; etc.), modular composite representation has some specific points of emphasis on the semantics and the modular nature of large vectors.

\paragraph{Snaider's Modular Composite Representation}  Snaider has emphasized a ''modular'' approach to building hypervectors: each compositional structure is a combination of sub-vectors that remain ''locally interpretable'' within the global hypervector. By carefully defining how binding and bundling share sub-blocks (or modulo-indexing) of the vector, one can maintain a sense of modularity in the overall composite representation.

Snaider's framework for composite representation gives particular meaning to the three core operations most commonly. used in HDC applications: binding, bundling and permutation.

\paragraph{Binding}  Binding associates two (or more) hypervectors together so that they become a single combined entity, while retaining the possibility of ''unbinding'' or partial matching. For instance, if we want to represent ''Property A has Value X,'' we bind $\mathbf{A}$ and $\mathbf{X}$.

Common implementations of binding are:

\begin{itemize}
  \item Componentwise multiplication in $\{ \pm 1\}^{D}$. If $\mathbf{a}$ and $\mathbf{b}$ are both length- $D$ hypervectors with $\pm 1$ components, then the binding $\mathbf{c}=\mathbf{a} \odot \mathbf{b}$ is also length- $D$, where $c_{i}=a_{i} \times b_{i}$. This preserves near-orthogonality for distinct pairs.
  \item XOR in binary $\{0,1\}$ or $\{ \pm 1\}$. Some HDC frameworks treat binding as bitwise XOR. The principle is analogous: the result is close to random unless you have the same vector to ''un-bind.''
\end{itemize}

The semantic meaning here is relatively simple:

\begin{itemize}
  \item ''(A, X) is one composite item.'' The result is a unique vector that can be recognized only if we ''rebind'' with $\mathbf{A}^{-1}$ (which is $\mathbf{A}$ itself if we use $\pm 1$ multiplication) to retrieve $\mathbf{X}$.
  \item Binding is the mechanism for linking roles and fillers in a compositional structure: e.g., ''role = property, filler = value.''
\end{itemize}

Mathematically, if $\mathbf{a}$ and $\mathbf{b}$ have $\pm 1$ components, then their binding looks like

$$
\mathbf{c}=\mathbf{a} \odot \mathbf{b}, \quad c_{i}=a_{i} b_{i} .
$$

\noindent To invert, we do $\mathbf{c} \odot \mathbf{a} \rightarrow \mathbf{b}$, because $( \pm 1) \times( \pm 1)$ is invertible.

In Snaider's modular approach, sometimes the binding might be done per block or modulo segment, so sub-blocks represent different properties while still using multiplicative or XOR binding within each sub-block.

\paragraph{Bundling} Bundling aggregates multiple hypervectors into a single  vector -- roughly like a "set union" or "superposition"   If we want to combine ''Feature1,'' ''Feature2,'' ... into one representation, we ''bundle'' them.

This is generally implemented via elementwise addition or majority vote: If $\mathbf{x}, \mathbf{y}$ are length- $D$, their bundling $\mathbf{z}$ could be

$$
z_{i}=\operatorname{sign}\left(x_{i}+y_{i}\right)
$$

\noindent or simply $\mathbf{z}=\mathbf{x}+\mathbf{y}$ if we keep real entries. Then a final sign or normalization is done. Another variant is to store the sum unthresholded.

 The result of bundling is a ''superposition'' of the input vectors, so partial overlap can be discovered by dot product. If we do sign or majority, the final vector is again in $\pm 1\}^{D}$.

 In Snaider's modular approach, the bundling might occur in segmented or partially overlapping blocks, giving separate sub-blocks for different roles.

\paragraph{Permutation}  Permutation shifts or reindexes a hypervector to represent a structural or positional change. For instance, if we represent a sequence $\mathbf{x}_{1}, \mathbf{x}_{2}$, we can store them by successively permuting some base vector, or if we want ''Slot1:Value1, Slot2:Value2,'' we do a random/circular permutation for each slot so it's distinct.

The basic semantic meaning here is: ''This is the same vector, but placed in a different role or position.'' So the same base item, once permuted, indicates ''the item in slot \#2'' or ''the item used in a next step of the sequence.''

This may be implemented as follows, for instance:
--
\begin{itemize}
  \item Circular shift or ''index permutation'' of a length- $D$ vector. For a circular shift by 1 , if $\mathbf{x}=$ $\left(x_{1}, x_{2}, \ldots, x_{D}\right)$, then $\pi(\mathbf{x})$ might be $\left(x_{D}, x_{1}, \ldots, x_{D-1}\right)$.
  \item Random permutation: Another approach is a random shuffle based on a ''slot ID.'' If we have a stable random permutation $P$, then applying $P$ to $\mathbf{x}$ yields a new vector $\mathbf{x}^{\prime}$.
  
  \end{itemize}
  
  \noindent Typically, repeated permutations remain near-orthogonal to each other, which helps keep distinct roles separate.

Some simple mathematical observations are:

\begin{itemize}
  \item If $\sigma$ is a permutation on $\{1, \ldots, D\}$, then $\pi(\mathbf{x})_{i}=x_{\sigma(i)}$. In a circular shift, $\sigma(i)=i-1 \bmod D$.
  \item Permutation is invertible (applying $\sigma^{-1}$ retrieves the original vector). In Snaider's modular scheme, permutations might be done in sub-blocks or some structured approach to keep roles distinct.
\end{itemize}

\subsubsection{Semantic Meaning of a ''Modular Composite Representation''}

Snaider's MCR involves putting binding, bundling, and permutation together as follows:

\begin{enumerate}
  \item Binding: Binds pairs or n-tuples of roles/fillers into single vectors so that each pair is invertible (like ''Property => Value'').
  \item Bundling: Aggregates multiple items or sub-structures into a single superposed vector (like ''set membership'' or ''co-occurring features'').
  \item Permutation: Places items in distinct roles or positions (like indexing an array or shifting in a sequence).
\end{enumerate}

In the MCR approach, each sub-block (or modulo partition) of the hypervector might represent different roles or different aspects of the representation, so one can do partial bundling/binding in each sub-block. This ensures sub-structures remain ''locally decodable,'' which is helpful for large compositional reasoning. The composition is ''modular'' because we can segment or distribute different semantic parts into designated zones, yet still unify them all in a single big vector. Hence, a modular composite representation allows large structures to be built in a single hypervector, with partial invertibility for roles, sets, and sequences, while remaining robustly noise-tolerant thanks to high dimensionality.

\paragraph{Example Calculational Details} To clarify the concepts, we give a simplified example of how each operation might proceed numerically in $\{ \pm 1\}^{D}$ :

\begin{enumerate}
  \item Binding: $\mathbf{c}=\mathbf{a} \odot \mathbf{b}, c_{i}=a_{i} \times b_{i}$.
  \item Bundling: $\mathbf{z}=\operatorname{sign}(\mathbf{a}+\mathbf{b}+\ldots)$, or a majority vote.
  \item Permutation: $\mathbf{p}=\pi(\mathbf{x}), p_{i}=x_{\sigma(i)}$.
\end{enumerate}

In Snaider's ''modular'' approach, one might define a dimension $D=m \times n$. Each property or sub-slot uses a block of $m$ components, so binding might occur only within that block, or across blocks in a controlled manner. For example, ''feature \#j in sub-block \#j.'' Summation-based bundling across sub-blocks can produce a single global representation, but each sub-block can still be partially extracted or unbound.

\subsubsection{A Commonsense Reasoning Example}  Taking this a step further, let's walk through how to encode some everyday entities (''apples,'' ''bananas'') and their features (''color,'' ''shape,'' etc.), and how we might store multiple items or a short sequence. This is a didactic example rather than an attempt to recount how things would work in an actual large-scale system, but it conveys the essence of Snaider's approach.

\paragraph{Base Vectors for Roles (Features)}   Let's say we define a few ''roles'' (or properties) we commonly assign to everyday objects, each with a base hypervector:

\begin{itemize}
  \item $\mathbf{v}_{\text {color }}$ in $\{ \pm 1\}^{D}$,
  \item $\mathbf{v}_{\text {shape }}$ in $\{ \pm 1\}^{D}$,
  \item $\mathbf{v}_{\text {edibility }}$ in $\{ \pm 1\}^{D}$.
\end{itemize}

\noindent (We might have more, like ''location,'' ''typical usage,'' etc. Each is a random vector of large dimension $D$ .)

\paragraph{Base Vectors for Feature Values}.  We also define base vectors for different values each role can take:

\begin{itemize}
  \item For color: $\mathbf{v}_{\text {red }}, \mathbf{v}_{\text {yellow }}, \mathbf{v}_{\text {green }}, \ldots$..
  \item For shape: $\mathbf{v}_{\text {round }}, \mathbf{v}_{\text {long }}, \ldots$.
  \item For edibility: $\mathbf{v}_{\text {edible }}, \mathbf{v}_{\text {inedible }}$.
\end{itemize}

\noindent Again, each is a random hypervector in $\{ \pm 1\}^{D}$. Because they're high-dimensional and randomly generated, they're pairwise near-orthogonal.

\paragraph{Base Vectors for Entities} We can also define a named base vector for each entity:

\begin{itemize}
  \item $\mathbf{v}_{\text {apple }}, \mathbf{v}_{\text {banana }}, \ldots$.
\end{itemize}

The purpose here is: We want to form a composite representation of, for example, ''An apple is red, round, and edible.'' Then we can store multiple items, possibly bundling them into a single ''memory state.''

\paragraph{Binding (Role + Value)}. In Snaider's style, we often bind the role and the value, e.g.:

$$
\mathbf{v}_{\text {Color }=\text { Red }}=\mathbf{v}_{\text {color }} \odot \mathbf{v}_{\text {red }}
$$

\noindent where '' $\odot$ '' denotes componentwise multiplication in $\pm 1$ domain (or XOR in a binary domain). The result is a unique vector representing ''color=red.'' We can invert or ''un-bind'' by multiplying again with $\mathbf{v}_{\text {color }}$ if we want to retrieve ''red,'' or with $\mathbf{v}_{\text {red }}$ if we want to retrieve ''color.''

We can do similarly  for Shape and Edibility:

\begin{itemize}
  \item Shape=Round: $\mathbf{v}_{\text {Shape }}=$ Round $=\mathbf{v}_{\text {shape }} \odot \mathbf{v}_{\text {round }}$
  \item Edibility=Edible: $\mathbf{v}_{\text {Edibility }=\text { Edible }}=\mathbf{v}_{\text {edibility }} \odot \mathbf{v}_{\text {edible }}$
\end{itemize}

\noindent If the entity is ''banana => color=yellow => shape=long => edibility=edible,'' we form the analogous bound vectors.

\paragraph{Binding an Entity with Its (Role=Value) Pairs}  We might also bind each (Role=Value) vector with the entity base vector $\mathbf{v}_{\text {apple }}$ to create a single combined item, e.g.:

$$
\mathbf{v}_{(\text {Apple,Color }=\text { Red })}=\mathbf{v}_{\text {apple }} \odot\left(\mathbf{v}_{\text {color }} \odot \mathbf{v}_{\text {red }}\right)
$$

and so on for shape, edibility, etc.   We can do this individually or skip directly to bundling them, as shown below.

\paragraph{Bundling (Combining Multiple Pairs into One Representation)} To store ''An apple is red, round, and edible'' in one hypervector, we typically bundle the different ''(role=value)'' bound vectors:

$$
\mathbf{v}_{\text {appleComposite }}=\operatorname{Bundle}\left(\mathbf{v}_{\text {(Apple,Color=Red) }}, \mathbf{v}_{(\text {Apple,Shape=Round })}, \mathbf{v}_{(\text {Apple,Edibility=Edible })}\right)
$$

 If we use addition followed by sign, we do something like
 
$$
\mathbf{v}_{\text {appleComposite }}=
$$
$$
\operatorname{sign}\left(\mathbf{v}_{\text {apple }} \odot\left(\mathbf{v}_{\text {color }} \odot \mathbf{v}_{\text {red }}\right)+\mathbf{v}_{\text {apple }} \odot\left(\mathbf{v}_{\text {shape }} \odot \mathbf{v}_{\text {round }}\right)+\mathbf{v}_{\text {apple }} \odot\left(\mathbf{v}_{\text {edibility }} \odot \mathbf{v}_{\text {edible }}\right)\right)
$$

Thus, $\mathbf{v}_{\text {appleComposite }}$ is a single high-dimensional vector storing that the entity ''apple'' has:

\begin{enumerate}
  \item Color=Red,
  \item Shape=Round,
  \item Edibility=Edible
  \end{enumerate}
  
\noindent all superimposed. 

In typical hyperdimensional settings, we can do partial unbinding or dot-product matches to check which roles/values are present.

\paragraph{Permutation (Positional or Sequence Info)} Now suppose we have two entities in a sequence.   We might want to say: ''In grocery store order: first is Apple, second is Banana.'' If we define a random or circular permutation operator $\pi_{1}$ to represent ''Slot \#1,'' and $\pi_{2}$ to represent ''Slot \#2,'' we can store them:

$$
\mathbf{v}_{\text {slot1Apple }}=\pi_{1}\left(\mathbf{v}_{\text {appleComposite }}\right) \quad, \quad \mathbf{v}_{\text {slot2Banana }}=\pi_{2}\left(\mathbf{v}_{\text {bananaComposite }}\right)
$$

Then we bundle them:

$$
\mathbf{v}_{\text {2ItemGroceryList }}=\operatorname{Bundle}\left(\mathbf{v}_{\text {slot1Apple }}, \mathbf{v}_{\text {slot2Banana }}\right) .
$$

This single hypervector now encodes ''Slot 1 => apple(with color=red, shape=round, etc.), Slot 2 => banana(with color=yellow, shape=long, etc.).''

\paragraph{Retrieval (Partial Unbinding / Dot-Product)}  From $\mathbf{v}_{\text {appleComposite, }}$ if we multiply by $\mathbf{v}_{\text {apple }}$ we get something like:

$$
\mathbf{v}_{\text {appleComposite }} \odot \mathbf{v}_{\text {apple }} \approx \operatorname{Bundle}\left(\mathbf{v}_{\text {color }} \odot \mathbf{v}_{\text {red }}, \mathbf{v}_{\text {shape }} \odot \mathbf{v}_{\text {round }}, \mathbf{v}_{\text {edibility }} \odot \mathbf{v}_{\text {edible }}\right) .
$$

\noindent Then we might do a dot product with each known ''role $\odot$ value'' to see which are present. Because of the near-orthogonality properties, each bound pair that's included typically yields a higher dot product. That is how we discover that color=red, shape=round, edibility=edible.

Similarly, from the 2ItemGroceryList vector, we can unpermute with $\pi_{1}^{-1}$ or $\pi_{2}^{-1}$ to discover which item was in slot 1 or 2.

\subsubsection{Summing up}

Overall, as we have just illustrated in a simple example, in Snaider's approach to modular composite representation, we:

\begin{enumerate}
  \item Bind each property's base vector with its value base vector to get ''(property=value).''
  \item Optionally bind that with the entity vector, or we can keep entity base vectors separate.
  \item Bundle multiple property-value pairs into a single composite hypervector for that entity.
  \item If we want sequences or positional roles, we apply permutation to shift vectors, then bundle them.
\end{enumerate}

\noindent All of these operations remain in the $\{ \pm 1\}^{D}$ space (or an integer hypervector space), are noise-tolerant due to high dimensionality, and are partially invertible via unbinding. 

This machinery provides a way to represent everyday knowledge about ''apples are red, round, edible,'' or ''the first item is an apple, second is a banana,'' in a single large hypervector while preserving compositional structure.   It can also be used to represent all sorts of other knowledge, e.g. patterns in states of ActPC networks, in ways that leverage and expose their compositional structure.

\subsection{Combining Compositional Hypervector Algebra with a Kernel-PCA-Like Metric}

Now we connect these MCR hypervectors with ActPC-Geom by explaining how they can be used as an extended, semantically improved version of the approximate kPCA vectors we proposed above as compressed output for the neural approximators of the ActPC information manifold's inverse measure-dependent Laplacian.

We give here a conceptual outline of an algorithmic approach for extending a set of $k$-dimensional kernel-PCA vectors (which we can call ''base vectors'' or ''basis vectors'') into high-dimensional hypervectors suitable for Snaider-style MCR (Modular Composite Representation) -- while retaining approximate orthogonality from kPCA, and ensuring a fixed set of ''core ontology concepts'' is well-treated by both the MCR hypervector operations and the original kPCA decomposition. This outline can be adapted or refined as needed, but it sketches a practical path to bridging the two representations.   

(The importance and origin of the ''core ontology concepts'' is deferred to Section \ref{sec:ontology}; for now let us simply consider them an optional, potentially valuable input to the algorithm.)

\paragraph{Starting Point: kPCA Vectors}  Let us assume that as a starting-point we have

$$
\left\{\mathbf{u}_{1}, \mathbf{u}_{2}, \ldots, \mathbf{u}_{k}\right\}
$$

\noindent as the main kPCA basis (each $\mathbf{u}_{i} \in \mathbb{R}^{k}$ or $\mathbb{R}^{m}$, depending on how we define it). Alternatively or complementarily, we may have $\mathbf{N}$ entity-specific vectors in some dimension $k$, each capturing the entity's main features from kernel PCA.

Each vector $\mathbf{u}_{i}$ has the orthogonality or near-orthogonality property from kPCA in the space of interest. We want to embed these into a large dimension $D \gg k$, such that linear or near-linear relationships in the original kPCA space are (mostly) preserved.

\subsubsection{Overall Strategy: Two-Block or ''Hybrid'' Embedding} A common approach in forming purpose-specific hypervector embeddings is to build a two-block (or multi-block) structure in a high-dimensional hypervector:

$$
\mathbf{h}_{i}=\left[\operatorname{scaledEmbed}\left(\mathbf{u}_{i}\right) \mid \mathbf{r}_{i}\right] .
$$

We divide the hypervector dimension $D$ into:

\begin{enumerate}
  \item A sub-block (size $\sim k$ or slightly bigger) that encodes the original kPCA vector in some normalized or scaled way,
  \item A random-lift sub-block (size $D-k$ ) that provides near-orthogonal expansions plus a ''unique signature'' for Snaider's MCR operations (binding/bundling/permutation).
\end{enumerate}

\paragraph{The ''ScaledEmbed $\left(\mathbf{u}_{i}\right)$ '' Sub-Block} Next,
\begin{itemize}
  \item We simply copy $\mathbf{u}_{i}$ (or a scaled version) into a sub-block of the dimension $D$.
  \item Optionally, we do a small random orthonormal transform to expand from dimension $k$ to dimension $\ell \geq k$, but not too large.
  \item Then we place that $\ell$-dim vector in a sub-block of the final dimension $D$.
  \item This preserves, at least approximately, the relative angles or dot products from the original kPCA space, so that linear combinations or distance in that sub-block reflect the original kernel PCA geometry.
\end{itemize}

\paragraph{The '' $\mathbf{r}_{i}$ '' Random-Lift Sub-Block}
\begin{itemize}
  \item We generate a random vector $\mathbf{r}_{i} \in\{ \pm 1\}^{D-\ell}$ or $\backslash$ mathbb $\{R\}^{\wedge}\{D$-lell $\}$ for each $\mathbf{u}_{i}$.
  \item This part ensures that each vector $\mathbf{h}_{i}$ is near-orthonormal to other vectors if $\mathbf{r}_{i}$ are chosen with large dimension and random seed, and also that each vector has a ''unique signature.''
  \item We can define separate sub-blocks for ''role vectors'' vs. ''filler vectors'' or ''core ontology concepts'' if we want MCR's ''modular'' logic.
\end{itemize}

\noindent Hence, each final hypervector $\mathbf{h}_{i} \in \mathbb{R}^{D}$ (or $\{ \pm 1\}^{D}$ ) is a concatenation (or partial overlap) of ''kPCA sub-block'' + ''random sub-block.'' This approach ensures we keep two functionalities:

\begin{enumerate}
  \item Approximate kPCA-based distance or linear combos in one sub-block,
  \item MCR compositional binding/bundling/permutation in the random sub-block.
\end{enumerate}

\subsubsection{Handling Core Ontology Concepts}  Next, for a few different reasons, it will be desirable in carrying out this sort of process to ensure a fixed set of ''core ontology'' concepts have stable, meaningful hypervectors that integrate well with both the ''kPCA side'' and the ''random-lift side.''  

One way to do this would be as follows.  

Firstly, for each core concept $c$, if the concept is known in the original kernel PCA space, we have a vector $\mathbf{u}_{c}$. We can embed it in the ''kPCA sub-block'' portion as above.

But we can also do more.   We can also create a predefined random sub-block with random signature $\mathbf{r}_{c}$ for that concept. Possibly we could even create sub-sub-blocks for ''concept,'' ''property,'' etc., if we want more modular control.

These can be combined via, say,

$$
\mathbf{h}_{c}=\left[\operatorname{scaledEmbed}\left(\mathbf{u}_{c}\right) \mid \mathbf{r}_{c}\right]
$$

The ''random portion'' ensures that concept is robustly distinct from others in MCR operations, while the ''kPCA portion'' ensures it aligns or is near orthonormal to other relevant concepts or states in the old kernel PCA geometry.

If some core concepts are not effectively represented in the kPCA space, we can define their ''kPCA sub-block'' as e.g. zero or a small random vector, so they remain ''unmapped'' in that sub-block.

\subsubsection{Integrating Snaider-Style MCR Operations}

Now how do the MCR operations work here?   Fairly straightforwardly, it seems.

\paragraph{Binding} When we bind two hypervectors $\mathbf{h}_{A}, \mathbf{h}_{B}$ in $\{ \pm 1\}^{D}$ for compositional semantics:

\begin{enumerate}
  \item In the ''kPCA sub-block,'' we might define the binding as just componentwise multiplication or XOR. This typically scrambles the interpretability of the kPCA portion somewhat.
  \item In the ''random-lift sub-block,'' we also do componentwise multiplication, which is standard for MCR. This sub-block is where most of the compositional logic is robustly executed.
  \item If we want to preserve the ''kPCA geometry'' in the final result, we might consider only doing* binding in certain sub-regions**. For instance, the ''kPCA sub-block'' might remain unaltered if the logic is to keep a stable projection. Meanwhile, the ''random sub-block'' is used for the actual binding operation. This yields a partially ''modular'' approach.
\end{enumerate}

Hence, the combined vector includes partial ''unchanged kPCA sub-block'' plus partial ''bound random sub-block,'' ensuring we can still do partial linear combos with the unaltered sub-block while we handle composition in the random sub-block.

\paragraph{Bundling} Bundling multiple hypervectors (like ''this entity has features $X, Y, Z$ '') is typically elementwise addition or majority vote, so here:

\begin{itemize}
  \item We can do it across the entire vector, but again it might overshadow the kPCA portion if they differ.
  \item Alternatively, we can do bundling only in the random sub-block and keep the kPCA sub-block as some weighted average or remain separate.
  \item Each approach is a design choice depending on how strongly we want the ''kPCA geometry'' to remain interpretable in the composite vector.
\end{itemize}

\paragraph{Permutation}  For representing ''slot \#1,'' ''slot \#2,'' etc., we define a random or circular shift operator that we only apply to the random-lift sub-block, leaving the kPCA sub-block in place. That ensures the ''positional role assignment'' is captured in the random portion, while the kPCA portion remains the same so the system can still see the original vector's geometry.

\subsubsection{A Step-by-Step ''Vector Expansion'' Algorithm}  Incorporating these ideas, we now outline an algorithm for producing the final hypervectors from kPCA vectors:

\begin{enumerate}
  \item Given dimension $k$ from kernel PCA, and target hypervector dimension $D$.
  \item Partition dimension $D$ as $(\ell+R)$. Let $\ell \geq k$. The first $\ell$ components store the ''kPCA sub-block,'' the remaining $R$ are the ''random-lift sub-block.''
  \item Insert sub-blocks corresponding to "core ontology" concepts as needed
  \item Construct Orthonormal Extension:
\begin{itemize}
  \item (Optional) If $\ell>k$, define a small random orthonormal matrix $\mathbf{Q} \in \mathbb{R}^{k \times \ell}$ so each $\mathbf{u} \in \mathbb{R}^{k}$ is mapped to $\mathbf{u}^{\prime}=\mathbf{u} \cdot \mathbf{Q} \in \mathbb{R}^{\ell}$. Then place that $\mathbf{u}^{\prime}$ in the first $\ell$-dim portion.
  \item Otherwise, if $\ell=k$, we directly copy $\mathbf{u}$ into sub-block \#1.
\end{itemize}
\end{enumerate}

\paragraph{Management of Random Signatures:}  For each concept or entity, we can define a random signature $\mathbf{r} \in\{ \pm 1\}^{R}$, and potentially store these in a dictionary.   For ''core ontology'' concepts, we can fix these signatures to remain consistent across expansions.   These signatures can then be concatenated:

$$
\mathbf{h}(\mathbf{u})=\left[\mathbf{u}^{\prime} ; \mathbf{r}\right]
$$

\noindent where $\mathbf{u}^{\prime}$ is dimension $\ell$ and $\mathbf{r}$ dimension $R$.  (Possibly we might want to normalize or threshold to $\{ \pm 1\}^{D}$, as well.)

\subsubsection{Summary}

There are multiple moving parts here and a number of micro level decisions to be made based on experimental practice.   However, a sensible high level approach seems clear:

\begin{itemize}
  \item Partition the final hypervector dimension.
  \item Embed the original kPCA vectors in the first sub-block (sometimes with random orthonormal extension)
  \item Embed core ontology concepts in their own sub-blocks as desired
  \item Random-lift a second sub-block for robust compositional MCR operations.
  \item Design the binding/bundling/permutation operations so they either (a) only act on the random sub-block or (b) carefully incorporate the kPCA sub-block in a reversible manner, depending on how strongly we want to preserve the original linear geometry.
\end{itemize}

With that design, as elaborated above, we get both:

\begin{enumerate}
  \item The near-orthogonality and interpretability from kernel PCA in a sub-block
  \item The compositional Snaider-like MCR logic from the random sub-block
  \item Effective incorporation of concepts from a given core ontology as needed
\end{enumerate}

This synergetic approach addresses the desire to keep the ''kPCA orthogonality'' while enabling hypervector operations like binding, bundling, and permutation for a ''modular composite representation'' that accommodates both normal linear combos and advanced compositional manipulations.

We then obtain a symbolic-subsymbolic bridge of a fairly unique nature: A space of entities (hypervector) obeying a simple, sensible symbolic algebra, but also serving as an approximation of the state of a subsymbolic network, designed specifically to be useful in training neural approximators to this subsymbolic network's overall dynamics in a way that enables acceleration of the subsymbolic network's operations via geometric methods.

\section{Neural Architecture for Learning Fuzzy FCA Lattices from ActPC System States} \label{sec:ontology};
  
Now we turn to the notion of a ''core ontology'' of concepts relevant to modeling the states of an ActPC system.   We have articulated above how to create a hypervector embedding that combines statistical modeling of a collection of ActPC system states (e.g. via kPCA approximations) with faithful modeling of the logical relationships between elements of a core descriptive ontology.   But we have deferred discussion of where this ontology comes from.   We finally address this matter here.

There may be a meaningful role for a hand-coded initial ontology, and we will turn that toward the end of this section.   However, we believe that any such hand-coded ontology will be more effectively leveraged within the context of a flexible, data-driven ontology-learning framework, and the latter will be our main focus here.
  
 \subsection{Learning Fuzzy FCA Lattices} 
  
Tuning a compositional hypervector embedding to display the desired algebraic symmetries requires a concrete collection of concepts and properties to use for iteratively evaluating and improving the embedding.  For this purpose, in an ActPC setting, it seems most sensible to use a collection of concepts and properties that correlates naturally to the ActPC systems states that the hypervectors are embedding.  This leads to the idea of using machine learning methods to automatically induce concepts and properties from these sets of system states.  

One way to conceive of such concept lattices is using Fuzzy FCA (Fuzzy Formal Concept Analysis, \cite{samal2022learnfca}.  FCA (Formal Concept Analysis) typically organizes concepts and their properties into a lattice: each concept is a set of entities (in this case system states) + the maximal set of properties they share.   Fuzzy FCA extend this to encompass the case where properties can hold to some degree $\in[0,1]$. 

There is a literature describing relatively simplistic heuristics for learning Fuzzy FCA concept lattices from datasets, however in the present context we believe a neural learning approach will probably be more effective.

In the approach we envision, a fuzzy concept lattice is evaluated by how well it helps a neural approximator predict an ActPC system's behavior. Specifically, we only allow descriptions in terms of concepts and properties from the lattice as inputs to the approximator network, and then measure how accurately the network can predict the system's outcomes based on its state (where it see the state only via the concepts it belongs to and the properties it possesses). If it does well, that means the discovered concepts are indeed relevant to the system's dynamics.

As described in Section \ref{sec:hyper}, above, the process of constructing compositional hypervectors is then directed to pay attention to these concepts, and to make sure that its compositional algebra works effectively for relatively small logical combinations of these concepts.

\subsection{Potential Use of a Core Ontology}

A variation on the above approach arises if one has a ''core ontology'' of concepts one has reason to believe will be valuable for structuring the embedding space driving the system's information geometry.   In this case one can attempt a methodology such as:

\begin{itemize}
\item Expose the system, one by one, to each item from the core ontology, and create a state vector as the centroid of the states achieved in this process
\item ''Clamp'' the neural approximator to use these core-ontology concepts as part of its concept lattice.   I.e.
\begin{itemize}
\item the network has pre-wired output neurons that indicate the degree to which its input belongs to each of these concepts
\item when learning features, the network considers how well these features distinguish the core-ontology concepts, along with its learned concepts
\end{itemize}
\end{itemize}

One direction of research, for instance, would be to attempt this with a philosophically-grounded ontology like the one articulated in \cite{goertzel2024hyperseed}.

\subsection{Defining a Neural Architecture}

We now articulate in more detail what a neural architecture learning a fuzzy concept lattice in this way might look like.   Of course there are many different ways to do this, and our goal here is to articulate one initial approach to guide experimentation, rather than to attempt to limn out in detail what precise architecture may finally emerge from an experimental process.

\subsubsection{Notational Setup}

The process begins with raw material comprising state vectors 

$$
\left\{\mathbf{x}_{i}\right\} \subseteq \mathbb{R}^{k} 
$$

\noindent  for $i=1, \ldots, m$.  These may e.g. be derived from an approximate kPCA of ActPC network states.   In this case, eeach $\mathbf{x}_{i}$ is the raw embedding of system state $S_{i}$.

We may also begin with a Core Ontology: A set of pre-defined concepts $\mathcal{C}_{\text {core }}=\left\{C_{1}, \ldots, C_{r}\right\}$ that we believe are valuable for structuring the fuzzy concept lattice.

The main players in the architecture are then:

\begin{itemize}
  \item FCL: A learned set of concepts $\left\{F_{1}, \ldots, F_{n}\right\}$ plus degrees of membership $f_{j}(\mathbf{x}) \in[0,1]$. This forms the ''fuzzy concept lattice'' $\mathcal{L}$.
  \item Neural Approximator $N$ : Maps $\left\{f_{j}(\mathbf{x})\right\}$ to predicted system outcomes $\hat{y} \in \mathbb{R}^{\ell}$.
\end{itemize}

\subsubsection{Architecture Overview}
\begin{center}
\includegraphics[max width=\textwidth]{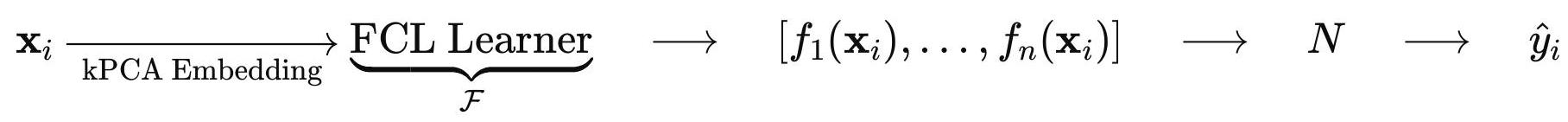}
\end{center}

\begin{itemize}
  \item FCL Learner $(\mathcal{F})$ : A neural net that outputs fuzzy membership degrees for each concept $F_{j} \in \mathcal{L}$.
  \item Approximator $N$ : A second neural net that uses these membership degrees as input to predict $\hat{y}_{i}$.
  \item Learning algorithm (Backprop or better yet PC): We do end-to-end updates. The error $\hat{y}_{i}-y_{i}$ flows back to shape both the fuzzy concept definitions and the approximator's weights.
\end{itemize}

\paragraph{FCL Learner $\mathcal{F}$} This network has

\begin{itemize}
\item Input: $\mathbf{x} \in \mathbb{R}^{k}$ (the state vector from e.g. kPCA).\\
\item Output: $\mathbf{f}(\mathbf{x}) \in[0,1]^{n}$. Each component $f_{j}(\mathbf{x})$ is the membership in concept $F_{j}$.
\end{itemize}

\noindent and rough internal architecture of a small MLP or multi-head network, where 

\begin{itemize}
\item For each concept $F_{j}$, we define an output neuron $o_{j}$ passed through a sigmoid or softplus.
  \item For core ontology concepts, we can ''clamp'' or ''pre-wire'' partial definitions. For instance, if concept $C_{\text {core }}$ must be recognized, we add a partially fixed neuron or partiallayers ensuring that membership in $C_{\text {core }}$ is learned but anchored in known cues. E.g.:
  \item A concept's neuron's weights are partially trainable, partially fixed, ensuring it consistently distinguishes that concept from data.
\end{itemize}

The key point is, this net must produce a''fuzzy concept-lattice''-like structure. One can impose additional constraints or losses to encourage a partial order among concepts if needed, or at least yield meaningful overlaps. But typically, we just train it to produce membership degrees.

\paragraph{Neural Approximator $N$} This network has

\begin{itemize}
\item Input: $\mathbf{f}(\mathbf{x}) \in[0,1]^{n}$
\item Output: $\hat{\mathbf{y}} \in \mathbb{R}^{\ell}$
\end{itemize}

and again the rough architecture of an MLP or multi-layer net, potentially a quite large one if the system's outcomes are complex.
Potentially this network could include skip connections to ''core concepts,'' ensuring they appear distinctly in the hidden layers.

The key point here is: If $\hat{\mathbf{y}}$ accurately predicts the system's outcomes from fuzzy concept memberships alone, that indicates $\mathcal{F}$ has discovered a concept set $\left\{F_{j}\right\}$ that is relevant for capturing the system's dynamics.

\paragraph{Loss and Co-Training}  These two networks must be co-trained.   For each sample ( $\mathbf{x}_{i}, y_{i}$ ), we compute:

\begin{enumerate}
  \item $\mathbf{f}_{i}=\mathcal{F}\left(\mathbf{x}_{i}\right)$.
  \item $\hat{y}_{i}=N\left(\mathbf{f}_{i}\right)$.
  \item Loss: $\mathcal{L}=\sum_{i}\left\|\hat{y}_{i}-y_{i}\right\|^{2}$ or $\left\|\hat{y}_{i}-y_{i}\right\|$ plus any regularization on $\mathbf{f}_{i}$.
\end{enumerate}

 We do end-to-end training: $\nabla_{\theta_{\mathcal{F}}, \theta_{N}} \mathcal{L}$. Possibly we explicitly incorporate constraints for core ontology concepts or a small penalty if fuzzy membership is too contradictory.

\paragraph{Incorporation of Core Ontology}  To heavily militate toward core ontology usage, we can define pre-wired output neurons for each core concept $C_{r}$. 

For instance, concept \#1 is ''Food,'' concept \#2 is ''Danger.'' The architecture ensures for membership neuron \#1:

\begin{itemize}
  \item Some partial or entire weight vector is fixed or partially pre-trained, forcing it to rely on known indicative features.
  \item The rest is trainable to refine membership boundaries in a data-driven manner.
\end{itemize}

\noindent Hence these ''core concepts'' are integrated from the start, and the net learns additional ''discovered concepts'' for the remainder. We can unify them by letting the net produce an $n=(r+s)$-dim membership vector, where $r$ is the number of core concepts, $s$ the number of newly discovered ones.

\subsubsection{Post-Learning: Hypervector Construction}

Now once all this learning has succeeded and $\mathbf{f}(\mathbf{x})$ is stable, we can define a hypervector for each state $\mathbf{x}$ :

$$
\mathbf{h}(\mathbf{x})=\operatorname{Bind}\left(\ldots \operatorname{Bind}\left(\mathbf{v}_{F_{j}}, f_{j}(\mathbf{x})\right) \ldots\right)
$$

\noindent or better yet, in a Snaider ''modular'' approach as described above, we might do a ''two-block'' method:

\begin{enumerate}
  \item Block A: The original kPCA embedding or a random-lift from it,
  \item Block B: Snaider's composition of membership degrees and concept base vectors (like ''(Concept$\_j$ $\odot$ membership$\_j$) bundling across j'').
\end{enumerate}

These hypervectors can be used by another neural approximator $N_2$ or multiple specialized approximators.

\subsubsection{Architecture Summary}
So the final neural pipeline is:

$$
\mathbf{x}_{i} \xrightarrow{\text { FCL net } \mathcal{F}} \mathbf{f}_{i} \rightarrow N\left(\mathbf{f}_{i}\right) \quad \rightarrow \hat{y}_{i}
$$

where:

\begin{itemize}
  \item FCL net $\mathcal{F}$ has partial ''core concept'' neurons (some pre-defined) and additional discovered concept neurons, all trained to minimize the final error.
  \item Approximator $N$ uses only these concept memberships to do predictions, ensuring the discovered concepts are validated by how well they characterize the system's state for outcome prediction.
\end{itemize}

The architecture co-trains the fuzzy concept lattice learner and the outcome predictor, guaranteeing that the discovered concept set $\left\{F_{j}\right\}$ is truly relevant to the system's dynamics while still honoring ''core ontology'' constraints.

After training, we embed or convert each $\mathbf{f}_{i}$ or the concept sets into a hypervector form for MCR usage, enabling advanced compositional operations if desired.

\subsection{Conclusion}

Summing up, it appears that potentially, adopting a fuzzy concept-lattice approach to identify relevant concepts and features combos for use in constructing the information geometry of an ActPC-Geom system and then encoding those combos in a compositional hypervector embedding does offer a coherent path to building more semantic (and thus likely more effective) internal representations in a hybrid ActPC network.  The fact that the concepts being composed are ''demonstrably relevant to the dynamics'' is crucial: it ensures these compositional elements are not just random or arbitrary, but actually help the system's predictive-coding loops.  This synergy stands to improve the compositional generalization and memory retrieval capabilities in the ActPC approach, bridging the discrete rewriting side and the continuous side with a single unified measure-based geometry in hypervector space.

While implementation overhead and dimension management must be handled carefully, the logic of the approach is clear, and this appears another case where the efficiency of modern software and hardware infrastructures for training neural models can be leveraged in the service of a broader cognitive architecture.
  
\subsubsection{A Potential Ensemble of Approximators}

It is interesting to note that, following on from the particulars of the concept learning algorithm suggested above, we have the potential create an ensemble of neural approximators for the inverse measure-based Laplacian of the overall ActPC network being guided.  We are looking at:

\begin{itemize}
\item  One approximator ($N_0$) based ''simply'' on approximate kPCA vectors
\item One approximator ($N_1$) whose inputs are drawn solely from the concepts from the learned fuzzy FCA lattice
\begin{itemize}
\item This would be expected to give lower accuracy than $N_0$ for situations close to the historical training distribution, but might provide stronger generalization out of distribution, depending on the abstraction power of the concepts in the lattice
\end{itemize}
\item  One approximator ($N_2$) based on the hypervectors constructed from judiciously combining blocks corresponding to the kPCA vectors and the concepts in the lattice.  
\begin{itemize}
\item This conceptually should provide both accuracy in-distribution and generalization out-of-distribution, but will likely have some peculiarities due to the particularities of the hypervector construction
\end{itemize}
\end{itemize}

Whether we want to use only $N_2$ once it is built or use an ensemble of the three networks (or an even greater number of specialized networks of various sorts) remains to be determined via experiment.   There are many degrees of freedom here, but clearly also a lot of power for interleaving symbolic and subsymbolic aspects in subtle ways.

\section{Toward ActPC-Geom Based Transformers}

The principles of ActPC-Geom as we have articulated them are quite general and can be applied to a great variety of neural architectures.   Co-adapting the neural architecture as learning occurs is a valuable and viable approach.   However, it's also interesting to think about implementing specific sorts of neural architectures and training their weights with the ActPC-Geom approach.   The transformer architecture for attention-guided next-token predictions is one obvious candidate.

In this section we consider the following thought-experiment: Suppose one were to try training a transformer neural net based on ActPC type methods (using PC for the basic pretraining for next token prediction, and then ActPC for the reinforcement involved in instruction tuning, but retaining the basic structure of a transformer with multiple attention heads and a number of hierarchical layers)....  We frame an argument as to why the information geometry approach might help PC/ActPC converge more efficiently to accurate next-token predictions.

The focus is on how the geometry-aware neural learning and updating process aligns with the iterative, distribution-centric style of PC/ActPC and the structured, multi-head attention design of a transformer.

\subsection{Transformers under a Predictive Coding/ActPC Regime}

A standard transformer has:

\begin{itemize}
  \item Multiple layers, each with attention sub-layers and feedforward sub-layers.
  \item A large number of parameters $(\mathbf{W})$ that shape how tokens are embedded, how attention is computed, and how feedforward transformations occur.
\end{itemize}

\noindent When reframed under predictive coding:

\begin{itemize}
  \item Each layer (or sub-layer) tries to predict some representation of the next layer (or next token), measuring mismatch (error) if the actual representation deviates from predictions.
  \item ActPC extends predictive coding into an RL-like setup, introducing rewards or penalties (e.g., from an instruction-tuning step) that shape the same error-minimizing updates.
\end{itemize}

Obvious challenges in making this work would include:

\begin{itemize}
  \item Massive Parameter Space: Transformers often have billions of parameters, which can lead to slow or unstable training.
  \item Iterative Inference + Training: If we attempt local PC updates in each sub-layer, we must keep track of partial derivatives or local divergences and adapt them in real time.
  \item Sparse Rewards: Instruction tuning or RL-fine-tuning often adds a new dimension of reward signals, compounding the complexity of distribution mismatch.
\end{itemize}

However, there are also fairly clear ways that information geometry could help meet these challenges.

For instance, there is an interesting potential to do local distribution matching in each attention head.   That is: In a PC/ActPC approach to transformers, each attention head effectively predicts or filters how tokens (or states) relate to one another. If we treat each head's output distribution as a probability distribution, then we can make an analysis like:

\begin{itemize}
  \item {\bf Ground Metric}: The ''cost'' of moving from one attention configuration to another can be defined (at least conceptually) with a metric $\omega$.
\item {\bf Measure-Dependent Operators}: A measure-dependent Laplacian encodes how a local mismatch in attention distributions (or feedforward transformations) can be efficiently ''transported'' to the correct distribution.
\end{itemize}

By adopting a geometry-based update (via following a Wasserstein natural gradient flow), the system should:

\begin{itemize}
  \item Move in parameter space in ways that minimize ''transport cost'' (i.e., abrupt or tangential corrections are smoothed out).
  \item Effectively capture how local distributions at the attention head or feedforward output shift from iteration to iteration.
\end{itemize}

Predictive coding in a transformer would then proceed by repeatedly refining each layer's representation of the tokens, top-down and bottom-up. This iterative approach is akin to a ''dynamical system,'' where each small step tries to reduce local mismatch:

\begin{itemize}
  \item {\bf Geometric Update}: If each small step uses a geometry-aware update, it has a better chance of converging smoothly.
  \item {\bf Global Coherence}: The system's heads and layers are not updated in purely decoupled ways; geometry-based metrics can reduce contradictory or oscillatory updates across layers.
\end{itemize}

There seems potential here to overcome the problem of vanishing/exploding gradients that arises in standard backpropagation-based methods, where large or unstructured parameter updates can stall or overshoot.  By contrast, a measure-dependent geometry can reveal ''natural directions'' for parameter changes, informed by how the network's (layer-wise) distributions actually rearrange.   Instead of meandering in a high-dimensional space, the system can potentially follow a more direct ''transport path'' toward lower error states --diminishing local minima or ill-conditioned curvature.

\subsection{Synergy of Instruction Tuning with PC/ActPC}

In a traditional transformer training methodology, when shifting from unsupervised next-token prediction to instruction tuning, we incorporate human or environment feedback about which outputs are ''desirable.''

ActPC does things a little differently, via merging these external signals into the same predictive-coding error minimization framework (reward-laced error terms).

That is, under an information-geometric approach we have the following intersecting aspects:

\begin{enumerate}
  \item Local Distribution: Each sub-layer's predicted representation or attention weighting is considered a distribution over tokens or hidden states.
  \item Reward: Tells us that certain next-token sequences or certain ''attention/hidden trajectories'' produce better alignment with instructions.
  \item Geometry-Aware Steps: Minimizing error + maximizing reward can be done via measuredependent gradient flows, leading to fewer large, detrimental updates in the high-dimensional space of a multi-layer transformer.
\end{enumerate}

Potential advantages of an ActPC-Geom approach versus a traditional backpropagation based approach include:

\begin{itemize}
  \item {\bf Stable Convergence}: If each iteration or each batch of instruction tuning uses a geometry-based step, the system is less likely to degrade previously learned knowledge (catastrophic forgetting).
  \item {\bf Data Efficiency}: Fewer updates are wasted on random local re-corrections; the system moves ''in the right direction'' more robustly each time.
\item {\bf Iterative Refinement}

\begin{itemize}
  \item The system is continuously refining local predictions.
  \item A measure-dependent operator (like Wasserstein) is well-suited to small local updates to probability distributions.
\end{itemize}
\end{itemize}

\subsection{Online Learning in ActPC-Geom Transformers}

Another substantial advantage of ActP-Geom based transformers would be their amenability to online learning.

Typical backprop-trained transformers operate via Batch/Minibatch Updates:

\begin{itemize}
  \item They gather data in batches or episodes, compute a global gradient by running a forward and backward pass, then do a single large parameter update.
  \item Attempting to do this at every new token or new environment step is computationally huge (billions of parameters).
\end{itemize}

Their training involve a full Forward-Backward Graph, such that:

\begin{itemize}
  \item The computation graph spans multiple layers and attention heads, requiring $\mathcal{O}\left(n^{2}\right)$ or $\mathcal{O}\left(n^{3}\right)$ with large context windows.
  \item Doing this ''instantly'' as new tokens or signals arrive in real-time is generally intractable, and can also degrade previously learned knowledge in unpredictable ways
\end{itemize}

Predictive Coding (PC), on the other hand, breaks large networks into smaller modules or layers, each computing and minimizing local prediction errors in real time.   This is intrinsically online learning friendly.  Active Predictive Coding (ActPC) extends this by incorporating reward or RL signals into the same error-minimizing framework. Instead of large global gradient steps, each module/layer adapts incrementally and locally -- retaining the same online learning friendliness of generic PC.

 In PC/ActPC, the network continuously refines its representations and parameters as new data arrives, akin to a dynamical system that settles into lower-error configurations.   There is no requirement to unroll a large computational graph for a big backward pass across every attention head/layer at once.   ActPC can update a small subset of rules or neuron states (or local weights) at each micro-iteration.   It does not rely on storing massive intermediate activations to compute full global gradients.

The various approximation methods we've introduced above complexify the mechanics of online learning a bit, but don't diminish its practicality.   Since the agent's distribution $\mathbf{x} \mapsto p\left(\mathbf{x} \mid \boldsymbol{\xi}_{t}\right)$ changes with new data or new actions, the approximate kernel PCA or hypervector embeddings used will need to be updated to incorporate new factor matrices, as part of online learning.   However, this is fundamentally unproblematic.   Incremental kernel PCA and related updates can be done with a small buffer of recent data, merging them into the existing embeddings. This can happen at a slower timescale than real-time micro-steps (e.g., perhaps every few seconds or minutes).

\subsubsection{ Step-by-Step Online Learning Procedure}

Below is a conceptual outline of how online learning might work in an ActPC-Geom based transformer:

\paragraph{1. Initialize ActPC-Transformer:}
\begin{itemize}
  \item The network has multiple layers, each with attention heads and feedforward sub-layers.
  \item Instead of backprop, each sub-layer is assigned local PC/ActPC update rules that manage (a) local error signals, (b) distribution representations.
\end{itemize}

\paragraph{2. Initialize Approximations:}
\begin{itemize}
  \item A neural net $f_{\theta}$ for the inverse measure-dependent Laplacian's low-rank factorization, trained (initially) on some offline or small dataset.
  \item Possibly an approximate kernel PCA embedding matrix, hypervector embedding, etc., capturing typical geometry from prior knowledge.
\end{itemize}

\paragraph{3. Real-Time Loop (each new token or environment input):}
  
\begin{enumerate}

  \item Local Error Computation: Each layer's representation compares predicted vs. actual hidden states or attention distributions for the newly arrived token(s).
  \item Distribution Feature Extraction: The agent extracts features $\mathbf{f}_{t}$ describing the current distribution, based on partial states/activations in the transformer.
  \item Neural Approximator:

$$
\hat{L}_{t}^{\dagger}=\operatorname{decode}\left(f_{\theta}\left(\mathbf{f}_{t}\right)\right)
$$

giving an approximate inverse measure-dependent Laplacian
\item Local Parameter Update:

$$
\boldsymbol{\xi}_{t+1}=\boldsymbol{\xi}_{t}-\eta \hat{G}\left(\boldsymbol{\xi}_{t}\right)^{-1} \nabla_{\xi} F\left(\boldsymbol{\xi}_{t}\right), \quad \text { where } \hat{G}\left(\boldsymbol{\xi}_{t}\right)=J_{\boldsymbol{\xi}_{t}}^{\top} \hat{L}_{t}^{\dagger} J_{\boldsymbol{\xi}_{t}} .
$$

\noindent This step is done locally in each attention head or sub-layer of the transformer. Each microupdate adjusts weights or states to reduce local error.
\end{enumerate}

\paragraph{Online Updating is Natural Here}  The amenability of ActPC to online learning carries through to ActPC based transformers in a natural way.   In the background and/or at a slower frequency, one. may update the approximate kernel PCA subspace or the neural net $f_{\theta}$ so that it remains relevant to the evolving environment distributions.  This keeps the geometry and the network up-to-date without doing a huge global backprop pass.  Each sub-layer is iteratively refined in small increments, relying on local predictive errors (the hallmark of PC/ActPC).    The main transform-layers are updated in a local sense, while all the while the network remains ''active'' and can produce tokens in real time.

\subsubsection{Example Scenario}  To better understand how these parts work together, consider a large ActPC-Transformer performing dialogue generation in an interactive environment, and think about what happens when a user provides an instruction or corrects a statement:

\begin{enumerate}
  \item Incoming Token: ''That is incorrect, the correct answer is ...''
  \item Local Error: The model's predicted representation for the next tokens is mismatched with the user's correction. Each transformer sub-layer sees the discrepancy.
  \item Neural Approximator: The network extracts new distribution features, passes them into $f_{\theta}$, obtains $\hat{L}^{\dagger}$.
  \item Micro-Updates: Each sub-layer's parameters do a small geometry-based update to reduce the local mismatch, factoring in any RL-like reward from user feedback.
  \item Online Convergence: Over repeated tokens/interactions, these updates accumulate. The system ''adapts'' to the user's instructions in real time, refining knowledge or style.
\end{enumerate}

\noindent The entire process is distributed and continuous.

\subsection{Potential Synergies Between Emergent In-Context Activation-Space Learning and Online Weight Updating}

Now we come to some points that are a little less obvious -- regarding the intersection of ActPC-Geom mechanisms with the emergent in-context learning one sees in larger transformers.

Some of the more interesting dynamics we've seen come from transformers have to do with learning in the ''activation space'' happening even without weights in the neural net being updated. This underlies the few shot learning that modern large transformers are capable of. This could presumably happen in a transformer trained via the ActPC-Geom approach as well, but there would be a different twist because of localized online learning. 

One could potentially have a new sort of cognitive dynamic, where few-shot learning in the activation space is dynamically interwoven with localized real-time update of the weights.

\paragraph{In-context few-shot learning in transformers}  To recap the phenomenon we are discussing here: In classical large transformers, a prompt (e.g., a short few-shot example) feeds into the model, shaping activation patterns through multi-head attention and feedforward layers. Even without any changes to the model weights, the network's activation flow effectively ''configures itself'' to handle the new task ephemerally. This ephemeral reconfiguration is:

\begin{itemize}
  \item Local to the forward pass: The hidden states $\mathbf{h}_{\ell}$ and attention patterns adapt to the prompt context.
  \item Vanishing Post Forward Pass: Once the forward pass is done, no permanent weight changes are made-this ''solution'' is ephemeral.
\end{itemize}

\paragraph{Predictive Coding and in-context learning}  If we consider this sort of in-context learning from a predictive-coding perspective, we may consider that:

\begin{itemize}
  \item The model's hidden states $\mathbf{z}^{\ell}$ at each layer $\ell$ are corrected iteratively via local mismatch from topdown or bottom-up predictions, often referred to as the ''PC dynamical system.''
  \item This iterative correction in activation (hidden states) can mimic the ephemeral few-shot ''learning'' we see in standard transformers, but it can happen more explicitly as repeated small inference steps that shape each layer's $\mathbf{z}^{\ell}$.
\end{itemize}

\noindent Further, we add to this picture

\begin{enumerate}
  \item Adaptive accommodation to reward or global objective signals.
  \item Local, gradient-free weight adjustments that minimize prediction error in each sub-layer or attention head, typically in small increments each time an error signal is detected.
\end{enumerate}

\noindent That is: Beyond ephemeral changes in the activation space, we now have small real-time updates to the network parameters themselves, potentially:

\begin{itemize}
  \item Retaining aspects of the ephemeral solution gleaned from the new context.
  \item Accumulating knowledge that might help with future tasks.
\end{itemize}

\subsubsection{Integration with Wasserstein Geometry and Neural Approximators}

To think about this potential new form of learning synergy in depth, we must return to measure-dependent Laplacian $\hat{L}$  that (in the ActPC-Geom paradigm) shapes the ActPC parameter updates in a geometry-aware manner, via its neural and embedding-vector approximations. Conceptually:

\begin{enumerate}
  \item Activation-Space Inference: At each layer $\ell$, hidden states $\mathbf{z}_{\ell}$ are refined in real time as new tokens or new context arrives (the ephemeral, few-shot style of ''learning in activation'').
  \item Local Weight Update: Simultaneously, or in small intervals, each sub-layer uses a measure-dependent operator $\hat{L}^{\dagger}$ to update a portion of its weights, guided by a local error function reflecting the mismatch between predicted and actual hidden states.
  \item Neural Approximator: The inverse measure-dependent Laplacian is approximated by a neural network that sees (embedding vectors or hypervectors produced from) distribution or layer features $\mathbf{f}$. This yields geometry-based updates for the local weight parameters, reflecting how the internal distribution changes as the ephemeral activation-based solution emerges.
\end{enumerate}

Integrating these various points, we can picture a two-timescale dynamic:

\begin{enumerate}
  \item Fast timescale: ephemeral in-activation few-shot learning. The transformer's hidden states $\mathbf{z}^{\ell}$ continuously rearrange in short-run inference loops, responding to new tasks in context.
  \item Slow timescale: real-time local weight updates. The measure-dependent Laplacian approach yields geometry-aware ActPC steps for $\mathbf{W}^{\ell}$. Over repeated exposures, these steps accumulate, embedding ephemeral solutions into the network's parametric memory.
\end{enumerate}

To illustrate, let's walk through the step-by-step dynamics in a single mini-episode:

\paragraph{1. Prompt Arrival (Few-Shot Context)}
\begin{itemize}
  \item The transformer receives a new input sequence containing a demonstration or instructions for a novel task.
  \item In a standard transformer, the forward pass activation patterns shift to reflect the ephemeral solution.
  \item In a PC/ActPC transformer, these ephemeral patterns evolve iteratively, as local prediction errors in each sub-layer are successively reduced:

$$
\mathbf{z}^{\ell} \leftarrow \mathbf{z}^{\ell}-\eta_{z}\left[\nabla_{\mathbf{z}^{\ell}}(\text { Local PC error })\right]
$$

  \item This process can be done in multiple micro-iterations or a short forward ''settling'' loop.
\end{itemize}

\paragraph{Local Real-Time Weight Update}

\begin{itemize}
  \item In addition to ephemeral $\mathbf{z}^{\ell}$ updates, each sub-layer or attention head has local weight parameters $\mathbf{W}^{\ell}$.
  \item ActPC states that partial mismatch (error) signals also drive small adjustments in $\mathbf{W}^{\ell}$.
  \item These updates are shaped by an approximate measure-dependent Laplacian:

$$
\mathbf{W}_{t+1}^{\ell}=\mathbf{W}_{t}^{\ell}-\eta_{W}\left[G\left(\mathbf{W}_{t}^{\ell}\right)\right]^{-1} \nabla_{\mathbf{W}^{\ell}} E^{\ell}
$$

\noindent where $E^{\ell}$ is the local error energy at layer $\ell$, and $\left[G\left(\mathbf{W}_{t}^{\ell}\right)\right]^{-1}$ is derived from the approximate kernel PCA + neural approximator that models the geometry of distribution changes in that sub-layer's representation.
\end{itemize}

\paragraph{3. Ephemeral vs. Persistent Changes}
\begin{itemize}
  \item As tokens flow in, the ephemeral updates in activation space solve the short-run problem. For instance, the model learns ''in context'' how to do a certain Q\&A pattern.
  \item Meanwhile, the persistent weight updates-though small-accumulate over repeated tasks, so if the user asks something similar next time, the ephemeral solution can be found more quickly or with fewer micro-iterations.
  \item This synergy means short-run ephemeral adaptation (few-shot style) and small permanent improvements are interwoven.
\end{itemize}

\paragraph{4. Dynamics in the ''Cognitive Kernel''}
\begin{itemize}
  \item Because each layer is doing local PC/ActPC updates, the entire network's ephemeral solution emerges from repeated local corrections in $\mathbf{z}^{\ell}$.
  \item The presence of small weight updates biases the system's ephemeral solution to converge in a direction that also yields better future performance, effectively bridging immediate in-context solutions with ongoing ''long-term'' learning.
\end{itemize}

As in-context learning is still only weakly understood on a theoretical level, it's difficult to limn the precise details of this sort of synergetic learning without running large-scale experiments, adjusting parameters and seeing what happens.   But it seems clear there's a lot of potential power here.

\paragraph{Why the Measure-Dependent Operator Matters for This Dynamic}
\begin{enumerate}
  \item Match to Distribution Shifts:

\begin{itemize}
  \item As the ephemeral solution emerges in activation space, the local distribution of hidden states or attention patterns changes from time step to time step.
  \item The measure-dependent Laplacian approach sees these changes not as random local gradients but as transport in the distribution manifold of hidden states.
  \item The weight updates reflect the ''real cost'' of repositioning parameters in a geometry aligned with how hidden states shift.
\end{itemize}

  \item Faster Convergence:
  
\begin{itemize}
  \item A naive Euclidean approach might require many micro-iterations of partial error correction to integrate ephemeral patterns.
  \item The geometry-based approach can ''hop'' more directly to parameter configurations that reduce mismatch, accelerating assimilation of ephemeral knowledge into stable rules.
\end{itemize}

  \item Incremental Nature of Neural Approximator:

\begin{itemize}
  \item The distribution features from the ephemeral states feed into the neural approximator, which quickly outputs an approximate inverse operator $\hat{L}^{\dagger}$.
  \item This ensures the local step is consistent with how the hidden-state distribution is actually shifting during ephemeral reasoning.
\end{itemize}

\end{enumerate}

\subsubsection{Potential Emergent Advantages}

The potential advantages of this synergetic learning approach appear significant.

\paragraph{Blending In-Context Learning with Weight Consolidation}
\begin{itemize}
  \item Many practical tasks only appear once or sporadically. The ephemeral in-activation solution is powerful, but ephemeral knowledge can vanish after the session.
  \item With ActPC, some of that ephemeral solution is ''anchored'' via small updates that accumulate if that pattern is encountered repeatedly. Over many exposures, ephemeral solutions become partially or fully ''internalized'' in the network's weight structure.
\end{itemize}

\paragraph{Less Catastrophic Interference} Because updates are local and small, the model is less likely to break existing capabilities. The measure-dependent geometry further reduces abrupt parameter changes.

\paragraph{Continuous Life-Long Learning}
\begin{itemize}
  \item Over time, the network gradually expands its knowledge base, guided by ephemeral solutions that appear in activation space for new tasks.
  \item Real-time assimilation means the system doesn't need offline fine-tuning phases.
\end{itemize}

\paragraph{ Persistent Weight Updates}  While ephemeral adaptation in the activation space occurs (as in normal few-shot usage), small local weight updates also occur in real time, guided by predictive-coding error signals.  These updates can partially ''lock in'' aspects of the ephemeral solution encountered in the prompt, so that next time a similar pattern arises, the model's baseline distribution of representations is already shifted in the correct direction.

\paragraph{ Better Handling of Repeated or Extended Few-Shot Scenarios} If the user or environment repeatedly introduces new tasks, incremental learning accumulates. The ephemeral solution for a new pattern, if encountered multiple times, can become stably embedded in the network. This goes beyond typical few-shot LMs that rely on ephemeral context alone.\\

\paragraph{Potential for Iterative ''Inner Loop + Outer Loop''}

The local iterative inference in activation space (the ''inner loop'' in this context) merges seamlessly with incremental structural changes in weights (the ''outer loop'' in this context). If a new pattern is complex enough to require multiple forward passes (deliberation), the ephemeral solution can also drive small but meaningful changes in network connectivity or attentional patterns.

\subsubsection{Emergent ''Deliberative'' Cognition}

This sort of emergent learning dynamic weaving together in-context activation-space learning and online localized weight updating has potential to support more advanced forms of deliberative cognition than standard transformers.

Standard LLMs can show emergent chain-of-thought or reasoning ''styles,'' but these are implicitly encoded in feedforward passes. They do not deliberately revise their own parameters within that session to refine strategies.

In the ActPC-Geom approach, on the other hand, deliberation can be viewed as iterative or recurrent inference in activation space, guided by local error signals at each layer.   The measure-dependent geometry helps steer the parameter updates so the system actively ''tunes itself'' in real time, refining not only ephemeral states but also small aspects of the network's weighting.  This synergy leads to a two-level approach to deliberation:

\begin{enumerate}
  \item Fast ephemeral shaping of activation states to handle the immediate question or context,
  \item Slow local weight changes that push the network's baseline distribution toward better internal coherence for that line of reasoning.
\end{enumerate}

\noindent One would expect this to lead to a more incisive or stable form of chain-of-thought because:

\begin{itemize}
  \item Activations in each step get ''amplified'' by the slight weight modifications that unify partial solutions discovered so far.
  \item The measure-dependent Laplacian approach can identify minimal parameter shifts that drastically reduce the mismatch in higher-layer reasoning states, potentially ''unblocking'' a line of reasoning or clarifying ambiguities in the middle of a multi-step reflection.
\end{itemize}

\subsubsection{Impact on Long-Term Memory Integration}

Standard transformers are relatively poor at associative memory retrieval, even when their memories contain the needed knowledge to respond to a query.   Retrieval Augmented Generation (RAG) and related frameworks like GraphRAG are used to work around these issues, via interactions between LLMs and external knowledge bases (which e.g. can fish relevant information out of external KBs via non-LLM query methods and then pull them into the LLM's current context), but it's often difficult to tune them to work effectively.    In the end, one really needs tighter interconnection between the query resolution process and the information retrieval process than what RAG and related methods provide.

In Section \ref{sec:hopfield} we propose a method for integrating associative long-term memory into ActPC-Geom based transformers via inserting Hopfield-net style connectivity within the layers.   However, it also seems that, even without this sort of enhancement,
ActPC based transformers have significant potential to bring RAG-like processes within the transformer itself.   If the network can quickly rewire or reinforce parameter connections in real time, the boundary between ephemeral context and ''internal memory'' is blurred in a highly cognitively relevant way.  For instance:

\begin{itemize}
  \item A short, retrieved piece of text might not only shape the ephemeral activation patterns but also cause local link updates that better connect that textual knowledge into the model's longer-term representation of domain knowledge.
  \item If, during ephemeral reasoning, the model forms strong associations between a newly discovered concept in the working memory context and some existing concept, the local error signals can nudge the relevant attention/FFN weights to encode that associati
  \item Over repeated use, the model might no longer need external retrieval for that piece of knowledge because it's (partially) integrated into the network's parameters.
  \item This reduces the necessity for repeated external retrieval of the same snippet or the same chunk, since the association is now in the network itself.
\end{itemize}

Realistically, we might still use external retrieval to handle large or dynamic knowledge bases, but the difference is that the boundary between ephemeral context and ''stable memory'' in the net's parameters becomes more permeable. If a piece of knowledge is used often enough in ephemeral context, the model accumulates it in the weights. This yields a continuum from purely ephemeral, in-context knowledge to permanently integrated knowledge.

If a user or system re-invokes a new concept or skill multiple times, local updates will accumulate. The model's baseline distribution becomes more adept at that skill, potentially reducing the context length needed in future prompts. In a standard few-shot model, the user must keep re-providing examples.

Further, in this approach, the ephemeral chain-of-thought that emerges in activation space can feed back immediate local error signals, prompting small but context-specific weight modifications to reinforce crucial tokens or patterns. This can produce a more robust and stable chain-of-thought approach, akin to a short-run memory consolidation in the network's architecture.

\subsubsection{Potential for Deliberative Metacognition}   There are also potentially broader implications of this sort of dynamic for reflective meta-cognition:

\begin{itemize}
  \item With iterative inference, the model can ''notice'' that its ephemeral solution is incomplete or inconsistent.
  \item Real-time local weight updates can shift relevant heads or feedforward sublayers to fix the mismatch.
  \item Over multiple steps, a more thorough ''internal check'' can occur, weaving ephemeral reasoning with stable modifications that persist for subsequent steps in the conversation or session.
\end{itemize}

\subsubsection{ Summary / Hypothesis}

An online-learning ActPC-based transformer could transcend standard few-shot learning by interweaving ephemeral solutions in the activation space with small local weight updates in real time. This might:

\begin{enumerate}
  \item Enable an ongoing ''deliberative'' dynamic, bridging ephemeral chain-of-thought with incremental network modifications for future reuse.
  \item Empower the network to incorporate new knowledge from ephemeral context directly into its ''longterm memory'' (the net's parameters) without requiring large offline fine-tuning or reliance on repeated retrieval from external memory systems.
  \item Potentially render certain retrieval-augmented techniques less essential, as recurring ephemeral knowledge can be integrated into the net's stable parameters if it proves repeatedly useful, obviating the strict ephemeral/long-term boundary.
\end{enumerate}

Thus, the synergy of few-shot learning in activation space plus local ActPC-based weight updates can yield a more adaptive and integrative learning process than standard backprop-based transformers can currently achieve, opening the door to a more ''cognitive'' style of real-time, deliberative, and self-improving AI.

\subsection{Online-Learning Performance Requirements for Emergent Deliberative Cognition: A Very Preliminary Analysis}

Having established that synergetic interaction between online learning in activation space and weight updating is likely highly valuable for upgrading transformers' cognition, we now turn to quantitative aspects.   How fast must or should weight updating be relative to activation spreading for the right sorts of cognitive synergy to manifest themselves?

Conceptually, it seems clear that for online learning in a PC/ActPC-based transformer to effectively complement ephemeral few-shot adaptation, weight updates must typically be somewhat slower than the ephemeral activation updates (to let the ephemeral solution stabilize) but fast enough to incorporate new knowledge within the same session or conversation. 

In practice, the weight updates likely need to be:

\begin{itemize}
\item Slower Than Activation Convergence
\begin{itemize}
  \item Let each sub-layer's activation states (the ephemeral solution) settle into an approximate stable configuration for the current context.
  \item Only after this partial settling do we measure the stable residual error signals, then make small updates to the sub-layer's parameters.
\end{itemize}

  \item Fast Enough to Aid Within the Session

\begin{itemize}
  \item If a new pattern or skill arises mid-session, we want at least some portion of the session or conversation to benefit from partial assimilation of that skill into the weights.
  \item So, we can't wait hours or thousands of forward passes before applying any weight changes.
\end{itemize}
\end{itemize}

Consider a hypothetical example: In a real-time conversation, the user provides a new puzzle, and the model tries an ephemeral chain-of-thought approach (over 3-5 inference steps). After the puzzle is solved or the user says ''Yes, that's right,'' each sub-layer performs a small parameter update. This entire cycle might take 2-3 seconds in total:

\begin{enumerate}
  \item (0-1s) Model quickly iterates local activation refinements, forming ephemeral logic about the puzzle.
  \item (1s) Gains partial confidence, produces an answer.
  \item (1-2s) The user says ''Yes, correct!'' (reward).
  \item (2s) The network triggers small local updates, guided by measure-dependent geometry, to more firmly embed the puzzle-solving pattern.
  \item (2-3s) The conversation moves on, ephemeral states shift for a new question. Some newly updated weights remain, facilitating a future puzzle of similar type.
\end{enumerate}

One can envision multiple methods for triggering weight updating based on the system's cognitive progress in a given incidence, e.g.

\begin{itemize}
\item  Confidence Thresholding

\begin{itemize}
  \item Wait until the local activation states have ''settled'' (i.e., the local error $\left\|\mathbf{z}^{\ell}-\hat{\mathbf{z}}^{\ell}\right\|$ stops significantly decreasing).
  \item Then apply a small weight update.
  \item If the session context changes drastically, ephemeral states reconfigure and only after they settle again do we do another weight update.
\end{itemize}

\item Reward-Triggered
\begin{itemize}
  \item If a user or environment event signals a success or failure, that can trigger a wave of local parameter updates, locking in or correcting the ephemeral solution.
  \item This ensures the weight changes occur precisely when new knowledge or skill is validated.
\end{itemize}
\end{itemize}

While the specific optimal timing will depend on the situation, crudely it seems a ratio of 2:1 to 10:1 ephemeral updates to parameter updates might be a plausible rule of thumb. This sort of timing may effectively enable ephemeral adaptation (few-shot learning in activation space) proceeds without interference from overly frequent weight changes, while still letting small local weight updates accumulate promptly, capturing beneficial ephemeral insights for future re-use.

This leads to the question whether a 10:1 ratio (ephemeral activation updates vs. weight updates) is actually feasible in practice, given the overhead of online weight updating in a large transformer-like network.   In the following we conclude the answer is probably yes, but only if one embeds some special methods in the system, such as speeding up or focusing weight updates on the critical sub-network that's actually engaged in the few-shot reasoning, thus lowering computational load and making real-time online adaptation more attainable.

\subsubsection{Meeting the Challenge of Weight Updates in Real Time}

In a large ActPC-based transformer, even a local weight update can still require:

\begin{itemize}
  \item Accessing parameter sets for multiple attention heads and feedforward sub-layers.
  \item Computing local error signals and (in the measure-dependent geometry approach) partial operators ( $\hat{L}^{\dagger}$ or its factorization).
  \item Applying small parameter deltas in real-time.
\end{itemize}

\noindent If we tried to do these steps every time we do ephemeral activation refinements (which can happen at a high frequency), overhead might become prohibitive. A ratio of 1:1 would likely be too large a burden for most hardware.   On the other hand, it might also not be desirable, if one follows the above ideas regarding the value of having two different time scales involved (so that in-context learning has a chance to settle in a bit before being disrupted by major weight changes).

A 10:1 ratio, on the other hand, is still ambitious but appears potentially feasible.  In this approach, if ephemeral micro-iterations happen on the order of tens or hundreds of milliseconds each, weight updates might happen once every second or so.

Achieving that on large networks means we must keep the cost of each weight update sufficiently low -- likely implementing partial or approximate updates (rather than updating the entire multibillion parameter set in one shot).

Available tooling is only partially on our side here.   Current deep-learning infrastructure is heavily optimized for forward/backward passes, not for small incremental parameter changes in many scattered sub-layers.   If, for instance, each layer has to lock memory sections for weight updates in real-time, concurrency and overhead could balloon.   Still, 10:1 is not unimaginable -- if we limit the scope of each update and rely on partial or approximate updates in appropriate manner.

\paragraph{Localizing to the Nodes/Links Most Engaged}.  One practical strategy would be to focus real-time weight updates on the part of the network that is currently:

\begin{itemize}
  \item Receiving the most activation (i.e., the sub-layers/heads that are key to the current few-shot reasoning),
  \item Driving the largest local error signals (the ''center of mismatch''), or
  \item Most relevant to the user's query (identified by attention maps or gating signals).
\end{itemize}

\noindent For instance, if only certain attention heads or certain feedforward ''expert modules'' are strongly activated for the new task/prompt, we can concentrate the computational budget there:

\begin{enumerate}
  \item Sparse Weight Updates: Skip or defer updates to sub-layers that are mostly dormant for this particular context.
  \item Error-Driven Focus: If local mismatch signals are near-zero in some layers, no reason to spend time on their parameter updates. Focus on where the error is large.
\end{enumerate}

\paragraph{Micro-Learning in the Active Subnetwork} By restricting updates to $\sim 1-5 \%$ of the network's parameters that are actually ''hot'' at a given point in time, the overhead for computing partial geometry-based steps is drastically lowered.

If we only update $\hat{L}^{\dagger}$ for that local subgraph, the cost of updating the needed approximations in real-time may be feasible.

This approach means the network, at any given moment, invests its real-time learning resources where ephemeral reasoning is happening, making that 10:1 ratio (or even $5: 1$ or $3: 1$, perhaps ) more plausible.

\paragraph{Random or Priority Sampling}.  Instead of updating all relevant parameters at once, we could:

\begin{enumerate}
  \item Sample a small fraction of them each micro-step, applying the measure-dependent gradient. Over multiple steps, each parameter gets updated often enough to matter.
  \item Prioritize parameters with large local gradient magnitudes or large mismatch signals.
\end{enumerate}

\noindent This strategy could be combined with the strategy of focus on the active subnetwork.   I.e. the active subnetwork gets updated immediately, and the rest of the network gets updated via priority sampling.  This is the strategy OpenCog has long deployed in its ECAN attention allocation subsystem -- updating attention levels in the attentional focus rapidly, and doing so in the rest of long-term memory using stochastic importance sampling \cite{goertzel2016controlling}.

\paragraph{''One-Shot'' or ''Few-Shot'' Weight Corrections}  We might allow the network to do a mini ''outer loop'' each time a chunk of ephemeral reasoning stabilizes. For example:

\begin{enumerate}
  \item After the ephemeral states converge for the current segment of text, do a short burst of partial weight updates (like a micro-batch of 10-20 parameter subsets).
  \item Then move on. This can still keep the overhead small relative to the ephemeral forward pass.
\end{enumerate}

\paragraph{Example Workflow to Achieve 10:1 Ratio}.  These pieces might be put together roughly as follows, for instance:

\begin{enumerate}
  \item Ephemeral Iterations (Activations)
\begin{itemize}
  \item Let the network run $~ 10$ iterative refinement steps in activation space over $\sim 200-300 \mathrm{~ms}$, guiding the hidden states to a stable partial solution.
\end{itemize}

  \item Local Parameter Update (1 step)

\begin{itemize}
  \item Identify the top sub-layers or heads responsible for the largest mismatch.
  \item Run the measure-dependent geometry on those sub-layers' parameters only.
  \item Apply small updates within 50 ms .
\end{itemize}

\item Continue Next Token

\begin{itemize}
  \item Move on to the next token or next partial chunk of text, with ephemeral states re-initialized or partially carried over.
\end{itemize}

\end{enumerate}

\noindent In this hypothetical narrative, ephemeral steps:weight updates = 10:1, and total overhead is kept in the same ballpark as normal forward pass computations if partial updates are used.

Our overall, tentative conclusion here is: While out-of-the-box it may be hard to do 10:1 in a giant network, targeted, partial online learning plus hardware/software optimizations can potentially make a 10:1 ephemeral-to-weight-update ratio realistic, preserving real-time reactivity and the synergy between ephemeral few-shot adaptation and stable incremental learning.

\section{Embedding Associative Long-Term Memory in ActPC Transformers} \label{sec:hopfield}

One weakness of current transformer neural nets, already noted above, is that they are poor at associative retrieval from long-term memory... this is why. methods like Retrieval Augmented Generation are so useful, even though they are often just enabling an LLM to look up (via relatively crude but still usefully different means) information it already has embodied in its own primary neural-net memory.   This shortcoming can likely already be addressed to a significant measure by the improvements to transformer architecture suggested in the previous sections.  However, it may also be valuable to address the problem head-on in a more direct way, via integrating into the transformer-like architecture some different neural connectivity patterns that are known to be highly effective for associative long-term memory.

A hint of a possibly productive direction may be found in the observation that, compared to backpropagation, predictive coding based learning is a bit more like (a more sophisticated version of) the Hebbian learning used e.g. in Hopfield associative memory networks, which are in their own way quite good at associative long-term memory retrieval \cite{amit1990attractor}.   There has also been recent work updating Hopfield network methodologies to leverage modern neural net tools and ideas \cite{ramsauer2020hopfield}.   All this brings up the question whether some particular architectural and/or control mechanism adjustment could/should be used in ActPC-Geom based transformers to cause them to form ''asymmetric Hopfield net like'' lateral connections within a layer, potentially then leading to more robust associative memory dynamics (which would synergize well with in context learning that transformers already do well, enabling it to better interact with the long term memory already in the transformers). 

In this section we will flesh out this idea a bit, and explore some qualitative considerations regarding how to make it work.   We will give a conceptual proposal for how transformer-like networks, trained via predictive coding (most likely with information-geometry accelerations) could incorporate Hopfield-like associative retrieval dynamics. 

\subsection{Transformers \& Predictive Coding in Relation to Hebbian/Hopfieldian Learning}

\paragraph{Why Transformers Struggle with Associative Retrieval} In standard transformers, we do multi-head attention over a window of tokens, but we do not have a strong mechanism for ''retrieving old patterns from deep inside the net's memory.'' The model's parameters have ''memorized'' some associations, but the forward pass does not easily re-activate them unless the relevant tokens are in the current context window.

Extended context transformers can handle bigger windows, but still rely on attention across many tokens. This does not replicate the classic Hebbian ''if pattern partial match, re-activate stored pattern'' dynamic typical of Hopfield-like attractor networks.

Hence: Transformers tend to be poor at ''associative leaps'' unless you feed them an explicit retrieval or context chunk. RAG attempts to fix that by letting them ''look up'' relevant knowledge from external memory.

\paragraph{ Predictive Coding as a Hebbian-Like Mechanism} Predictive coding, on the other hand, relates much more closely to Hebbian learning dynamics.   PC can be viewed as a recursive minimization of local mismatch (prediction error) in a generative network. In some forms:

\begin{itemize}
  \item Activations can reconfigure iteratively rather than in a single forward pass.
  \item Weights can also adapt in near-real-time (or micro-iterations) to reduce error.
  \item This is reminiscent of older Hebbian or Hopfield-like ''associative'' updates, where partial matches lead to strong pattern completion over a short iterative cycle.
\end{itemize}

\subsection{Augmenting ActPC-Geom Transformers with Lateral Hopfield-Like Layers}

The core idea presented in this section is to augment a transformer's layers with lateral (intra-layer) connections that let partial activation patterns ''pull each other in."   

Doing this, potentially, can lead to a mini Hopfield-like attractor dynamic within each layer or across layers, so that if the net sees a partial activation sub-pattern, it (over a small iterative refinement) re-activates the entire associated pattern.

In the context of backprop-based training of large transformer networks, we would expect this to lead to non-convergence or other tricky issues with gradients, for similar reasons to why complicatedly, richly recurrent network architectures often perform poorly with backprop.  However, PC based learning should be much more amenable to this sort of architecture, due to its conceptual synergy with Hebbian-style learning, as well as other subtler reasons.

If we adopt an info-geometry-based approach to local updates, we might let each micro-iteration move the network's state or weights in a geometry-aware manner, speeding up the attractor-based pattern recall. That is: the natural gradient might yield a direct path to the ''correct'' stored pattern if partial cues match a known attractor.

\subsubsection{Where to Add Lateral Connections?}

Consider a standard transformer block: it has

\begin{enumerate}
  \item Multi-Head Self-Attention
  \item FeedForward sublayer
  \item Residual connections
  \item LayerNorm
\end{enumerate}

\noindent We propose adding lateral or recurrent connections within a layer, specifically allowing:

\begin{itemize}
  \item Neurons (or attention heads) to update each other's activations iteratively, in a small ''recurrent pass'' that attempts to minimize predictive error among themselves.
  \item The ''asymmetric Hopfield net'' style means each connection from node $i \rightarrow j$ can have a distinct weight from $j \rightarrow i$. The local update rule is not symmetrical in the naive sense.
\end{itemize}

\subsubsection{Mechanism to Trigger a Hopfield-Like Retrieval}
We want the partial activation pattern present in the ''current context'' to pull out a larger pattern stored in the layer's parameterized memory. For instance:

\begin{enumerate}
  \item The layer sees input $\mathbf{x}$.
  \item The lateral net (or ''Hopfield block'') runs a few micro-iterations, refining $\mathbf{x}$ into $\mathbf{x}^{*}$, a stable attractor that corresponds to some known pattern.
  \item This pattern acts as a ''completion'' or ''retrieval'' from the layer's ''long-term memory'' embedded in the lateral weights.
\end{enumerate}

\subsubsection{Predictive Coding Angle}  In a predictive coding approach, each small iteration tries to reduce mismatch between the current $\mathbf{x}$ and what the internal model expects. If partial cues match some stored pattern, the layer's internal generative loop ''pulls'' the activations to that pattern to reduce error.  This is akin to how traditional Hopfield networks do ''pattern completion.''

\paragraph{Potential Synergies with Transformer Architecture/Dynamics}
\begin{enumerate}
  \item Multi-Head Attention already does some partial matching. But it's mostly ''token to token'' or ''layer to layer.'' It doesn't have a stable attractor dynamic.
  \item A Hopfield-Like lateral sub-layer could run after or alongside attention each step, letting the layer's activations settle into an attractor if partial cues appear that match a stored pattern.
  \item In-Context Learning: The partial pattern in the context might cause a self-activation of some memory from the net's parameters, enabling a robust recall of associated knowledge. This might yield more associative leaps than standard transformer attention does alone.
\end{enumerate}

\subsection{Implementation Issues}

The addition of Hopfield-like layers to a transformer is a conceptually straightforward, and the idea that ActPC-Geom should work training this sort of architecture seems plausible.   However, there are a few "minor implementation issues" that may need to be considered.

\subsubsection{Overhead vs. Gains}
\begin{itemize}
  \item We need iterative updates for each forward pass. That might slow down inference unless we keep the iteration count small or do partial concurrency.
  \item The parameterization of a Hopfield-like sub-layer is also non-trivial. We might store a big weight matrix for lateral connections. This can blow up parameter counts.
\end{itemize}

\subsubsection{Stability \& Training Complexity}
\begin{itemize}
  \item Assuming the network is trained end-to-end with predictive-coding scheme, we must ensure stable attractors that enable associative long-term memory without degrading global performance.
  \item A normal transformer might get confused if partial Hopfield updates produce unexpected or contradictory signals.
  \item Carefully balancing the ''transformer'' path and the ''Hopfield-lateral'' path is key, or else we might get "weird in the wrong way" attractors that degrade performance.
\end{itemize}

\subsubsection{Asymmetric vs. Symmetric?}  Hopfield networks are classically built on a symmetric weight matrix for guaranteed convergence, but ''asymmetric'' designs can still produce stable attractors or cycles. The advantage of asymmetric is that it might be more biologically or cognitively plausible, and capable of broader and subtler forms of knowledge representation, but we might lose guaranteed global minima. Predictive coding can offset some of these issues by an error-correction dynamic, but the analysis is more subtle.   This is a fascinating and important area for research.

\subsection{Recap}

Summing up, the dynamic we are after here is:

\begin{enumerate}
  \item Pattern Completion: A partial pattern in the context could re-activate the entire pattern stored in the net, solving the ''lack of robust associative memory'' problem.
  \item Predictive Coding helps enforce iterative refinement to a stable solution, letting the net do multiple micro-iterations per forward pass if needed.
  \item Information Geometry can help ensure updates navigate quickly to an attractor, possibly reducing overhead.
\end{enumerate}

Assuming all these pieces come together as envisioned: The net can retrieve from large LTM stored in the lateral parameters, without naive external retrieval—some of the same effect RAG aims for, but internally integrated.

If the design and implementation challenges are handled -- e.g., we do carefully tuned partial-lateral updates or run them for only a small iteration count -- the synergy with in-context learning could be powerful, letting the system do real associative retrieval from LTM, without a naive RAG-style pipeline.

\section{Symbolic/Subsymbolic Transformers via ActPC-Geom + ActPC-Chem}

In a sister paper \cite{goertzel2024actpc} we have proposed to apply ActPC principles to dynamically shape discrete networks of inter-rewriting rewrite rules evolving according to ''algorithmic chemistry'' style dynamics.  The ActPC-Chem approach suggested there involves Wasserstein-metric based information-geometric methods applies to a discrete variation of ActPC modeling.   It is also suggested there that connecting an ActPC-Chem network with a traditional ActPC-based neural network would lead to some natural synergies, including the ability to propagate prediction errors between the two coupled networks.   

The ideas presented here allow us to add to that story somewhat.   What we can now point out is that, if one hybridizes an ActPC-Chem network (or some other sort of discrete, symbolic ActPC network) with a continuous ActPC-Geom network, an additional form of synergy becomes available: namely, {\it shared geometry}.  One can use unified measure-dependent operators to guide updates across both symbolic and continuous components of a hybrid system.

Integration of ActPC-Geom with ActPC-Chem could be done in a variety of different ways and contexts.  For sake of concreteness, we will here flesh out the integration in the context of transformer-like networks.   However the savvy reader will easily be able to analogize the specific presented to other sorts of neural architectures.

\subsection{Neural-Symbolic ActPC-GeomChem Transformer: High-Level Architectural Sketch}

Imagine we have N ''layers'' (or ''blocks''), each layer comprising:

\begin{itemize}
  \item A Discrete Transformer-like Block:
  \begin{itemize}
  \item Instead of ''dense MLP + attention,'' it uses algorithmic chemistry of rewrite rules.
  \item Learning is done via discrete ActPC: local error signals drive updates to the rewrite rules, guided by a discrete measure-dependent Laplacian (or an approximate version).
  \end{itemize}
  \item A Continuous Transformer-like Block:
    \begin{itemize}
  \item A standard multi-head attention + feedforward sub-layer, but updated via continuous ActPC principles.
  \item It uses ''PC + RL + measure-dependent geometry,'' with partial local updates of continuous weights.
  \item Also includes a ''neural approximator'' for the geometry-based operator (the rank- $r$ inverse Laplacian) and a ''kernel-PCA embedding'' for large-scale feasibility.
    \end{itemize}
\end{itemize}

We can then articulate an "activation bridge" between the symbolic and subsymbolic networks:

\begin{itemize}
  \item At each layer, the discrete block produces some symbolic or partial symbolic representation (like a pattern of rewrite-rule matches or a distribution over discrete symbols). The continuous block produces a hidden vector.
  \item Cross-Links allow ephemeral states to pass from discrete to continuous (and vice versa). For instance, the continuous side might produce a continuous embedding that the discrete side interprets as a set of ''tokens'' or ''sub-graphs,'' and the discrete side might produce rewrite-rule ''evidence'' or ''token transformations'' that feed back into the continuous layer's next iteration.
\end{itemize}

We can also deploy a shared probability model, in which both sides maintain or reference a shared distribution $p(\mathbf{x})$ or partial distribution across states. The measure-dependent geometry (Wasserstein distance, or discrete-continuous analog) can unify how we measure ''cost'' or ''transport'' in rewriting and in continuous updates.

But how does this support synergetic next-token prediction?

\subsection{Next-Token Prediction with RL Refinement}

\paragraph{Continuous Transformer's Usual Workflow}
\begin{itemize}
  \item Input tokens enter a standard multi-head attention pipeline.
  \item The model predicts the next token distribution (in continuous embedding space).
  \item RL signals can come from user feedback or environment tasks, guiding an ActPC-based local reward-laced error minimization.
\end{itemize}

\paragraph{Discrete Rewrite-Rule Side}

\begin{itemize}
  \item They also see or transform the sequence of tokens at a symbolic level.
  \item They maintain rewrite rules that describe how tokens might evolve or unify symbolically.
  \item Discrete ActPC attempts to minimize mismatch between the predicted discrete rewrite outcomes and observed tokens, plus any reward signals.
\end{itemize}

\paragraph{Neural/Symbolic Synergy}
\begin{enumerate}
  \item Ephemeral Few-Shot in Activation Space:

\begin{itemize}
  \item The continuous side can do ephemeral chain-of-thought in continuous embeddings.
  \item The discrete side can do ephemeral rewriting sequences, unrolling partial transformations.
\end{itemize}

\item Shared Partial Solutions:
\begin{itemize}
  \item If the discrete rewriting proposes a certain next-token pattern that yields a lower mismatch or better reward, it can pass that back to the continuous side, shaping the continuous layer's ephemeral hidden states.
  \item If the continuous side finds a strong local embedding that implies a certain rule sequence is beneficial, it can prime the discrete rewrite rules.
\end{itemize}

  \item Unified RL:

\begin{itemize}
  \item A single reward function merges the success/failure signals from next-token performance or environment tasks. Both discrete and continuous sides see local ''actPC-style'' signals, leading them to converge in a cooperative manner.
\end{itemize}

\end{enumerate}

\subsection{Shared Activation Spreading and Few-Shot Adaptation}

There is also a fairly natural story regarding how few-shot learning in the activation space may work synergetically across the symbolic and subsymbolic sides.

\paragraph{Joint Ephemeral States} When a prompt or new tokens arrive:

\begin{enumerate}
  \item Continuous Sub-Network quickly forms ephemeral activation patterns representing partial chain-of-thought.
  \item Discrete Sub-Network tries rewriting the token sequence or partial symbolic states to incorporate the new context.
  \item Bridges: The ephemeral states from each side are shared (like continuous embeddings and discrete sub-graph rewrites), so each side can glean insight from the other's ''in-context solution.''
\end{enumerate}

\paragraph{Enhanced Few-Shot} Because each side is capable of ephemeral adaptation, the synergy might yield:

\begin{itemize}
  \item More robust ephemeral solutions: The discrete rewrite side can quickly see if certain symbolic transformations help the continuous side converge. The continuous side can produce partial embeddings that help the discrete side skip certain rewriting steps.
  \item Possibly more stable or explainable ephemeral solutions: the discrete rules can be read as symbolic chains-of-thought, the continuous states as subsymbolic embeddings.
\end{itemize}

\subsection{Shared Information Geometry \& Online Weight Updates}

Due to the local learning implicit in the general ActPC methodology, both the discrete rewrite rules and the continuous neural parameters can get updated online:

\begin{itemize}
  \item Discrete: The measure-dependent Laplacian might define how to shift rewrite rule probabilities or partial conditions in a geometry-aware manner.
  \item Continuous: As before, the neural net does small steps guided by the approximate Laplacian in parameter space.
\end{itemize}

This updating can leverage a shared metric and joint probability distribution.   We can define a combined distribution $\mathcal{P}=\mathcal{P}_{\text {discrete }} \times \mathcal{P}_{\text {continuous }}$, and based on this the measure-dependent operator capturing cost-of-transport in the joint space might be factorized or approximated. This yields a single geometry that:

\begin{itemize}
  \item Encourages consistent changes in discrete rule probabilities and continuous weights when they refer to the same tokens or tasks.
  \item Could unify the local error signals in a single operator or ''block diagonal'' approach.
\end{itemize}

\subsection{Potentials for New Cognitive Frontiers}

In this sort of hybrid architecture, because each side can do local real-time updates, if the agent enters a ''deliberative mode,'' ephemeral solution attempts might produce partial stable error signals that trigger small rewrites or small continuous weight changes, bridging symbolic manipulations and subsymbolic pattern adjustments seamlessly.

Basic advantages might include:

\begin{itemize}
  \item The discrete rewriting can handle more ''structured'' transformations (like rewriting a formal expression or chunk of code), while the continuous side handles fuzzy pattern recognition or natural language embeddings.
  \item Shared ephemeral states let them cross-pollinate in the same session, possibly leading to more ''logical'' or ''coherent'' chain-of-thought.
\end{itemize}

Deeper potentials might also exist:

\begin{itemize}
  \item The agent can pass partial symbolic reasoning steps to the continuous side, which checks them in a distributed embedding approach. Or vice versa, the continuous side proposes approximate expansions that the symbolic side tries to refine or confirm.
  \item Real-time learning in both sub-networks can lock in effective sub-routines discovered during ephemeral reasoning.
  \item If the discrete side's rewriting rules discover a new pattern that yields reward, local geometry-based updates can quickly incorporate it into the continuous side's attention patterns.
  \item If the continuous side sees an image or environment data, it forms embeddings that help the discrete rewrite rules expand or unify symbolic content.
\end{itemize}

These capabilities could well greatly decrease the need for external "chain of thought" prompting to drive series of queries:

\begin{itemize}
  \item Because discrete rewriting can handle ''structured text'' or ''graph transformations'' directly, and the continuous side can store large-scale patterns, we get a synergy that might reduce reliance on external retrieval or fragile chain-of-thought prompting.
  \item Over repeated exposures, learned link weights in both sides embed the newly discovered patterns. The system's ''cognitive kernel'' becomes a self-contained synergy of symbolic \& subsymbolic knowledge.
\end{itemize}

\subsection{Recap}

We have described an "ActPC-GeomChem transformer-like architecture" that combines key aspects of ActPC-Geom and ActPC-Chem into a shared neural-symbolic transformer architecture with a number of potential advantages including:

\begin{itemize}
  \item More flexible, ''deliberative'' synergy: ephemeral in-context solutions that unify symbolic rewriting with fuzzy pattern embeddings.
  \item A single integrated system that can do real-time online learning in discrete and continuous forms, bridging symbolic logic and neural pattern recognition.
  \item Over time, repeated synergy locks in newly discovered solutions, reducing reliance on external retrieval or repetitive prompting and fostering genuinely integrative symbolic-subsymbolic intelligence.
\end{itemize}

What we have described here in the transformer context can also be taken as an exemplar for bringing ActPC-Geom and ActPC-Chem together in the context of other sorts of neural architectures.

Many details remain to be resolved, and much of this can only happen via experiment.  Conceptually, though, the simple key points underlying these details are that

\begin{itemize}
\item ActPC enables tight unification of neural and symbolic components of a hybrid systems via common entrainment in the same predictive coding based learning dynamic
\item Information geometry modeling has potential to make this unification efficient via building shared probabilistic models across the neural and symbolic portions of a hybrid neural-symbolic system
\end{itemize}

\noindent These simple points are what, we believe, will enable the complex details to resolve in highly functional and efficient ways.

\section{From Compositional Hypervectors to System-Level ActPC Compositional Reasoning}

Let us now connect a few of the multiple points made in the preceding sections.   Suppose we use hypervectors with a compositional algebra to form a compressed representation of ActPC system states, so as to enable passably efficient neural approximation of the inverse measure-dependent Laplacian needed to provide information-geometric guidance of ActPC learning.   Suppose we do this within an ActPC-Geom based transformer network (with or without ActPC-Chem integration, but we can leave this as a side point for the moment)?   How might the compositional, symbolic algebra of the embedding hypervectors manifest itself in the overall behavior of the transformer network?   

In this section, we will try to work out a specific "toy example" in this direction, a sort of sustained ''thought experiment."   

Suppose we have an ActPC-Geom-based transformer answering a question like ''The home country of the sport associated with Giorgio Chinaglia is'' (whose answer is UK even though Chinaglia is Italian).  

How might the compositional algebra involved in the hypervector embedding deep in the guts of the ActPC-Geom network help guide the overall ActPC based transformer system to answer the question in a way that reflects understanding of the compositionality of the sentence, rather than answering ''Italy'' incorrectly. 

Let's assume (for sake of the thought experiment) that the hypervectors include the property decomposition ''Giorgio\_Chinaglia = sport\_played x soccer + home\_country x UK + ...''), so the question is, conceptually, how this works its way into the actual behavior of the ActPC-Geom transformer-like network.

Two caveats should be obvious here:

\begin{itemize}
\item The actual dynamics of a real network of this nature will almost surely be different from any toy thought-experiment run-through
\item Modern large LLMs like GPT4 get this sort of question right, but earlier models like GPT3 got it wrong. 
\end{itemize}

\noindent Our hypothesis is that ActPC-Geom based transformers could get this sort of question right in different ways from how GPT4 does, and in a way that then also allows them to get harder compositionality-based problems right, including those which GPT4 and o1/o3 etc. cannot get.  But for now we just want to flesh out a simple example of how all the parts of this proposed system might work together, on a conceptual basis.

\subsection{Set-Up}

\paragraph{The Entities and Their Compositional Hypervectors}  We suppose that in this neural-symbolic ActPC system, names and concepts (like ''Giorgio Chinaglia,'' ''soccer,'' ''home country,'' ''Italy,'' ''UK,'' etc.) each have hypervector embeddings that allow them to be bound or composed:

\begin{itemize}
  \item Giorgio\_Chinaglia might be internally decomposed as something like:\\
$\mathbf{v}_{\text {Chinaglia }}=\operatorname{Bind}\left(\mathbf{v}_{\text {ChinagliaName }}, \operatorname{Bundling}\left(\mathbf{v}_{\text {Nationality }}:\right.\right.$ Italian, $\mathbf{v}_{\text {Sport }}:$ Soccer $\left.)\right)$\\
But in our scenario, we also note a property ''HomeCountry(sport)'' is the UK for soccer. So the system's knowledge might store:

$$
\mathbf{v}_{\text {Soccer }}=\operatorname{Bundling}\left(\mathbf{v}_{\text {HomeCountry }}: \mathbf{v}_{\mathrm{UK}}, \mathbf{v}_{\text {sportType }}: \mathbf{v}_{\text {BallGame }}, \ldots\right)
$$
\end{itemize}

\noindent (We are ignoring the numeric details for now, but we assume the ''compositional algebra'' in hypervectors can represent that soccer is a ball game, whose home country is the UK, historically, etc.)

\paragraph{Kernel PCA + Expansion Mechanism}  We assume that the transformer-like system uses a kPCA + expansion approach to unify discrete and continuous features. For example:

\begin{enumerate}
  \item Discrete aspects: ''Nationality = Italian,'' ''Home country of soccer = UK,'' etc. are turned into a kernel-based similarity.
  \item Continuous aspects: numeric attributes, embeddings of ''Chinaglia's stats,'' etc.
  \item Hypervector expansions: random-lift or binding/bundling that yields a single high-dimensiona vector for each concept or statement. 
\end{enumerate}

\noindent Hence: The system's memory has a representation that effectively encodes:

\begin{itemize}
  \item ''Chinaglia $\mapsto$ (Nationality = Italy, Sport = soccer, ...),''
  \item ''Soccer $\mapsto$ (HomeCountry = UK, ...).''
\end{itemize}

\paragraph{The ActPC-Based Transformer}  We consider a transformer that does multiple micro-iterations per layer, in a predictive-coding style:

\begin{enumerate}
  \item Local Expansions: The layer sees the partial input (the question ''The home country of the sport associated with...?''), enumerates possible next states or partial property combinations.
  \item Geometric Analysis: The system tests these expansions for consistency with the knowledge in the network (via a measure-based geometry that says ''Which expansions reduce predictive error?'').
  \item Iterate: Over a few sub-iterations, the layer refines the internal representation, eventually forming a stable set of activation patterns that reflect ''Chinaglia $\rightarrow$ soccer $\rightarrow$ home country is UK.''
\end{enumerate}

\paragraph{Information-Geometry Acceleration}  Each micro-iteration is guided by a measure-based approach that references the ''distance'' in hypervector space to states known to yield correct predictions. If partial cues suggest ''Chinaglia is soccer,'' the system checks: ''Which expansion leads to minimal error if the question is about the home country of that sport?''

Possibly, via this approach, the network quickly identifies ''Soccer's home is the UK'' as the minimal-error path.

\subsection{Step-by-Step: ''Why Not Italy?''}

Question: ''The home country of the sport associated with Giorgio Chinaglia is...?''

Naive Response: If the system fixated on ''Chinaglia is Italian,'' it might guess ''Italy.'' 

But let's see how the compositional hypervector approach avoids that:

\begin{enumerate}
  \item Input: The question text arrives. The question tokens embed into the first transformer layer. The network sees references to ''Chinaglia'' and ''the sport associated with him.''
  \item Local Expansions: The layer tries partial expansions:

\begin{itemize}
  \item Possibly one expansion says: ''Nationality(Chinaglia) = Italy $\rightarrow$ maybe the answer is Italy?''
  \item Another expansion says: ''Wait, the question is about the sport's home country, not the person's home country. Let me see: sport(Chinaglia) = soccer. Then home\_country\_of(soccer).''
\end{itemize}

\item Compositional Overlap:
\begin{itemize}
  \item The hypervector representation for ''Chinaglia'' includes a binding that ''Chinaglia $\rightarrow$ soccer,'' and ''soccer $\rightarrow$ (HomeCountry=UK).''
  \item This is discovered when the partial overlap ''(Chinaglia, soccer part) \textbackslash mapsto \textbackslash mathbf\{v\}\{\textbackslash text\{Chinaglia\}\}lodot\textbackslash mathbf\{v\}\{\textbackslash text\{sport\}\}\$...'' leads the system to unbind or read out ''soccer's property homeCountry=UK.''
  \item If the aggregator tries the path ''Chinaglia's nationality = Italy => answer=Italy,'' the partial order or measure-based geometry might produce higher error because the question specifically says ''the home country of the sport.'' The local mismatch is large if we conflate the person's nationality with the sport's home.
\end{itemize}

\item  Predictive Error Minimization:
\begin{itemize}
  \item The system checks: ''If I produce the answer = 'Italy', do I reduce mismatch with known facts?'' Possibly it sees a bigger error: the text says 'home country of the sport,' not the person.
  \item The geometry or partial-order aggregator then discards the ''Italy'' path as it fails to reduce error.
\end{itemize}

\end{enumerate}

According to this narrative: Within a few micro-iterations in the transformer block, the emergent stable solution is ''UK,'' because that partial alignment yields minimal mismatch and is consistent with the ''soccer => UK'' property in the hypervector-coded memory.

\subsection{The Role of the Learned Hypervectors}

What's happening behind the scenes in this example would include, presumably, the following:

\paragraph{Compositional Binding}. Implementation might store something like:

\begin{itemize}
  \item $\mathbf{v}_{\text {Chinaglia }} \approx \operatorname{Bind}\left(\mathbf{v}_{\text {NameChinaglia }}, \mathbf{v}_{\text {sport }}: \mathbf{v}_{\text {Soccer }}, \ldots\right)$.
  \item $\mathbf{v}_{\text {Soccer }} \approx$ Bundling $\left(\mathbf{v}_{\text {HomeCountry }}: \mathbf{v}_{\mathrm{UK}}, \ldots\right)$.
\end{itemize}

\noindent When the question ''(Chinaglia => sport => home country => ?)'' arrives, partial overlap in the hypervector domain triggers or retrieves the ''UK'' portion. Because the partial overlap reveals ''Chinaglia => soccer => UK,'' the network's aggregator sees that bridging is less error than any bridging to ''Italy.''

\paragraph{Kernel-PCA-like geometry}  The system might do partial dimension reduction of these hypervectors, so that e.g.\\
$\left\|\mathbf{v}_{\text {Chinaglia }}-\mathbf{v}_{\text {ChinagliaAndSoccer }}\right\|$ is small, while ''Chinaglia => Italy'' is also recognized but flagged as less relevant for ''sport's home.'' The geometry ensures that ''Chinaglia + soccer => UK'' sits in a region that yields the minimal predictive error for the question.

\subsection{Big Picture: ''Understanding'' vs. ''Guessing''}

Contrasting a simplistic traditional transformer dynamic with the neural-symbolic compositionality-based dynamic we're after here, we may say:

\begin{itemize}
  \item A naive ''GPT-3 style'' approach might wrongly focus on the nationality of the person, because it is following attention flows without accounting for the underlying structural semantics of the entities being attended
  \item Larger and more advanced transformers overcome this particular sort of simple small-scale error via leveraging a larger number of larger combinations to focus their attention, so e.g. their training explicitly accounts for various cases where people are associated with different countries as regards different roles or aspects
  \item The ActPC-based approach with hypervector compositional embeddings would (we conjecture) avoid the naive error in a different way -- via being more explicitly structured, letting partial pattern ''Chinaglia => sport => soccer => home country => UK'' pop out.
  \item Over repeated micro-iterations, the system would arrive at ''UK'' because that compositional route is the one that truly yields minimal mismatch with all sub-part knowledge stored in the hypervector-coded memory.
\end{itemize}

Hence: The system ''understands'' the difference between the person's nationality vs. the sport's home country. The compositional representation should, we believe, be able to foster correct responses for more complicated chain-of-thought queries, where partial aspects of the question highlight subchunks in the memory.

So ... what could possibly go wrong?  Well -- for example -- 

\begin{enumerate}
  \item In a real large LLM: The actual code vectors are enormous, and the network's gradient updates are extremely complex. The story above is just a toy example.  Real dynamics may deviate in ways we can't predict from these toy considerations.
  \item Dimension: The hypervector approach might require an overly large dimension for robust binding, incurring unrealistic cost.
  \item No Guarantee: If the training data never gave enough negative examples (like ''a person from one place playing a sport from another'' or at least other more laterally related examples), the system might still conflate them.
\end{enumerate}

But conceptually, it does seem the proposed apporach can yield a more compositional ''internal route'' to the correct answer.  

As we have seen, in this toy scenario:

\begin{itemize}
  \item The compositional hypervector embeddings store, for Chinaglia, both ''Italian nationality'' and ''soccer => UK'' in the same high-dim representation.
  \item The ActPC-based transformer uses predictive coding micro-iterations and a measurebased aggregator to check expansions.
  \item The partial question ''home country of the sport associated with $X$ '' triggers the aggregator to re-activate ''soccer => UK,'' discarding ''Italy'' because that route is
\end{itemize}

Hence, the question is answered in a way reflecting deeper compositional reasoning, rather than superficial pattern association or ''naive immediate confusion with the nationality of the person.''

Going beyond toy to reality will complexify things and place heavy demands on approximations enabling scalability, but will also provide systems with massive amounts of data to bring to bear on indirectly related questions, leveraging the many powerful generalization mechanisms ActPC-Geom has at its disposal.

\subsection{Digging Further into Binding and Bundling}

Fleshing out our toy example a little further, we now speculate a bit about how the binding and bundling operations in the hypervector algebra might work so as to enable the dynamics suggested above.

(Of course -- in line with the caveats already given above -- the details presented are conceptual rather than an attempt to spell out precisely what would happen in a fully implemented approach -- real high-dimensional embeddings can be more complicated and will need to be refined via experiment.)

\subsubsection{Representation \& Storage}

\paragraph{Representing ''Chinaglia => soccer => homeCountry=UK''} Let's denote the following base vectors:

\begin{itemize}
  \item $\mathbf{v}_{\text {Chinaglia }}, \mathbf{v}_{\text {Soccer }}, \mathbf{v}_{\text {Italy }}, \mathbf{v}_{\text {UK }}$ for these entities or values.
  \item $\mathbf{v}_{\text {Nationality }}, \mathbf{v}_{\text {HomeCountry }}, \mathbf{v}_{\text {AssociatedSport }}$ for these properties.
\end{itemize}

\paragraph{Storing a ''Person => Sport'' Fact} To store ''Chinaglia's sport is soccer,'' we might define an operation like:

$$
\mathbf{v}_{\text {Chinaglia }} \rightarrow_{\text {soccer }}=\operatorname{Bind}\left(\mathbf{v}_{\text {Chinaglia }}, \operatorname{Bind}\left(\mathbf{v}_{\text {AssociatedSport }}, \mathbf{v}_{\text {Soccer }}\right)\right)
$$

\noindent or in a simpler ''two-argument binding,'' something like:

$$
\mathbf{v}_{\text {Chinaglia }} \rightarrow_{\text {soccer }}=\operatorname{Bind}\left(\mathbf{v}_{\text {Chinaglia }}, \mathbf{v}_{\text {Sport:IsSoccer }}\right)
$$

\noindent where $\mathbf{v}_{\text {Sport:ISSoccer }}=\operatorname{Bind}\left(\mathbf{v}_{\text {AssociatedSport }}, \mathbf{v}_{\text {Soccer }}\right)$.

\paragraph{Storing ''Soccer => HomeCountry=UK''} Similarly:

$$
\mathbf{v}_{\text {Soccer }} \rightarrow \mathrm{UK}=\operatorname{Bind}\left(\mathbf{v}_{\text {Soccer }}, \operatorname{Bind}\left(\mathbf{v}_{\text {HomeCountry }}, \mathbf{v}_{\mathrm{UK}}\right)\right)
$$

\noindent So, in the system's memory (some large, possibly distributed store of hypervectors), we keep these ''fact vectors" for diverse utilization.

\subsubsection{The ''Question'' Partial Overlap in Hypervector Space}

Question: ''The home country of the sport associated with Giorgio Chinaglia is...?''

At some stage, the system forms or bundles partial cues from the question.  For instance, we have:

\begin{itemize}
  \item ''Chinaglia''
  \item ''Associated sport''
  \item ''Home country''
\end{itemize}

\noindent and we want to see which known facts in memory overlap well with these partial cues.

\paragraph{Partial Binding of the Question}. One might produce a question vector:

$$
\mathbf{v}_{\mathrm{Q}}=\text { Bundle }\left(\mathbf{v}_{\text {Chinaglia }}, \mathbf{v}_{\text {AssociatedSport }}, \mathbf{v}_{\text {HomeCountry }}\right)
$$

\noindent (plus additional bits for ''question structure,'' etc.). Then the aggregator or measure-based geometry tries to find which ''fact vectors'' in memory have a strong overlap or small distance from $\mathbf{v}_{\mathrm{Q}}$.

\paragraph{Overlap with ''Chinaglia to Italy'' vs. ''Soccer => UK''}  If we also had a vector $\mathbf{v}_{\text {Chinaglia }} \rightarrow_{\text {Italy }}=\operatorname{Bind}\left(\mathbf{v}_{\text {Chinaglia }}, \operatorname{Bind}\left(\mathbf{v}_{\text {Nationality }}, \mathbf{v}_{\text {Italy }}\right)\right)$, it might not overlap strongly with the partial property ''(Chinaglia => sport => homeCountry).'' Because Nationality $\neq$ HomeCountryOfSport.

Meanwhile, the vector for ''(Chinaglia to soccer) + (soccer to UK)'' collectively is close to or consistent with ''(Chinaglia + associatedSport + homeCountry).'' The aggregator sees a higher dot product or smaller measure-based error.

Hence, in the measure-based aggregator step, the ''Chinaglia => soccer => UK'' path has a better overlap or yields less predictive mismatch for ''(Chinaglia's sport => home country?),'' so the system selects ''UK.''

\paragraph{ Info-Geometry Acceleration in Micro-Iterations}  During each micro-iteration in the ActPC-based transformer:

\begin{enumerate}
  \item The system tries expansions such as ''Chinaglia => nationality=Italy => (??),'' ''Chinaglia => sport=soccer => homeCountry=??,'' etc.
  \item Each expansion is turned into a new hypervector $\mathbf{v}_{\text {candidate. }}$. The aggregator or measure-based geometry references the question vector $\mathbf{v}_{\mathrm{Q}}$.
  \item If the dot product $\left\langle\mathbf{v}_{\text {candidate }}, \mathbf{v}_{\mathrm{Q}}\right\rangle$ is higher, or distance $\left\|\mathbf{v}_{\text {candidate }}-\mathbf{v}_{\mathrm{Q}}\right\|$ is smaller, then that candidate is more consistent with the partial cues.
  \item ''Chinaglia => nationality=Italy => homeCountry=??'' might not unify well with ''sport => homeCountry,'' so that partial route yields higher mismatch.
  \item ''Chinaglia => soccer => UK'' has bigger overlap with $\mathbf{v}_{\mathrm{Q}}$ because ''soccer => homeCountry=UK'' matches exactly the property ''(homeCountry(sport(Chinaglia)))=?.''
\end{enumerate}

\noindent After a few steps, the system ''collapses'' to ''UK'' as the best final answer that matches the question structure.

\subsubsection{Binding and Bundling}  So why does binding help here?

\begin{itemize}
  \item If ''Chinaglia => soccer'' is a single bound vector, then any mention of ''Chinaglia's sport'' plus partial mention of ''soccer => homeCountry=??'' will ''unlock'' or ''guide'' the aggregator to look for the next piece: ''soccer => UK.''
  \item If we used a purely additive or purely continuous embedding, we might not have as strong a combinational or invertible structure to parse out that ''Chinaglia => soccer => homeCountry=UK.''
\end{itemize}

\noindent Hence: The partial overlap or partial unbinding in hypervectors is key to ''compositional'' retrieval.

\subsubsection{Summarizing the Speculative Mechanism}

The hypervector algebra operations have clear meaning in the toy example, as we have seen:

\begin{enumerate}
  \item Binding for facts: ''( $\mathrm{A}=>\mathrm{B}=>$ property=Value)'' is stored as a hypervector.
  \item Bundling for questions: The question states partial properties (''Chinaglia => sport => homeCountry=?).''
  \item Partial Overlap or measure-based aggregator: Compares the question's hypervector to stored or newly expanded state vectors.
  \item Minimal Error leads to concluding ''the sport is soccer => soccer's home is the UK,'' not ''Italy.''
\end{enumerate}

\noindent Thus, conceptually:

\begin{itemize}
  \item Binding keeps the associations ''Chinaglia => soccer'' and ''soccer => UK'' in invertible form.
  \item Bundling merges partial cues from question.
  \item Overlap or measure-based geometry identifies the correct chain.
  \item Predictive-coding micro-iterations use the partial match to reduce error, preferring expansions that complete the partial pattern in a logically consistent way—leading to ''UK'' as final answer.
\end{itemize}

In this way, the binding/bundling operations in the hypervector algebra are used specifically to store and retrieve composite knowledge like ''(Chinaglia => soccer => homeCountry=UK).'' 

When the question ''(Chinaglia => sport => homeCountry=??)'' arises, partial match and unbinding push the measure-based aggregator to retrieve ''(soccer => UK),'' discarding ''(Chinaglia => Italy).'' Over a few micro-iterations, the ActPC-based transformer converges on the correct completion ''UK.''  So the compositional algebra in hypervector space helps guide the overall network to an answer that acknowledges the difference between the person's nationality and the sport's home origin—all driven by the partial matches and error minimization in predictive coding.

\subsection{Dissecting Aggregator Functionality}

It is interesting to dissect the guts of this hypothesized cognitive dynamics a little further, and ask how exactly the proposed "aggregation" function might function internally in hypervector space, comparing a ''question vector'' to stored or newly generated ''fact vectors.'' 

We focus here on telling a story explaining why partial overlap or small $\ell_{2}$ distance in hypervector space translates to ''this new expansion is good,'' and how the aggregator decides which expansions to keep or discard.

\subsubsection{Recap of Set-Up: A Hybrid ActPC Network with a ''Measure-Based Aggregator''}

\begin{enumerate}

  \item ActPC Loop:

\begin{itemize}
  \item The network has iterative micro-steps, each generating expansions (possible next states, partial solutions, or ''fact vectors'').
  \item The aggregator collects these expansions, measures their ''predictive cost'' or ''distance from optimum,'' and discards the ones that do not reduce error.
\end{itemize}

  \item Hypervector Representations:

\begin{itemize}
  \item Each concept or partial solution is encoded in a high-dimensional vector, possibly with binding/bundling.
  \item The aggregator's ''cost'' or ''error'' is often computed as some distance in that high-dimensional space: e.g., an $\ell_{2}$ distance to a relevant query or question vector
\end{itemize}

\item Aggregator's Input \& Output

\begin{itemize}
  \item Input: 
  
  \begin{itemize}
  \item A set of newly proposed expansions:

$$
\left\{\mathbf{v}_{\text {candidate }, 1}, \mathbf{v}_{\text {candidate }, 2}, \ldots, \mathbf{v}_{\text {candidate }, k}\right\}
$$

\item along with:

\begin{itemize}
  \item The ''context vector'' or ''question vector'' $\mathbf{v}_{\mathrm{Q}}$, summarizing the partial cues from the question or from the current ActPC state.
  \item Potentially, a set of previously accepted expansions or partial solutions stored in memory.
\end{itemize}

\end{itemize}

  \item Output: A refined set of expansions that reduce the overall predictive error. Possibly the aggregator produces the single best expansion or a small subset for the next iteration.
  \end{itemize}
\end{enumerate}

\subsubsection{Internal Aggregator Steps}  So how might this work internally?

\paragraph{Evaluate Overlap or Distance}.  For each candidate $\mathbf{v}_{\text {candidate }, i}$ , we compute a measure of ''match'' or ''error'' relative to $\mathbf{v}_{\mathrm{Q}}$. In accordance with standard practice in hypervector systems, we might define:

\begin{itemize}
  \item A dot product:

$$
\operatorname{score}_{i}=\mathbf{v}_{\mathrm{Q}} \cdot \mathbf{v}_{\text {candidate }, i}
$$

(the higher, the better overlap), or

  \item An $\ell_{2}$ distance:

$$
\operatorname{dist}_{i}=\left\|\mathbf{v}_{\mathrm{Q}}-\mathbf{v}_{\text {candidate }, i}\right\|
$$

\noindent (the smaller, the better match).

  \item Or more elaborate measure-based approaches, e.g. ''Wasserstein-like'' if we interpret each hypervector as a point distribution. But dot product or $\ell_{2}$ is typically simpler for high-dimensional vectors.
\end{itemize}

\paragraph{Sort or Threshold} The aggregator then

\begin{itemize}
  \item Sorts the candidates by ascending distance or descending overlap.
  \item Potentially discards all candidates whose score is below a threshold or whose distance is above a threshold. The top few might remain as plausible expansions.
\end{itemize}

\paragraph{Combine With Known Memory} If the aggregator keeps track of ''fact vectors'' from long-term memory (like ''Chinaglia => soccer => UK,'' ''Chinaglia => Italy,'' etc.), it can also compare $\mathbf{v}_{\text {candidate, } i}$ with each memory vector. If a candidate expansion is redundant or strictly worse than an existing memory vector, aggregator discards it. For instance:

\begin{itemize}
  \item If ''Chinaglia => soccer => UK'' is already in memory as a vector $\mathbf{v}_{\text {Chinaglia2Soccer2UK, }}$ a newly proposed vector that tries ''Chinaglia => Italy => ???'' might be overshadowed if it leads to higher mismatch with the question vector.
\end{itemize}

\paragraph{Possibly Update Weights or Next Iteration}. After picking the best expansions, the aggregator signals to the ActPC loop:

\begin{itemize}
  \item ''Keep expansions 1 and 3, discard the rest.''
  \item The network then does the next micro-step focusing on expansions 1 or 3, refining them further if needed.
  \item The aggregator might also do partial learning, e.g. adjusting some local measure-based weights if we want to re-scale certain property bindings.
\end{itemize}

\paragraph{The Example with ''Chinaglia => soccer => UK''}. Then, given
\begin{itemize}
  \item Question Vector $\mathbf{v}_{Q}$. Summarizing ''Chinaglia => associatedSport => homeCountry => ???''.
  
\end{itemize}

\noindent we have

\begin{enumerate}
  \item Candidate: ''Chinaglia => nationality=Italy => ???,'' coded as $\mathbf{v}_{\text {Chinaglia2Italy }}$ 
  
  \begin{itemize}
  \item We measure:

$$
\operatorname{dist}\left(\mathbf{v}_{Q}, \mathbf{v}_{\text {Chinaglia2Italy }}\right)
$$

\noindent or the dot product. 
\item Likely a moderate match but not strong for the ''homeCountry(sport(Chinaglia))'' idea.
\end{itemize}
\item Candidate: ''Chinaglia => soccer $=>$ UK,'' coded as $\mathbf{v}_{\text {Chinaglia2Soccer2UK}}$

\begin{itemize}
\item We measure: 

$$
\operatorname{dist}\left(\mathbf{v}_{Q}, \mathbf{v}_{\text {Chinaglia2Soccer2UK }}\right)
$$

\noindent or dot product. 
\item This is presumably lower distance or higher overlap, because partial cues ''Chinaglia => associatedSport => soccer => homeCountry => UK'' line up neatly with the question vector's components.
\end{itemize}
\end{enumerate}

Hence the aggregator:

\begin{itemize}
  \item Accepts $\mathbf{v}_{\text {Chinaglia2Soccer2UK. }}$
  \item Rejects $\mathbf{v}_{\text {Chinaglia2Italy }}$.
\end{itemize}

\subsubsection{Why Binding/Unbinding is Key}

If the aggregator tries to see if $\mathbf{v}_{\text {Chinaglia2Soccer2UK }}$ explains the partial question, it might do partial unbinding.

We could define an unbinding step like:

$$
\mathbf{v}_{\text {Chinaglia2Soccer2UK }} \odot \operatorname{Bind}\left(\mathbf{v}_{\text {Chinaglia }}^{-1}, \mathbf{v}_{\text {AssociatedSport }}^{-1}\right)
$$

\noindent to see what remains. Possibly that yields something close to $\operatorname{Bind}\left(\mathbf{v}_{\text {HomeCountry }}, \mathbf{v}_{\mathrm{UK}}\right)$.

\begin{itemize}
  \item If so, the aggregator sees it's consistent with ''HomeCountry => UK,'' which is what the question asks for.
  \item By contrast, unbinding with the property ''Chinaglia => nationality => ???'' might yield nonsense or a large mismatch for the question's property slots, so it's not chosen.
\end{itemize}

\subsubsection{Summary of Aggregation Process}

In sum, according to the narrative we unfold here for this toy problem, the aggregator's steps might look like:

\begin{enumerate}
  \item Compute the overlap/distance of each candidate with the question vector $\mathbf{v}_{Q}$.
  \item (Optionally) attempt partial unbinding to see if it yields the needed property ''homeCountry(sport(Chinaglia)).''
  \item Select the candidate that yields minimal mismatch. ''(Chinaglia => soccer => UK).''
  \item Discard less-fitting expansions like ''Chinaglia => Italy.''
  \item Return the final ''UK'' as the correct fill for ''home country of the sport'' or proceed to the next iteration with that candidate.
\end{enumerate}

Thus the aggregator, referencing ''strong overlap or small distance,'' leverages binding/bundling to find the correct chain, thus guiding the ActPC-based system toward the right answer.

\subsection{Intelligent Search of Multiple Partial Unbindings}

This in-depth analysis of the aggregation process suggests interesting possible extensions.   One wonders if there could be a role for exploring multiple partial unbindings of different combinations of vectors in parallel -- so that the aggregator would then be doing a sort of heuristic search process on potential unbindings, trying to find a good match?   This would surface a sort of traditional AI exploration process in the guts of the application of compositional hypervector algebra to geometric neural net guidance.

\paragraph{Why Parallel Partial Unbinding?}.   If we store ''Chinaglia => soccer => homeCountry =UK'' as a single hypervector (via nested or chained binding/bundling), that single vector might look like

$$
\mathbf{v}_{\text {Chinaglia2Soccer2UK }}=\operatorname{Bind}\left(\mathbf{v}_{\text {Chinaglia }}, \operatorname{Bind}\left(\mathbf{v}_{\text {AssociatedSport }}, \ldots\right), \ldots\right)
$$

We may then find partial overlaps: A question vector $\mathbf{v}_{Q}$ (like ''Chinaglia => sport => homeCountry=??'') might partially overlap with that big bound vector.

However, the plot may be thicker than this: Realistically, we could also have \emph{other} expansions or facts, each of which is also a large bound hypervector. The aggregator needs to figure out which expansions best match $\mathbf{v}_{Q}$.

\paragraph{Why Attempt Multiple Unbindings?} There could be multiple reasons:

\begin{itemize}
  \item One Candidate might match partial cues but not the rest.
  \item Another Candidate might also partially match, but if we do a small unbinding step in a different combination of property vectors, we discover a better match.
  \end{itemize}
  
By exploring multiple partial unbindings in parallel, we do a local search for which candidate expansions have the highest synergy with the question's structure, potentially finding better solutions that a single-step check might miss.

Hence we avoid a naive single nearest-neighbor approach and do more of a heuristic or beam search in hypervector space.

\paragraph{How the Aggregator's Search Might Work}. It's not hard to make an algorithmic sketch of a more sophisticated process:

\begin{enumerate}

  \item Initial Candidates: The aggregator receives multiple expansions $\left\{\mathbf{v}_{c_{1}}, \mathbf{v}_{c_{2}}, \ldots, \mathbf{v}_{c_{k}}\right\}$. Some might be newly generated; some might be from memory.
  
  \item Score Basic Overlap: For each $\mathbf{v}_{c_{i}}$, compute a quick ''overlap measure'' with the question vector $\mathbf{v}_{Q}$ (e.g. dot product or $\ell_{2}$ distance). If some are obviously too far off, discard them.
  
  \item Partial Unbinding Steps: For the top few expansions, attempt different combinations of partial unbinding. For example:

\begin{itemize}
  \item If $\mathbf{v}_{c_{i}}$ is known to contain ''(Entity $\rightarrow$ property $\mathrm{X} \rightarrow$ property $\mathrm{Y} \rightarrow$ property Z ),'' we might do unbinding with $\mathbf{v}_{\text {property }}$, or some combination, to see if the remainder lines up with ''homeCountry,'' ''sport,'' etc.
\end{itemize}

  \item Heuristic Checking:

\begin{itemize}
  \item Each partial unbinding yields some ''residual vector'' or ''unbound chunk.'' Compare that chunk to the question's property slots (like ''homeCountry(?).'')
  \item If it aligns well, we have a strong match for that route.
\end{itemize}

\item Search Frontier:
\begin{itemize}
  \item For each partial unbinding that yields a strong alignment, we keep it in the next iteration's search frontier.
  \item We discard or rank-lower expansions that produce only weak partial alignments.
\end{itemize}

\item Stop Condition:

\begin{itemize}
  \item If we find a route that unbinds to exactly the property needed by the question (e.g. ''(soccer => UK)''), we consider that a success.
  \item Alternatively, we might keep going until expansions converge or we run out of search budget.
\end{itemize}
\end{enumerate}

Such iterations can be done in parallel, exploring different regions of the search space:

\begin{itemize}
  \item Each expansion and partial unbinding can be assigned to a separate thread or subprocess. They all do a quick check of ''(residual overlap with question).''
  \item The aggregator collects results in a lock-free or message-based manner, merges the top results that show the best synergy.
\end{itemize}

\paragraph{ Interpreting ''Heuristic Search'' in the ActPC/InformationGeometry Paradigm}   ActPC usually frames expansions vs. local mismatch minimization.  As we have articulated, we can incorporate a search-like aggregator, which does multiple expansions concurrently, checks which yield minimal error, and focuses on those next.  How might this leverage information geometry?

 Information Geometry can help each partial unbinding or expansion update quickly reduce error. The aggregator might do a small geometric step in hypervector space for each candidate, then measure improvement, continuing or discarding.

Hence, the aggregator is no longer just ''pick best single expansion,'' but ''explore multiple partial unbindings in parallel to see which better aligns with the question's partial structure,'' akin to a local BFS or beam search in hypervector space, and leveraging  geometric information to guide its exploration.

\paragraph{Potential Gains} include

\begin{itemize}
  \item Robustness:

\begin{itemize}
  \item If one route to unbinding or partial expansions is slightly confused, another route might produce a better match.
  \item We avoid ''locking in'' on a suboptimal candidate too soon.
\end{itemize}

  \item Deeper Compositional Discovery:

\begin{itemize}
  \item Some expansions might require a multi-step chain of unbindings to reveal the property that the question demands. The aggregator's multi-route search helps discover those.
\end{itemize}

  \item Scalability:

\begin{itemize}
  \item On HPC hardware or many CPU threads, we can parallelize these partial unbinding checks.
\end{itemize}

\end{itemize}

The main downside here is:  Exploration of multiple unbindings introduces another layer of complexity into an already complex system.   However, this complexity can be at least made elegant via handling it recursively -- i.e. via implementing the heuristic search within a symbolic system such as Hyperon which is then modeled within the same distributional space as the ActPC-Geom network.

On the whole, it appears that exploring multiple partial unbindings in parallel can give the aggregator a heuristic search dimension. Instead of a single overlap check, the aggregator tries various ways to ''unpack'' the candidate hypervectors to see which route yields consistent alignment with the question or partial property specification. This can yield more powerful compositional retrieval, especially if the expansions are multi-layered (like ''Chinaglia => soccer => UK => ...''). The aggregator effectively does a small local search among possible ways to interpret or decode the bound hypervectors, enabling robust ActPC-based compositional reasoning.   

In the ideal case: We arrive at a hopefully powerful cognitive synergy within which AI search methods, potentially guided by neural methods for search control, are used to effectively deploy hypervector algebra to accelerate neural methods.

\part{Pathways to Optimized Implementation}

We have outlined a variety of speculative structures and algorithms here, and clearly the next order of business in the ActPC-Geom research direction is to implement some small-to-moderate-scale examples and see how the various aspects of the approach pan out in terms of practical experimentation.   Once things have been refined at a smaller scale, then the time will come to design and deploy large-scale implementations.

However, the role of scalability in modern AI is such that when introducing new approaches, it feels appropriate to address the route to massive scalability right from the start, at least to a limited degree.

In some ways the scaling of ActPC-Geom should come for free from the use of modern GPU-based libraries for matrix manipulations -- compared say ActPC-Chem or Hyperon, ActPC-Geom is a fairly "conventional" sort of architecture.   It is founded on neural networks, and the additional mechanisms it posits are largely based on numerical vectors or hypervectors.   All this is much less exotic than the large and general rewrite-rule networks Hyperon or ActPC-Chem deal with.

However, here we will look at ActPC-Geom scalability from some additional perspectives -- focusing in particular on hybrid architectures combining ActPC-Geom with ActPC-Chem, where neural-symbolic integration leads to novel needs and opportunities for optimization of various sorts.

We will consider two main aspects of optimization here:

\begin{itemize}
\item Achieving efficient concurrency of neural-symbolic operations via algorithmic rearrangements using Galois connection math
\item Design of customized HPC architectures for the mix of operations involved in ActPC-Geom + ActPC-Chem
\end{itemize}

\section{Efficient Concurrent Hybrid ActPC via Galois Connections}

In \cite{goertzel2021patterns} it is argued that the core concepts from Mu and Oliviera's paper "Programming with Galois Connections" \cite{mu2012programming} are broadly applicable to the optimized implementation of AI algorithms, and in particular to algorithms playing central roles in neural-symbolic AGI systems.   Examples given include evolutionary program learning and uncertain logical theorem proving.   In this section we will argue that a similar conclusion applies to the core operations of ActPC-Geom, but with particular importance for the case where ActPC-Geom is hybridized with ActPC-Chem to form a neural-symbolic PC based learning system.

An ActPC-Geom system on its own can be largely optimized via standard multi-GPU / SIMD parallel methods, though its use of symbolic processing in certain aspects such as compositional hypervector algebra means it does have a clear role for the sorts of optimization that the Galois-connection math brings.   However, once one introduces close coupling with a high volume of discrete operations via hybridization with ActPC-Chem, then one is in a domain where the Galois-connection approach potentially has more critical value for enabling efficient implementations.

\subsection{Handling Approximate Stochastic Dynamic Programming with Galois Connections}

We will apply Galois connections to ActPC networks via modeling certain of their operations as a form of approximate stochastic dynamic programming.   As argued in \cite{goertzel2021patterns}, a large percentage of AGI-relevant algorithms can be modeled in this way -- which is not terribly surprising since, for instance, reinforcement learning in itself can be viewed as a sort of approximate stochastic dynamic programming.

We begin by giving an extension of the ''Dynamic Programming Theorem'' (DPT)  from Mu and Oliviera in the setting of Galois Connections (GCs), relaxed to handle approximate, stochastic dynamic programming rather than the standard precise, deterministic DP. We first restate the original (precise) theorem, then modify each core piece -- particularly the transition relation and the fold/hylomorphism definitions -- so they can incorporate approximate ( $\varepsilon$-bounded) or probabilistic updates. Finally, we assert an analogous inclusion theorem guaranteeing that the least fixed point of the approximate system is ''contained in'' or ''close to'' the solution derived by the GC approach, with high probability or up to a small error.   This is a more rigorous version of argumentation presented more heuristically in \cite{goertzel2021patterns}.

\subsubsection{Extending Deterministic DP Theorem to the Uncertain Case} \label{sec:mu}

In the standard approach from Mu and Oliveira's framework, the Dynamic Programming Theorem states (in one phrasing):

\begin{equation*}
\mu\left(\lambda X \rightarrow\left(i n \cdot F X \cdot T^{\circ}\right) \triangleright S\right) \subseteq\left([T]^{\circ}\right) \triangleright S \tag{DPT-Exact}
\end{equation*}

\noindent where:

\begin{enumerate}
  \item $\mu(\lambda X \rightarrow \ldots)$ denotes the least fixed point of a functional that includes the fold/hylomorphism (through in $\cdot F X$ ) plus a transition relation $T^{\circ}$.
  \item (the ''shrink'' operator) expresses the ''hard'' part of a Galois connection, which narrows candidate solutions to those meeting the optimality criterion $S$.
  \item $[T]^{\circ}$ is the ''derived'' or ''fused'' fold that merges the transition relation into a single dynamic-programming-like recursion.
  \item The right-hand side $\left([T]^{\circ}\right) \downarrow S$ is the ''closed form'' or direct DP solution.
\end{enumerate}

The theorem shows that recursively constructing solutions by a fold (the left-hand side) is contained in the ''global'' solution derived by the Galois-connection approach (the right-hand side). In classical DP, both sides coincide under mild conditions (monotonicity, well-foundedness, etc.).

\paragraph{Stochastic / Approximate Modifications} If we assume system is not purely deterministic, i.e.

\begin{enumerate}
  \item The ''transition'' or ''candidate-generation'' step might yield a probabilistic or $\varepsilon$-approximate set of subproblems.
  \item The cost or evaluation step might have noise or approximate computations.
\end{enumerate}

\noindent then instead of a precise relation $T \subseteq S \times S$ (mapping state to sub-states), we may have $\widetilde{T}$ that, for each state $s$, returns a (random) set of next states $\widetilde{T}(s)$ with some probability distribution, or a set that is within $\varepsilon$-distance from the true set $T(s)$.

We can represent this as:

$$
\widetilde{T}: S \rightrightarrows \mathcal{P}(S)
$$

\noindent where '' $\rightrightarrows ''$ denotes a multi- or stochastic relation, possibly with a small ''error margin'' from the original $T$.

Similarly, the fold/hylomorphism:

$$
i n \cdot F X
$$

\noindent which enumerates subproblems, may be replaced by an approximate version:

$$
i n \cdot \widetilde{F} X
$$

\noindent where $\widetilde{F}$ is either:

\begin{itemize}
  \item A random operator, meaning $\widetilde{F} X$ yields sub-states or partial costs in a stochastic manner,
  \item An $\epsilon$-approximate operator, bounding the difference $\|\widetilde{F} X-F X\| \leq \epsilon$ for some metric $\|\cdot\|$.
\end{itemize}

\paragraph{ Extended Theorem Statement}  To extend the deterministic DP theorem to these cases, let:

\begin{enumerate}
  \item $\widetilde{T}^{\circ}$ be the converse of the approximate or stochastic relation $\widetilde{T}$.
  \item $\widetilde{F}$ be the approximate generator for the sub-hylomorphism.
  \item $S$ remain the same optimization or partial order used in the GC.
  \item remain the same ''shrink'' operator from Galois-connection-based optimization.
\end{enumerate}

\noindent We define the left-hand side functional:

$$
\tilde{f}(X):=\left(i n \cdot \widetilde{F} X \cdot \widetilde{T}^{\circ}\right)>S
$$

\noindent Now its least fixed point (or least solution in partial-order sense) is

$$
\mu(\tilde{f})=\mu\left(\lambda X \rightarrow\left(i n \cdot \widetilde{F} X \cdot \widetilde{T}^{\circ}\right)>S\right)
$$

\noindent We also define the ''fused'' or ''closed-form'' approach:

$$
\left([\widetilde{T}]^{\circ}\right) \triangleright S
$$

\noindent where $[\widetilde{T}]^{\circ}$ is the same or a very similar hylomorphism-based operator that merges $\widetilde{T}^{\circ}$ into a single pass, but now approximate or stochastic as well.

To extend the DP theorem appropriately, we want an inclusion or near-inclusion statement.   We would like to write something like

$$
\mu(\tilde{f}) \subseteq_{\epsilon}\left([\widetilde{T}]^{\circ}\right) \triangleright S
$$

\noindent meaning that:

\begin{itemize}
  \item With high probability (in a random setting), or up to an additive $\epsilon$ or multiplicative factor $\alpha$ (in an approximate setting), solutions found by the left side are contained or are not worse than the right side.
  \item The '' $\subseteq_{\epsilon}$ '' can be read as ''subset or bounding up to small error $\epsilon$.''
\end{itemize}

This can be formalized via statements in the direction of:

\begin{enumerate}
  \item Probabilistic: ''For all $\delta>0$, with probability $\geq 1-\delta$, the set of solutions from $\mu(\tilde{f})$ is contained in $\left([\widetilde{T}]^{\circ}\right) \checkmark S$, or equivalently, the approximate solution is $\epsilon$-close to the GC solution in some cost metric.''
  \item Approximate: ''If each local step is $\epsilon$-close to the original T, then the final DP solution is $\mathcal{O}(\epsilon)$-close to the precise DP solution overall.''
\end{enumerate}

In Appendix \ref{app:mu} we sketch a proof of this approximate stochastic extension of the Mu and Oliviera DP theorem.   In Appendix \ref{app:conditions} we review the key formal conditions needed to make the proof of the deterministic or uncertain versions of the DP theorem work, which in the end appear unproblematic in the context of the application to ActPC.

\subsection{ActPC as Approximate Stochastic Dynamic Programming}

We now flesh out the point that ActPC networks (particularly in RL-like scenarios) do something akin to approximate, stochastic dynamic programming.   Put simply, we have:

\begin{itemize}
  \item State: The network's internal states plus environment observations.
  \item Actions: Emergent from the network's output or policy layers.
  \item Rewards: Estimated or predicted by the network, driving local error minimization (or measure-based geometry if using Wasserstein).
  \item Approximate Updates: ActPC typically does small iterative updates that combine local mismatch signals and global reward signals in a distributed fashion, rather than monolithic, exact DP sweeps.
\end{itemize}

\noindent From a high-level, each step in an ActPC RL agent is an approximate dynamic-programming iteration, adjusting local estimates (value function or policy) in a partially stochastic manner.

\subsection{Leveraging Galois Connections}

In the Galois-connection (GC) framework, dynamic programming can be seen as a pair of adjoint operations:

\begin{enumerate}
  \item Easy Adjoint $(f)$ : Generating or enumerating candidate outcomes (sub-states, sub-actions, or next transitions).
  \item Hard Adjoint $(g)$ : ''Shrinking'' or ''optimizing'' these candidates according to a partial order or cost function.
\end{enumerate}

For ActPC with RL:

\begin{itemize}
  \item The ''easy'' enumerations are akin to the network's local expansions of possible transitions or substates (either explicitly or implicitly).
  \item The ''hard'' selection is akin to ''optimal choice of next step or next distribution,'' often a local process guided by reward minimization or measure-based geometry.
\end{itemize}

\subsubsection{Applying the Extended DP Theorem to ActPC}

As we've seen, in the standard Mu-Oliveira approach, one obtains:

\begin{equation*}
\mu\left(\lambda X \rightarrow\left(i n \cdot F X \cdot T^{\circ}\right) \triangleright S\right) \subseteq\left([T]^{\circ}\right) \triangleright S \tag{}
\end{equation*}

\noindent where:

\begin{itemize}
  \item $T$ is a deterministic transition relation,
  \item $S$ is the ''shrink operator'' enforcing optimality,
  \item $\mu(\ldots)$ is the least fixed point from a fold/hylomorphism approach.
\end{itemize}

\noindent When transitions and enumerations become approximate/stochastic (e.g. $\widetilde{T}, \widetilde{F}$ ), the theorem states the same containment or near-containment with high probability or up to an $\epsilon$-bound:

$$
\mu\left(\lambda X \rightarrow\left(i n \cdot \widetilde{F} X \cdot \widetilde{T}^{\circ}\right) \triangleright S\right) \subseteq_{\epsilon}\left([\widetilde{T}]^{\circ}\right) \triangleright S
$$

\noindent meaning the iterative approach's solution is ''within $\epsilon$ '' or ''with high probability'' close to the more direct GC-based solution.

ActPC with RL typically does local approximate updates (like a measure-based gradient), so the extended theorem suggests:

\begin{enumerate}
  \item Even if transitions are approximate or noisy, one can interpret the iterative ActPC updates as a ''fold-like'' approach.
  \item The resulting solution remains near or convergent (with high probability) to the ''optimal'' Galois-based DP solution, ensuring we do not drift arbitrarily from the best policy or best value function.
\end{enumerate}

\paragraph{Exploiting the ''Easy vs. Hard'' Breakdown}

A Galois-connection viewpoint breaks the problem into:

\begin{itemize}
  \item Easy side: enumerating candidate sub-states or sub-actions. In ActPC RL, this might be the local expansions or guesses about next hidden-state predictions (the ''fold'').
  \item Hard side: picking or refining which candidate yields the best reward (the ''shrink operator,'' $>S$ ). In ActPC, this is the local measure-based gradient or local ''policy improvement,'' selectively reinforcing or backtracking distribution elements that align with reward signals.
\end{itemize}

\subsection{Unraveling the Semantics of "Easy vs. Hard"}

It's interesting to dissect the different uses of the "easy" vs. "hard" sides of a Galois connection in the use of Galois connections we are proposing for ActPC, versus the use for translating specs into programs.

Mu \& Oliveira's framework for deriving programs from specifications often uses:

\begin{itemize}
\item Easy side (f): A ''fold'' or ''hylomorphism'' that generates or enumerates candidate solutions from a specification. In a classic functional-prog sense, this might be a fold that builds candidate recursive structures.
\item Hard side (g): A ''shrink operator'' $S$ that selects the minimal (or optimal) solution from among the enumerated set, applying constraints. This ''hard'' part shrinks the solution space to a single (or minimal) solution that meets the specification.
\end{itemize}

\noindent So the ''hard'' portion is the ''conceptual difficulty'' of imposing the optimization or minimality condition on an infinite or large set of enumerations. Meanwhile, the ''easy'' portion enumerates all expansions (the specification is often ''all possible programs or data structures''), which is a more mechanical or purely constructive step.

In ActPC with RL or expansions, we interpret:

\begin{itemize}
\item Easy side: The system expands or enumerates sub-states or sub-actions in each micro-step. This is akin to the ''fold'' or ''hylomorphism'' that generically unrolls possibilities. In discrete rewriting or continuous expansions, the ''easy'' part is ''generate a bunch of next states / partial solutions.''
\item Hard side: The aggregator or measure-based gradient picks which expansions or sub-states yield the ''best'' outcome, e.g. the minimal cost or highest reward. It discards suboptimal expansions. This is the ''shrink'' step that imposes an order or partial order to weed out solutions, just as in Mu?Oliveira's theorem.
\end{itemize}

\noindent. Hence, we again see ''one side enumerates, the other side prunes or chooses minimal solutions.'' So the structure is similar to Mu?Oliveira's viewpoint: a Galois connection between enumerations and constraints that produce the minimal (or partial-order?least) element.

The analogy between the "programming via Galois connections" idea and our suggestion here is then roughly:

\begin{itemize}
\item Mu \& Oliveira: The ''specs to programs'' domain is often purely functional or relational, deriving a single final program from a specification.
\item ActPC: The ''specs'' are the system's current or prospective states, and we repeatedly pick expansions that reduce local mismatch or improve reward in real-time.
\end{itemize}

\noindent and regarding the interpretation of ''Hard'':

\begin{itemize}
\item In Mu \& Oliveira, ''Hard'' means imposing the minimal solution subject to an ordering, so ''shrinking'' is the part that solves the big optimization or ''lowest point'' in the partial order.
\item In ActPC expansions, ''Hard'' similarly means ''reject expansions that don't help reduce mismatch or yield better reward,'' or ''pick expansions that do yield improvement.''
\end{itemize}

These differences reflect the differences in objectives:

\begin{itemize}
\item Mu \& Oliveira: The end-goal is a single final program or solution from a hylomorphic enumeration.
\item ActPC: The goal is to converge to minimal mismatch or maximal reward in iterative micro-steps. The aggregator ''shrinks'' expansions each iteration rather than producing a single final solution.
\end{itemize}

Thus, the conceptual parallel is: ''fold/hylomorphism => expansions'' vs. ''shrink => minimal solution.'' In both settings, the ''easy vs. hard'' names are conceptual references to ''generating a large set is (relatively) easy,'' while ''picking or optimizing from that set is the hard or constraining part.'' But the exact domain (program derivation vs. RL expansions) and objective (deriving a correct program vs. picking minimal cost expansions in real-time) differ, causing the two frameworks to appear somewhat different in usage, even though the Galois-connection viewpoint is consistent.

In Mu \& Oliveira's DP Theorem, an iterative approach that alternates between enumerations and ''shrinking'' can be shown to match (or be contained in) the final minimal solution derived by a direct fused approach?provided monotonicity conditions hold (which they do in our case, as we discuss in Appendix \ref{app:conditions}).   Hence, we know that each iteration's expansions plus aggregator shrink is a local approximation to the minimal solution in the partial order?accumulating or converging to the fixpoint if we keep enumerating enough expansions.

Because of these monotonic relationships and the Galois-connection perspective, we get a DP Theorem-style guarantee: repeating expansions and aggregator merges eventually yields a stable solution that is effectively the minimal mismatch or minimal cost in the partial order. In simpler terms, the iterative approach is not missing out on solutions that the ''fused'' approach might have found?the DP Theorem ensures the iterative fold + shrink approach converges to the same minimal fixpoint (or is included in it).

\subsection{Implementation}

Using a Galois-based approach, you systematically derive the ''shrink'' step for each sub-layer or substate. That ensures the agent's local acts of ''eliminating suboptimal expansions'' are indeed correct in an approximate sense, conforming to the partial order set by the reward function.

\subsubsection{Iterative Micro-Updates in a ''Fixed Point'' Manner}

The extended theorem frames the iterative approach as a least fixed point $\mu(\ldots)$. For ActPC:

\begin{itemize}
  \item The network is continuously minimizing local error plus reward mismatch.
  \item Each iteration is akin to ''fold/hylomorphism + shrink.''
  \item By referencing the extended theorem, we see these micro-iterations converge (with high probability or up to $\epsilon$ ) to the ''fused solution'' that a direct Galois-based DP would yield if it were able to do a big global pass.
\end{itemize}

\subsubsection{Reliability Despite Approximation}

ActPC is approximate and (especially with the information-geometry additions) stochastic, but the extended DP theorem in Galois connections says: ''As long as each local step is sufficiently monotone or $\epsilon$-bounded, the final emergent solution is near the global optimum w.r.t. the environment's cost ordering.''

Hence from an engineering perspective:

\begin{enumerate}
  \item We can adopt local approximate expansions (the ''easy side''),
  \item We define a ''shrink'' or measure-based ''policy improvement'' step using Galois connections,
  \item The result is ''guaranteed or highly likely'' to converge close to the best solution in standard DP sense.
\end{enumerate}

Of course, a variety of implementation features need to be handled in making this actually work, e.g.:

\begin{itemize}
  \item Focus-of-Attention: ActPC commonly updates a small subset of the network or hidden states each micro-step. The theorem can still hold if each partial update remains monotone in local partial orders.
  \item Geometry: If using Wasserstein, the agent's local measure-based operator references the same ground metric as the environment cost, which the extended GC-based DP theorem can handle (the ''shrinking'' becomes a geometry-based partial optimality selection).
  \item Convergence Diagnostics: In practice, one can track the agent's distribution or Q-values. The extended theorem says if updates remain ''boundedly approximate,'' the final distribution is $\epsilon$-close to the ''fused DP solution,'' so we can check if the local mismatch is stable within $\epsilon$.
\end{itemize}

The big picture however seems clear: ActPC in a reinforcement-learning setting is essentially doing approximate, stochastic dynamic programming. The extended Galois-connections-based DP theorem guarantees that, under certain ''monotonicity or bounding'' conditions, the iterative local approach in ActPC:

\begin{enumerate}
  \item Stays near the ''ideal'' DP solution in distribution/outcome space.
  \item Converges to or remains within an $\epsilon$-bound of the best possible policy or value function, with high probability.
  \item Achieves a mathematically structured method for bridging ''candidate expansions'' (the easy part) and ''optimal selection'' (the hard part) without ad-hoc guesswork.
\end{enumerate}

One can design ActPC's local expansions and local shrink steps systematically via Galois connections-guaranteeing that the final emergent behaviors approximate the global DP optimum. This unifies the local, gradient-based approach to RL with the well-founded DP logic of the Galois-connection framework.

\subsection{ActPC in terms of Recursion Schemes}

But how might we cash out these mathematical notions in terms of practical software implementation?   In this section we sketch a practical algorithmic breakdown of an iterative local approach to ActPC in a reinforcement-learning style, guided by the ideas of approximate/stochastic dynamic programming and Galois connections.   Our aim here is articulate concrete structures and processes, while still keeping things at a level of abstraction that is applicable to both ActPC-Geom, ActPC-Chem and hybrids thereof.

Conceptually, the idea is to implement ActPC in a loop that:

\begin{enumerate}
  \item Expands or samples candidate next states/sub-states from a local region of the network's distribution,
  \item Shrinks them with measure-based geometry or gradient steps, updating parameters $\theta$,
  \item Executes an action or motor command, gathering new data and updating distributions as a result
  \item Repeats.
\end{enumerate}

To articulate this more precisely we begin with some notation and set-up:

\begin{itemize}
  \item State Space $\mathcal{S}$ : The environment states (external) and possibly internal hidden states of the ActPC network.
  \item Action or Transition: The agent's choice $\alpha$ (which may be partial or emergent from the network).
  \item Reward: The environment's feedback $r$. The agent also internally maintains a cost function or reward distribution $p(r \mid \theta)$.
  \item Local ''Fold/Hylomorphism'': The ActPC network expands partial sub-states or predictions, then ''shrinks'' them using Galois-based optimization (the ''hard part'').
  \item Approximate/Stochastic: Each local iteration only approximates or samples from the full sub-state expansions, rather than enumerating all.
\end{itemize}

We can then propose the following algorithmic steps for a Galois-connection-inspired, recursion-scheme-based ActPC core loop:

\begin{enumerate}
  \item Initialize:

\begin{itemize}
  \item Parameter $\theta$ for the network, possibly random or pretrained.
  \item (Optional) a ''Focus-of-Attention'' subset of states or sub-layers for ActPC to update in this round.
\end{itemize}

  \item Observe / Predict:

\begin{itemize}
  \item The agent receives the current external state $s_{t} \in \mathcal{S}$ (and reward from the previous step, if any).
  \item The network forms an internal distribution or partial outcome prediction $p(r \mid \theta)$.
\end{itemize}

  \item Local Expand (Easy Part):
  
\begin{itemize}
  \item The ''fold/hylomorphism'' step enumerates or samples candidate sub-states. In DP or RL terms, these are possible next states or sub-actions.
  \item In a Galois perspective, a function $\widetilde{F}(X)$ might produce partial expansions from the current internal state distribution. (This is approximate: it picks a subset, or a random sample.)
\end{itemize}

  \item Local Shrink (Hard Part):

\begin{itemize}
  \item The system applies a ''shrink operator'' $\quad S$ that references an optimization order.
  \item Concretely, for each candidate, the network checks (in approximate form) which sub-state leads to better predicted reward or lower cost.
  \item This yields an updated local distribution or local parameter step $\theta \leftarrow \theta+\delta \theta$ that ''selects'' or emphasizes sub-states with higher utility.
\end{itemize}

\item{Apply Parameter Update:}
\begin{itemize}
  \item The measure-dependent operator referencing Galois-based geometry (e.g. a Wasserstein or partial gradient) modifies $\theta$.
  \item This is done in a small step (like $\theta_{t+1}=\theta_{t}-\eta \nabla_{\theta} \ldots$. or a local partial rewriting if we are in a symbolic domain.
\end{itemize}

\item{Act / Output:}
\begin{itemize}
  \item The network picks an action $\alpha_{t}$ from the updated distribution or the best sub-state from the local ''shrink.''
  \item The environment transitions to a next state $s_{t+1}$, delivering reward $r_{t+1}$.
\end{itemize}

\item{Iterate}
\begin{itemize}
  \item Over time, small approximate expansions (easy part) plus local ''shrink'' (hard part) converge (with high probability) to near-optimal behavior, as per the extended DP theorem for approximate/stochastic systems.
\end{itemize}

\end{enumerate}

\subsection{Why a Galois-Inspired Implementation Enables Efficient Concurrency}

A "standard" implementation of ActPC might look something like:

\begin{itemize}
  \item One or a few threads do a big forward pass or local error pass across the entire network's parameters.
  \item Possibly we do some naive data parallelism: each core processes a slice of data, then we sum gradients.
\end{itemize}

A Galois-connection perspective (particularly the ''easy part'' enumerations and ''hard part'' shrink operators) takes a different approach and naturally decomposes a global ActPC loop into smaller local steps:

\begin{enumerate}
  \item Local expansions: Each CPU thread can propose new candidate sub-states or partial solutions from a subset of the network or environment states.
  \item Local shrinks: Each thread applies an order-based ''shrink'' to narrow or optimize solutions within that subset, referencing a partial order for cost or reward.
\end{enumerate}

Because these expansions and shrinks are often monotonic w.r.t. a partial order, concurrency is straightforward to manage as long as merges are partial-order-aware. This decomposition is more explicit than a more monolithic standard ActPC implementation approach, which might do bigger synchronous updates with less structured concurrency.

Potential gains here include:

\begin{itemize}
  \item Better Exploitation of Cores: If each local Galois-based step is small and independent, the overhead of each thread can be modest, letting you use 100 CPU cores in near full parallel.
  \item Focus-of-Attention: By distributing expansions across multiple threads, the system can explore more sub-states or partial parameter sets simultaneously.
  \item Local Memory: If each thread mostly interacts with a small portion of the state or parameter space, we can exploit caches effectively, limiting RAM bandwidth.
\end{itemize}

On the other hand, there are also potentially substantial implementation overhead costs.  Implementing partial-order merges or concurrency-safe data structures is not trivial. You need to, for instance:

\begin{itemize}
  \item Maintain concurrency locks or lock-free structures (like atomic compare-and-swap on cost tables).
  \item Possibly do complex merges where we discard worse solutions or keep better ones.
\end{itemize}

\noindent All his might become nontrivial overhead if we do small, frequent merges among many threads.

A Galois-connection approach also adds extra abstraction layers (the ''shrink'' operator, adjacency-based expansions, partial domain bounding). If not well-optimized, the overhead in function calls or data structure manipulations can hamper performance in a way that eats away theoretical performance gains.

At the high level, what we can say is that:

\begin{itemize}
  \item If each iteration (expansion + shrink) is very small and local, concurrency can scale well to dozens of cores, possibly near-linear speedup-if merges or partial sync do not become the bottleneck.
  \item If merges are frequent and large, concurrency may saturate (lots of cache-line contention, or lock contention).
  \item If we can partition the network's states/parameters fairly well among the cores (''focus-of-attention'' style), then threads only rarely need to sync or unify solutions, leading to high parallel efficiency.
\end{itemize}

Hence, a 100-core server might see anywhere from a small factor speedup (2-10x) if merges are frequent, to near 50-90x if merges are relatively infrequent and expansions/shrinks remain quite local.   Of course, real speedups also hinge on memory throughput and the fraction of time spent on merges.

It is worth noting that the rho calculus \cite{meredith2005reflective} and its extension the metta calculus \cite{meredith2025metta} provide rigorous and practical foundations for implementing this sort of system.   The deployment of metta calculus within the OpenCog Hyperon MeTTa interpreter is designed specifically to keep these sorts of overheads in check.   So in the end, we believe all this is do-able, but we are just noting that it does require significant care to the details of the structures and processes at each level, or the theoretical advantages won't necessary flow through into the practical implementations.

\subsection{A Unified Partial Order / Probability Measure across Discrete and Continuous ActPC Networks}

One loose end not addressed in detail in earlier sections was the creation of a common Wasserstein-metric probability space across discrete and continuous nodes in a hybrid neural/ActPC-Chem system.   This comes up in the Galois connection context because it's needed to create a single partial order for a neural-symbolic ActPC system, but it is of course an issue of general importance beyond the scope of Galois-based implementation strategies.

Such a common metric probability space can indeed be created under appropriate design choices, but implementing such a unifying Wasserstein measure requires careful construction of a ground metric that spans discrete rewriting states and continuous error states. In this section we break down how to do this, and what caveats exist.

\paragraph{Stating the Core Issue}.  In a hybrid neural-symbolic ActPC,  we have two fundamentally different sorts of nodes, e.g.

\begin{itemize}
  \item Discrete rewriting: E.g., ''algorithmic chemistry'' rewriting rules, which produce new discrete structures or sub-states. We might define a cost or error measure for each rewrite step (uncertainty, complexity, etc.).
  \item Continuous neurons: E.g., real-valued node activations or parameter updates. We have a numeric cost function (e.g., MSE, or $\|\cdot\|$-based error, or something similar).
\end{itemize}

In Galois-connection parlance, we need a common partial order that allows us to unify expansions from discrete rewriting and from continuous updates in the ''shrink'' operator. Otherwise, we have two incommensurable cost dimensions, and can't easily decide which partial solution is ''better.''

The obvious approach here in an ActPC-Geom context is to interpret both discrete rewriting states and continuous states as distributions over some domain $\mathcal{X}$. Then, we define a Wasserstein distance $W(\cdot, \cdot)$ on $\mathcal{P}(\mathcal{X})$. Specifically:

\begin{enumerate}
  \item Each sub-state $\mathbf{S}$ (whether discrete or continuous) is mapped to a probability distribution $p_{\mathbf{S}}$.
  \item The ''optimal reward-achieving actions'' or ''target distribution'' is also some distribution $p_{\text {optimal }}$.
  \item The distance or partial order is then '' $p_{\mathbf{S}_{1}}$ is better or not than $p_{\mathbf{S}_{2}}$ ?'' if $W\left(p_{\mathbf{S}_{1}}, p_{\text {optimal }}\right) \leq$ $W\left(p_{\mathbf{S}_{2}}, p_{\text {optimal }}\right)$.
\end{enumerate}

\noindent Essentially, a sub-state or partial solution is closer in Wasserstein distance to the distribution of ''truly optimal'' states if it has lower cost. This yields a single partial order:

$$
\mathbf{S}_{1} \leq \mathbf{S}_{2} \Longleftrightarrow W\left(p_{\mathbf{S}_{1}}, p_{\text {optimal }}\right) \leq W\left(p_{\mathbf{S}_{2}}, p_{\text {optimal }}\right)
$$

\subsubsection{Hybrid Ground Metric for Discrete + Continuous}
To define one Wasserstein distance, we need one ground metric $\omega(\mathbf{x}, \mathbf{y})$ that can measure how ''far'' a discrete rewriting sub-state is from a continuous sub-state or from another discrete sub-state, etc. This might involve:

\begin{itemize}
  \item Discrete portion: Some form of ''edit distance'' on the rewriting subgraphs.
  \item Continuous portion: Euclidean or L2 distance for neural parameters.
  \item Combine them in a single metric, e.g.

$$
\omega\left(\left(d_{1}, c_{1}\right),\left(d_{2}, c_{2}\right)\right)=\alpha d_{\text {discrete }}\left(d_{1}, d_{2}\right)+\beta d_{\text {continuous }}\left(c_{1}, c_{2}\right)
$$

\noindent for tunable weights $\alpha, \beta$.

\end{itemize}

Then we interpret each sub-state $\mathbf{S}$ as distributing some probability mass on ''(discreteRepresentation, continuousRepresentation)'' in $\mathcal{X}$.  We can define or approximate '' $p_{\mathrm{s}}$ '' in that product space. Once we have that, the Wasserstein distance $W_{2}\left(p_{\mathbf{S}}, p_{\text {optimal }}\right)$ is well-defined.

\paragraph{Using This as a Common Partial Order}. Once each node or sub-state is mapped to a distribution in $\mathcal{P}(\mathcal{X})$, we define:

$$
\mathbf{S}_{1} \leq \mathbf{S}_{2} \Longleftrightarrow W\left(p_{\mathbf{S}_{1}}, p_{\text {optimal }}\right) \leq W\left(p_{\mathbf{S}_{2}}, p_{\text {optimal }}\right)
$$

Using this approach we can say that:

\begin{itemize}
  \item Both discrete rewriting expansions and continuous updates aim to reduce ''distance to the optimum.''
  \item The aggregator or ''shrink'' operator can pick states that yield a smaller Wasserstein distance to $p_{\text {optimal }}$.
  \item We effectively unify the cost function: the cost is ''how far'' we are from the optimum distribution, measured in $\omega$-based Wasserstein distance.
  \item The ActPC system can treat discrete expansions and continuous expansions identically when deciding which solutions to keep or discard.
  \item Concurrency is conceptually simple: each expansion thread just yields a new sub-state $\mathbf{S}_{\text {new }}$. We compute/approximate $W\left(p_{\mathbf{S}_{\text {new }}}, p_{\text {optimal }}\right)$. The aggregator merges them by picking the ones with smaller distance.
\end{itemize}

\paragraph{Tuning Could Be Tricky}  Tuning $\alpha$ vs. $\beta$ is nontrivial.  For instance, is the discrete mismatch more or less important than a certain difference in continuous space?   Rather than trying to achieve this sort of balance in a deeply theoretically grounded way, it may work best to simply set these weights pragmatically, based on fixing a desired balance between discrete and continuous components altogether (say, equal weighting of the two, or $\frac{2}{3}$ continuous vs $\frac{1}{3}$ discrete, or whatever) and then adaptively update the weights to roughly achieve the specified balance.

Putting this sort of heuristic weighting together with the use of numerous somewhat-crude approximations necessitated to make actual computation of the metric tractable (on either the discrete or continuous side), what we will arrive at in any large-scale practical system is more a rough guide to system dynamics, heavily inspired by formal theory, than something rigorously derived down to the last decimal point.   

But with this caveat in mind, we can say that: Yes, creating a common Wasserstein-metric for both discrete rewriting states and continuous states can unify them in one partial order: states that are ''closer'' (in the Wasserstein sense) to the ''optimal reward-achieving distribution'' are considered better.  It's not terribly problematic to address the problem of bridging discrete expansions vs. continuous expansions with a single partial order, enabling a common aggregator or ''shrink'' operator in an iterative local ActPC or dynamic-programming sense.

\section{Gesturing Toward a Specialized HPC Architecture for ActPC-Geom}

Even assuming the mathematical approximations and optimizations outlined above work effectively, one may still face a significant challenge in implementing ActPC-Geom scalably on contemporary computing hardware.   

However, we believe this challenge will be surmountable leveraging a combination of contemporary technologies.   The computational challenges of using backpropagation to train large neural networks have been met, in part, by targeted specialization of computing hardware and associated low-level software.  Analogous specialization could be carried out for ActPC-Geom.  This doesn't require as radical a deviation from currently standard architectures as one needs for effectively implementing symbolic AI or a radically cross-paradigm system like Hyperon, because we are still largely dealing with matrix multiplications here -- they are just organized differently than in typical backprop-based learning.
  
In this section we speculate a bit regarding potential custom software and hardware pipelines aimed at optimizing operation of ActPC-Geom on large networks, especially though not exclusively in the context of hybrid architectures connecting ActPC-Geom with ActPC-Chem and other symbolic models.

It may seem this is getting too far ahead of reality -- thinking about custom software and hardware infrastructures for an algorithm that's still at the speculative stage and not even proven to work yet!   However, the reality of modern AI is that the methods enjoying the greatest current success are doing so partly due to the prevalence of relatively affordable underlying processing infrastructures closely suiting their requirements.  So when outlining alternative approaches, it seems important to clarify that similar infrastructure optimizations could also be done for the alternatives.   

Of course, it would not make sense to put significant effort into creating this sort of optimized infrastructure  for ActPC-Geom until greater validation of the core ActPC-Geom ideas has been obtained.  It does seem important, though, to be quite clear that this sort of optimization path exists.  Because in the current rapid-progress AI climate, it is seeming decreasingly worthwhile to put significant effort into AI methods that can't be practically optimized for large-scale operation on hardware viably buildable in the relatively near term.

The outlining of an HPC architecture here also serves the expository purpose of reviewing all the different architectural and algorithmic aspects discussed above in a unified fashion.

\subsection{Toward ActPC-Geom Customized HPC Infrastructure}

To optimize ActPC-Geom, we want a large-scale HPC pipeline that efficiently runs a variety of operations, such as:

\begin{itemize}
  \item ActPC in a continuous manner (e.g., standard neural network predictive coding),
  \item ActPC-Chem in a discrete rewriting or ''algorithmic chemistry'' manner,\\
and unifies them under a single measure-based aggregator using:
  \item Wasserstein-based or info-geometry-based error minimization for local updates
  \item Approximate kPCA and hypervector embeddings for discrete+continuous states
  \item Sophisticated hypervector manipulations, including heuristic search for partial unbindings
  \item Fuzzy FCA concept-lattice auto-learning to identify which concepts and properties are relevant to geometric modeling
 \end{itemize}

\paragraph{Subsystems}  This sort of pipeline may involve subsystems such as:

\begin{enumerate}
  \item Neural Subsystem (Continuous ActPC):

\begin{itemize}
  \item Deals with real-valued sensor data, motor commands, and typical neural net layers.
  \item Predictive coding loops compute local mismatch in real time.
\end{itemize}

  \item Discrete Subsystem (ActPC-Chem):

\begin{itemize}
  \item Algorithmic rewriting or discrete transformations.
  \item Also harnesses predictive coding by turning each rewriting step into a local mismatch minimization.
\end{itemize}

  \item Compositional Hypervector Representation:

\begin{itemize}
  \item For each partial state or rewriting rule, we maintain a large hypervector that encodes the fuzzy properties discovered by the combination of approximate kPCA and ''Fuzzy FCA concept learning'' mechanism.
\end{itemize}

  \item Measure-Based Aggregator:

\begin{itemize}
  \item Galois-connection-inspired concurrency logic that merges expansions from discrete + continuous sides, chooses minimal Wasserstein-based error expansions, and prunes worse expansions.
  \item Possibly orchestrates micro-iterations or partial expansions in parallel across many HPC cores.
\end{itemize}

\item  Fuzzy FCA Learning:

\begin{itemize}
  \item A dedicated or integrated neural module that regularly (or incrementally) updates a set of fuzzy features (the ''concept lattice''), referencing both discrete rewriting states and continuous parameters.
  \item Outputs ''feature-value vectors'' for each partial state, which then get bound into hypervectors.
\end{itemize}

\end{enumerate}

\subsection{Architecture Modules}

We now run through the modules that would be needed in such a HPC pipeline, manifesting the conceptual subsystems described above:

\subsubsection{Fuzzy Feature Learning Module ( $\mathcal{F}$ )}
\begin{itemize}
  \item Role: Discover relevant fuzzy properties (features) from the system's states.
  \item Maps each state $S \in \mathcal{S}$ (which could be a combination of neural activation + discrete rewriting config) to a fuzzy membership vector $\mathbf{f}(S) \in \mathbb{R}^{N}$.
  \item These features are effectively the fuzzy concept lattice dimensions, learned via a neural approach or some heuristic plus a small net.
\end{itemize}

\paragraph{Implementation:}
\begin{enumerate}
  \item Neural: A small MLP or attention block that sees both discrete and continuous aspects of $S$ and outputs $N$-dim fuzzy membership values in $[0,1]$.
  \item Loss: Co-trained so that these fuzzy features let the rest of the system better predict system dynamics or target properties (see Utility/Prediction Module below).

  \item Updating:
  \begin{itemize}
  \item Runs periodically or in streaming mode, reflecting new states or rewriting expansions.
  \item Maintains a stable set of fuzzy features that evolve slowly, ensuring compositional consistency.
\end{itemize}
\end{enumerate}

\subsubsection{Hypervector Embedding \& Compositional Algebra $(\mathcal{H})$}
\begin{itemize}
  \item Role: For each state $S$, produce a high-dimensional hypervector $\mathbf{v}(S) \in\{ \pm 1\}^{D}$ or $\mathbb{R}^{D}$.
  \item This hypervector encodes the fuzzy concept membership from $\mathcal{F}$ plus the rewriting or continuous sub-structure.
\end{itemize}

In particular,

\begin{itemize}
  \item We define base vectors for each fuzzy feature $F_{j}$ and each discrete property/ rewriting rule, etc.
  \item The membership values from $\mathcal{F}$ are used as scaling or binding factors. For instance:

$$
\mathbf{v}(S)=\operatorname{Bundling}\left(\ldots, \operatorname{Bind}\left(\mathbf{v}_{\text {Feature } j}, \operatorname{MemVal}(S, j)\right), \ldots\right)
$$

\end{itemize}

The compositional structure here is significant:

\begin{itemize}
  \item If rewriting steps add new properties, we can bind them in.
  \item This structure allows partial unbinding for retrieving which features are active.
\end{itemize}

The structure must be adaptive: If $\mathcal{F}$ changes, $\mathcal{H}$ changes. We might need incremental re-embedding or partial  updates.

\subsubsection{ Neural Utility/Prediction Module $(\mathcal{U})$}
\begin{itemize}
  \item Role: Takes the hypervector $\mathbf{v}(S)$ as input (or a dimension-reduced version).
  \item Outputs predictions $\hat{\mathbf{y}}(S) \in \mathbb{R}^{m}$ about system's future or reward, used by the ActPC or rewriting logic.
  \item Wasserstein-based: We measure the mismatch between $\hat{\mathbf{y}}(S)$ and the ground-truth $\mathbf{y}(S)$ using a Wasserstein or other info-geometry measure. This ensures the entire system's error is ''transport-cost''-aligned, not just raw MSE or KL.
\end{itemize}

\subsubsection{Discrete Rewriting Subsystem (ActPC-Chem)}
\begin{itemize}
  \item Role: Maintains the set of rewriting rules, partial expansions.
  \item On each micro-step, it proposes expansions $\left\{\mathbf{v}_{c_{1}}, \ldots\right\}$ in hypervector form. Possibly using the fuzzy concept membership to decide feasible rewrites.
\end{itemize}

\subsubsection{Continuous Subsystem (ActPC)}
\begin{itemize}
  \item Role:The usual neural net layers that handle continuous signals.
  \item In synergy with the discrete rewriting states, also produce expansions in hypervector form.
\end{itemize}

\subsubsection{Aggregator / Galois Concurrency Manager ( $\mathcal{A}$ )}
\begin{itemize}
  \item Role: Runs a Galois-connection-style aggregator that merges expansions from the discrete rewriting side + the continuous side, and references the hypervector measure-based geometry or partial overlaps.
  \item Picks the expansions that yield minimal Wasserstein-based predictive error, discarding suboptimal expansions.
  \end{itemize}
  
The implementation here may be complex, including aspects such as:

\begin{enumerate}
  \item Parallel expansions in multiple HPC cores.
  \item Aggregator merges them by computing ''distance in hypervector space to $\mathbf{v}_{\mathrm{Q}}$ '' or to an ''optimal reference.'' Possibly tries partial unbindings or partial property checks for more advanced search.
  \item Prunes expansions that do not reduce error.
  \item Feeds back accepted expansions into the next micro-iteration or updates the rewriting states.
\end{enumerate}

\subsection{HPC Execution Outline}

The data flow implicit in the above breakdown looks roughly like:

\begin{enumerate}
  \item Entity / State $S$ arrives or is updated.
  \item Fuzzy Feature net $\mathcal{F}$ maps $S \mapsto \mathbf{f}(S)$.
  \item Hypervector module $\mathcal{H}$ binds these fuzzy features (plus other discrete/continuous data) $\rightarrow \mathbf{v}(S)$.
  \item Utility net $\mathcal{U}$ sees $\mathbf{v}(S)$, outputs $\hat{\mathbf{y}}(S)$.
  \item We measure mismatch $\operatorname{Loss}(\hat{\mathbf{y}}(S), \mathbf{y}(S))$ (Wasserstein-based, for example).
  \item In parallel, Discrete rewriting or Continuous expansions produce new candidate states $\left\{S^{\prime}\right\}$. Each is turned into $\mathbf{v}\left(S^{\prime}\right)$. The aggregator merges them, picks minimal-error expansions.
  \item We do partial or full HPC-based updates to $\mathcal{F}, \mathcal{H}, \mathcal{U}$ (and rewriting rules) via backprop or ActPC. Possibly in micro-iterations.
\end{enumerate}

Opportunities for concurrency are rife here:

\begin{itemize}
  \item Multiple HPC cores do expansions in parallel. Each expansion calls $\mathcal{F} \rightarrow \mathcal{H} \rightarrow \mathcal{U}$ (or partial versions) to evaluate error or synergy.
  \item The aggregator merges results in a partial-order sense (lowest error expansions survive).
  \item Meanwhile, $\mathcal{F}$ is updated periodically if new fuzzy features must appear or old ones can be refined.
\end{itemize}

There are also opportunities to mix components operating at different natural time-scales, e.g.

\begin{itemize}
  \item If we do real-time ActPC, each iteration might use the aggregator to pick expansions in tens of microseconds or milliseconds.
  \item The fuzzy concept net $\mathcal{F}$ might adapt slower, maybe every second or minute, so as not to disrupt stable property definitions too often.
\end{itemize}

\subsection{A Multilayer Concurrency Model}

This workflow corresponds naturally to a multi-layer concurrency model involving aspects such as: (1) cluster-level node distribution, (2) node-level concurrency across CPU cores and GPUs, and (3) sub-node-level concurrency for aggregator, expansions, and measure-based geometry.  Each of these would be a complex story in practice; here we outline a few of the key aspects that seem to be involved.

\subsubsection{Cluster-Level Architecture}

To set this sort of architecture up on a supercomputer would seem to involve ingredients such as:

\begin{itemize}
  \item N HPC Nodes: Each node has several CPU sockets (e.g., 2-4), each with multiple hyperthreaded cores, plus M GPUs (e.g. 4-8).
  \item High-Speed Interconnect: e.g. InfiniBand or a high-bandwidth internal fabric.
  \item Distributed Storage: Possibly a parallel file system for large model checkpoints or big dataset logs.
\end{itemize}

One would partition the overall HPC pipeline so that each node runs a subset of expansions or aggregator tasks in parallel, coordinating via a distributed concurrency layer. The system is primarily synchronous or loosely synchronous, but each node can do local expansions and aggregator merges independently in micro-steps, exchanging partial results or states across the cluster every so often.

\subsubsection{Node-Level Organization}. Within each HPC node in this architecture we would have neural-symbolic interactions:

\begin{enumerate}
  \item CPUs:

\begin{itemize}
  \item Several multi-core CPUs with hyperthreading.
  \item We assign certain tasks to the CPU threads:
  \item Discrete rewriting expansions (ActPC-Chem) can be ''lighter'' in GPU usage but heavy in symbolic logic, so many CPU threads can handle that in parallel.
  \item Aggregator concurrency logic (the partial-order merges, Galois-based aggregator) can also use CPU threads-these tasks can be quite branching/forkjoin in nature, well-suited to large CPU concurrency.
\end{itemize}

  \item GPUs:

\begin{itemize}
  \item Each node may have multiple GPUs that handle:
  \item Neural net forward/backward for the continuous ActPC portion.
  \item Neural-based fuzzy-FCA concept learning: If the fuzzy feature extraction $\mathcal{F}$ or the Utility/Prediction net $\mathcal{U}$ are large, they run on GPUs for matrix multiplications or attention blocks.
  \item Possibly large-scale hypervector binding/unbinding or partial dimension reduction could also be GPU-accelerated, especially if we do random features or partial SVD on large embeddings.
\end{itemize}

  \item Shared RAM:

\begin{itemize}
  \item CPU-accessible memory storing the data structures for rewriting states, aggregator partial order, intermediate expansions.
  \item GPU memory used for the biggest neural layers or large hypervector ops (like binding).
  \item We might do a NUMA-aware approach, pinning aggregator threads near their data, while GPU tasks handle big linear algebra.
\end{itemize}

\end{enumerate}

\subsubsection{Sub-Node HPC Workflow}

For the rewriting dynamics within an ActPC-Chem subsystem, we would have a variety of activities occupying multiple CPUs:

\begin{itemize}
  \item Each rewriting thread picks a set of partial states from the ''frontier.''
  \item Proposes expansions (new rewrite states).
  \item Calls the Fuzzy Feature net $\mathcal{F}$ on the GPU (if needed) or a CPU version if it's small, obtaining fuzzy membership vectors.
  \item Passes them to the Hypervector module $\mathcal{H}$ (which can run on GPU or CPU with specialized binding kernels).
  \item Produces candidate hypervectors $\mathbf{v}_{\text {candidate }}$.
\end{itemize}

\subsubsection{Continuous ActPC Expansions}

For the dynamics within a continuous ActPC-Geom subsystem, processing would be GPU-centric:

\begin{itemize}
  \item The continuous ActPC layers do local micro-iterations of predictive coding.
  \item Periodically produce candidate state embeddings $\mathbf{v}_{\text {candidate }}$ that represent the new continuous states after partial updates.
  \item Return these to CPU aggregator or store them in a shared data structure in CPU memory for aggregator usage.
\end{itemize}

\subsubsection{Aggregator}

The aggregator used for hypervector operations could productively leverage a dedicated CPU-Thread Pool, and potentially even a subset of CPU cores.   Each aggregator thread:

\begin{enumerate}
  \item Collects new candidate expansions $\mathbf{v}_{\text {candidate }}$.
  \item Compares them (in parallel) to the question vector(s), or ''target reference vectors,'' or stored best solutions, using a Wasserstein or $\ell_{2}$ measure. If it's $\ell_{2}$ in hypervector space, that's a straightforward dot-product or distance operation, which can be done quickly or partially GPU-accelerated if needed.
  \item Merges or discards them by partial order logic: if a candidate is strictly worse than an existing expansion, aggregator discards it.
  \item Possibly does partial unbinding via a multi-route search approach.
  \item The aggregator outputs a pruned list of expansions for the next micro-step or sets them aside in HPC memory.
\end{enumerate}

Concurrency can be handled in straightforward ways:

\begin{itemize}
  \item Each aggregator thread can handle a portion of expansions (like ''shard 1 gets expansions \#1-1000, shard 2 gets expansions \#1001-2000, ...''), or we can do a global lock-free structure with atomic merges.
  \item Periodically aggregator shards exchange updated minimal expansions so the global partial order remains consistent.
\end{itemize}

\subsubsection{Fuzzy-FCA Net $(\mathcal{F})$ Training}

This could be run on either GPU or CPU; but for a large network, mini-batches on GPU will be the best approach, with particulars such as:

\begin{itemize}
  \item Loss: The Utility net $\mathcal{U}$ on GPU trains to predict the numeric target from the fuzzy membership vectors.
  \item We do backprop or predictive-coding-style gradient flows that update both $\mathcal{U}$ and $\mathcal{F}$
  \item This might happen asynchronously: a CPU thread or GPU kernel collects a batch of states $(S, \mathbf{y}(S)$ ), runs them through $\mathcal{F}$ and $\mathcal{U}$, and accumulates a gradient. Then it updates $\mathcal{F}$ or $\mathcal{U}$ weights. Over time, the fuzzy concept definitions adapt.
\end{itemize}

\subsubsection{Periodic Re-Embedding}

Because $\mathcal{F}$ changes, the hypervector binding scheme might need incremental updates or partial dimension shifts. We might define a schedule:
 
  \begin{itemize}
  \item Every T steps or every so often, re-check the base hypervectors or do partial kernel expansions.
  \item Possibly do a distributed ''embedding manager'' on GPU that re-initializes or updates the large hypervector base vectors for newly created fuzzy features.
\end{itemize}

\subsubsection{Inter-Node Communication}

Suppose we have multiple HPC nodes, each node has this internal pipeline. They must occasionally exchange:

\begin{enumerate}
  \item Expansions: If expansions from node A might help node B. Possibly each node only handles a subset of states.
  \item Aggregator Info: Summaries of which expansions are minimal in partial order.
  \item Fuzzy-FCA Weights or '' $\mathcal{F}$ net parameters'': We can do data-parallel training, periodically averaging or merging model parameters across nodes.
  \item $\mathcal{U}$ net: Similarly can do a distributed parameter approach.
\end{enumerate}

A typical approach here, implementation-wise, is to do a synchronous or semi-synchronous model update every so many expansions.
The aggregator partial orders can be sharded by hashing states or expansions to nodes, exchanging minimal solutions that cross boundaries.

Nodes communicate minimal expansions or best solutions. If a new state from node A is ''strictly better'' than anything known at node B, B might incorporate it. The measure-based geometry is consistent if the hypervector base is the same across nodes (they share $\mathcal{F}$ and $\mathcal{H}$ parameters).

\subsection{Recap of HPC Design}

In the design we have sketched here, each HPC Node has the following structure:

\begin{itemize}
  \item A pool of CPU threads handles ActPC-Chem expansions and aggregator concurrency logic.
  \item GPUs handle continuous ActPC micro-iterations (for large neural net layers) and handle training of $\mathcal{F}$ (fuzzy-FCA net) and $\mathcal{U}$ (utility net).
  \item The aggregator merges expansions in hypervector space, measuring $\ell_{2}$ or partial unbinding overlap, using a Galois partial order approach.
  \item The fuzzy concept net is periodically updated via partial mini-batch training on GPU.
\end{itemize}

At the cluster level,

\begin{itemize}
  \item Each node focuses on a subset of expansions or states, but occasionally merges partial results via interconnect.
  \item Model parameters for $\mathcal{F}$ or $\mathcal{U}$ are distributed or data-parallel.
  \item The entire system is orchestrated by a global or hierarchical aggregator concurrency layer that ensures the partial order is consistent across nodes.
\end{itemize}

This provides a scalable HPC pipeline that implements the new ideas explored here:

\begin{itemize}
  \item System-wide Wasserstein error for $\mathcal{U}$.
  \item Compositional hypervector embeddings for states (both discrete rewriting + continuous).
  \item Neural-based fuzzy-FCA concept learning, evolving a fuzzy lattice.
  \item Periodic adaptation of hypervector base/embedding if $\mathcal{F}$ changes significantly.
  \item Galois concurrency to exploit multi-thread expansions at node-level and merges across nodes.
\end{itemize}

\subsection{Further Directions for Optimized Implementation}

\paragraph{Custom Hardware}  One could take the optimization one step further and spell out a hardware design for custom chips designed to coexist with standard CPUs/GPUs in a supercomputing setup and geared toward ActPC-Geom + ActPC-Chem pipelines.  One would introduce specialized on-chip components for discrete rewriting expansions, neural-based fuzzy concept learning, compositional hypervector operations, and measure-based aggregator concurrency.  Even assuming contemporary chip fab constraints (no exotic post-CMOS tech) there would still be room for significant architectural innovations (e.g. more processor-in-RAM, specialized dataflow units, memory-located matrix ops, etc.).   One could then conceive a distributed supercomputer integrating a mixture of standard CPU/GPU boards and these new custom boards/chips, orchestrated via a high-speed cluster interconnect.

\paragraph{Decentralized Deployment}  In a different direction, one could attempt to splay out the different modules articulated in our above HPC design across a decentralized network, rather than a single co-located distributed supercomputer.   This would entail additional complexities.  E.g. one might have a large slowly-updated information-geometry model spanning a large portion of a decentralized network, and then more rapidly-updated localized information-geometry models focused on specific localized portions of the network.  Tools such as SingularityNET, NuNet and MettaCycle could be used to coordinate decentralized modules with different owners and managers.  Rho calculus could be used, on the back end of MeTTa smart contracts, to manage concurrency within a single machine and distributed processing across machines in a unified fashion.

\bibliographystyle{alpha}
\bibliography{actpc-geom}

\newcommand{\etalchar}[1]{$^{#1}$}
\begin{thebibliography}{GBD{\etalchar{+}}23}

\bibitem[Ama16]{amari2016information}
Shun-ichi Amari.
\newblock {\em Information geometry and its applications}, volume 194.
\newblock Springer, 2016.

\bibitem[Ami90]{amit1990attractor}
Daniel~J Amit.
\newblock Attractor neural networks and biological reality: associative memory
  and learning.
\newblock {\em Future Generation Computer Systems}, 6(2):111--119, 1990.

\bibitem[Fri09]{friston2009free}
Karl Friston.
\newblock The free-energy principle: a rough guide to the brain?
\newblock {\em Trends in cognitive sciences}, 13(7):293--301, 2009.

\bibitem[GBD{\etalchar{+}}23]{goertzel2023hyperon}
Ben Goertzel, Vitaly Bogdanov, Michael Duncan, Deborah Duong, Zarathustra
  Goertzel, Jan Horlings, Matthew Ikle', Lucius~Greg Meredith, Alexey Potapov,
  Andre'~Luiz de~Senna, Hedra Seid~Andres Suarez, Adam Vandervorst, and Robert
  Werko.
\newblock Opencog hyperon: A framework for agi at the human level and beyond.
\newblock 2023.
\newblock Preprint.

\bibitem[GBIY16]{goertzel2016controlling}
Ben Goertzel, Misgana~Bayetta Belachew, Matthew Ikle?, and Gino Yu.
\newblock Controlling combinatorial explosion in inference via synergy with
  nonlinear-dynamical attention allocation.
\newblock In {\em Artificial General Intelligence: 9th International
  Conference, AGI 2016, New York, NY, USA, July 16-19, 2016, Proceedings 9},
  pages 334--343. Springer, 2016.

\bibitem[Goe21]{goertzel2021patterns}
Ben Goertzel.
\newblock Patterns of cognition: cognitive algorithms as galois connections
  fulfilled by chronomorphisms on probabilistically typed metagraphs.
\newblock {\em arXiv preprint arXiv:2102.10581}, 2021.

\bibitem[Goe24a]{goertzel2024actpc}
Ben Goertzel.
\newblock Actpc-chem: Discrete active predictive coding for goal-guided
  algorithmic chemistry as a potential cognitive kernel for hyperon \&
  primus-based agi.
\newblock {\em arXiv preprint arXiv:2412.16547}, 2024.

\bibitem[Goe24b]{goertzel2024hyperseed}
Ben Goertzel.
\newblock Introducing hyperseed-1.
\newblock \url{https://bengoertzel.substack.com/p/introducing-hyperseed-1},
  2024.

\bibitem[LM18]{li2018natural}
Wuchen Li and Guido Mont{\'u}far.
\newblock Natural gradient via optimal transport.
\newblock {\em Information Geometry}, 1:181--214, 2018.

\bibitem[Mer25]{meredith2025metta}
Lucius~Gregory Meredith.
\newblock Metta calculus.
\newblock \url{https://github.com/leithaus/rho4u/tree/main/metta-calculus},
  2025.
\newblock Accessed: 2025-01-07.

\bibitem[MO12]{mu2012programming}
Shin-Cheng Mu and Jos{\'e}~Nuno Oliveira.
\newblock Programming from galois connections.
\newblock {\em The Journal of Logic and Algebraic Programming}, 81(6):680--704,
  2012.

\bibitem[MR05]{meredith2005reflective}
L~Gregory Meredith and Matthias Radestock.
\newblock A reflective higher-order calculus.
\newblock {\em Electronic Notes in Theoretical Computer Science},
  141(5):49--67, 2005.

\bibitem[OK22]{ororbia2022neural}
Alexander Ororbia and Daniel Kifer.
\newblock The neural coding framework for learning generative models.
\newblock {\em Nature communications}, 13(1):2064, 2022.

\bibitem[RR07]{rahimi2007random}
Ali Rahimi and Benjamin Recht.
\newblock Random features for large-scale kernel machines.
\newblock {\em Advances in neural information processing systems}, 20, 2007.

\bibitem[RSL{\etalchar{+}}20]{ramsauer2020hopfield}
Hubert Ramsauer, Bernhard Sch{\"a}fl, Johannes Lehner, Philipp Seidl, Michael
  Widrich, Thomas Adler, Lukas Gruber, Markus Holzleitner, Milena Pavlovi{\'c},
  Geir~Kjetil Sandve, et~al.
\newblock Hopfield networks is all you need.
\newblock {\em arXiv preprint arXiv:2008.02217}, 2020.

\bibitem[Sam22]{samal2022learnfca}
Suraj~Ketan Samal.
\newblock Learnfca: A fuzzy fca and probability based approach for learning and
  classification.
\newblock 2022.

\bibitem[SF14a]{snaider2014modular}
Javier Snaider and Stan Franklin.
\newblock Modular composite representation.
\newblock {\em Cognitive Computation}, 6:510--527, 2014.

\bibitem[SF14b]{snaider2014vector}
Javier Snaider and Stan Franklin.
\newblock Vector lida.
\newblock {\em Procedia Computer Science}, 41:188--203, 2014.

\end{thebibliography}

\appendix

\part*{Appendices}
\addcontentsline{toc}{part}{\protect\numberline{}Appendices}

\section{Proof Sketches for Local Continuity and Scale-Matching Properties when Wasserstein Metric is Used for Both Inner and Outer Loops in ActPC-Geom} \label{app:scale-match}.
 
We sketch here proofs of the propositions made in Section  \label{sec:scale-prop} above.

\subsection{Proof Sketch for Local Continuity}

We break the argument for the Wasserstein Lipschitz ActPC Property into two parts: (A) local continuity, and (B) scale matching.  Let's start with the former, which is simple and straightforward.

By assumption, for all $\theta_{1}, \theta_{2}$ :

$$
\left|W_{2}\left(q, p\left(r \mid \theta_{1}\right)\right)-W_{2}\left(q, p\left(r \mid \theta_{2}\right)\right)\right| \leq L\left\|\theta_{1}-\theta_{2}\right\|
$$

Let $\Delta_{\text {env }}(\theta)=W_{2}(q, p(r \mid \theta))$. Then $\Delta_{\text {env }}(\theta)$ is an $L$-Lipschitz function in $\theta$. By definition, Lipschitz continuity implies local continuity: small changes $\left\|\theta_{1}-\theta_{2}\right\|$ in parameter space produce at most $L\left\|\theta_{1}-\theta_{2}\right\|$ changes in the external cost $\Delta_{\text {env }}(\theta)$.

This directly confirms that $\Delta_{\text {env }}$ cannot vary abruptly over small parameter shifts, ensuring the system sees a ''locally well-behaved'' cost landscape for changes in $\theta$.

Thus, Local Continuity follows immediately from the Lipschitz property.

\subsection{Proof Sketch for Scale Matching}

Scale matching is both messier and a little subtler.  The proof sketch is anchored in the idea that if the environment's reward distribution changes or exhibits variation over a particular scale in the ground metric $\omega$, then the agent's internal measure-dependent operator sees changes in distribution space over a proportionate scale, ensuring the agent's local distribution shifts remain aligned with how the environment's cost function changes.

\paragraph{Step 1: Local Lipschitz $\Longrightarrow$ Bounded External Cost Changes}  By assumption, for all $\theta_{1}, \theta_{2} \in \Theta$,

$$
\left|W_{2}\left(q, p\left(r \mid \theta_{1}\right)\right)-W_{2}\left(q, p\left(r \mid \theta_{2}\right)\right)\right| \leq L\left\|\theta_{1}-\theta_{2}\right\| .
$$

\noindent Hence, if $\left\|\theta_{1}-\theta_{2}\right\|=\|\delta \theta\|$, the external cost function $\Delta_{\text {env }}(\theta)=W_{2}(q, p(r \mid \theta))$ varies by at most $L\|\delta \theta\|$.

\paragraph{Step 2: Distribution Shift vs. Parameter Shift}  Focus on a small parameter change $\delta \theta$. The agent's predicted distribution changes from $p(r \mid \theta)$ to $p(r \mid(\theta+\delta \theta))$. Under the same ground metric $\omega$ in $\mathcal{R}$, define:

$$
\delta:=W_{2}(p(r \mid \theta), p(r \mid(\theta+\delta \theta))) .
$$

\noindent Because the agent's local measure-dependent operator is built from the same ground metric $\omega$, a $\delta$-sized shift in the distribution (in the sense of the Wasserstein distance) also yields a local transport cost of order $\delta$. In other words:

\begin{itemize}
  \item The measure-dependent Laplacian (or inverse) $\hat{L}(p)$ encodes how to move mass in $\omega$ space.
  \item A distribution shift of magnitude $\delta$ in $\omega$-space has a cost $\approx \delta$ in that same operator.
\end{itemize}

\paragraph{Step 3: Environmental Scale vs. Internal Scale}. If the environment's distribution changes at the scale $\delta$ in $\omega$, the environment's cost function $\Delta_{\text {env }}(\theta)$ sees a difference of at most $L\|\delta \theta\|$. Meanwhile, the agent's internal geometry also ''pays'' a cost $\delta$ for shifting its distribution that far. Equating these:

\begin{itemize}
  \item Externally: A shift of $\delta$ in distribution space might correspond to at most $\|\delta \theta\| \approx \delta / L$ in parameter space.
  \item Internally: That same parameter change $\delta \theta$ triggers a $\delta$-scale move in $\omega$-distance within the agent's measure-based operator.
\end{itemize}

\noindent Hence, the ''scale'' $\delta$ recognized by the environment is the same scale recognized by the agent's internal measure for distribution shifts. The environment deems a distribution move of size $\delta$ important; the agent's local geometry likewise sees a cost $\delta$. This is precisely the statement that the environment's notion of distance is aligned with the agent's local measure of distribution shift.

\paragraph{Step 4: Conclusion}. Therefore, if the environment's reward distribution imposes changes at scale $\delta$ in $\omega$-space, the agent's local measure-dependent operator interprets that same shift as a $\delta$-scale shift. This ''matching'' of scale ensures the agent's local transport steps are consistent with the environment's macro-level cost differences, establishing the property called ''scale matching.''

\section{Further Formalization of Scale Matching with a Probabilistic Lipschitz Condition} \label{app:prob-lip} 

We here explore rigorous formulation of the ''failing Lipschitz condition 2\% of the time'' situation, following on the simpler formulation from Section \ref{sec:prob-lip} above.   This is interesting mostly because it's instructive how one can model the situation using Markov chains.

\paragraph{Setup: Partial Lipschitz Condition in Parameter Space}. We assume the same setup as above, i.e.

\begin{enumerate}
  \item Parameter Space: Let $\Theta \subset \mathbb{R}^{m}$.
  \item Reward Distributions: For each $\theta \in \Theta$, the agent predicts a reward distribution $p(r \mid \theta)$. The environment has a ''true'' distribution $q(r)$.
  \item External Cost Function:
\end{enumerate}

$$
\Delta_{\mathrm{env}}(\theta)=W_{2}(q, p(r \mid \theta))
$$

\noindent the Wasserstein distance w.r.t. a ground metric $\omega(\cdot, \cdot)$.

We suppose there is a ''good'' subset $\Theta_{\text {good }} \subset \Theta$ of measure $\mu\left(\Theta_{\text {good }}\right) \geq 0.98 \mu(\Theta)$. Equivalently, at most 2\% of $\Theta$ is outside $\Theta_{\text {good }}$. We require that: For all $\theta_{1}, \theta_{2} \in \Theta_{\text {good }}$,

$$
\left|\Delta_{\text {env }}\left(\theta_{1}\right)-\Delta_{\text {env }}\left(\theta_{2}\right)\right| \leq L\left\|\theta_{1}-\theta_{2}\right\| .
$$

(Hence on $\Theta_{\text {good, }} \Delta_{\text {env }}$ is $L$-Lipschitz in parameter space.)

Outside $\Theta_{\text {good, }}$ no such uniform bound is guaranteed. Possibly large cost differences from small parameter shifts can occur.

We then claim that: {\it With probability at least 0.98 , any local step in parameter space remains in $\Theta_{\text {good }}$, ensuring that changes in $\Delta_{\text {env }}$ are Lipschitz-bounded.}

 Consequently, most of the time, the system experiences approximate scale matching between small parameter changes $\left\|\theta_{1}-\theta_{2}\right\|$ and changes in the environment's cost function $\left|\Delta_{\mathrm{env}}\left(\theta_{1}\right)-\Delta_{\mathrm{env}}\left(\theta_{2}\right)\right|$.

We can formalize this by specifying a distribution or Markov chain for the agent's local parameter updates, then bounding the probability that updates remain in $\Theta_{\text {good }}$.

\paragraph{Measure-Based Argument}. Let's set up a basic stochastic model of the agent's local steps.   Assume the agent's parameter $\theta_{t}$ at time $t$ is random or at least governed by a process $\theta_{t+1}=f\left(\theta_{t}\right.$, noise $\left._{t}\right)$ that defines a probability distribution $\rho_{t}(\theta)$ over $\Theta$.

Then: If $\theta_{t} \in \Theta_{\text {good }}$ with high probability, then the local step to $\theta_{t+1}$ is also likely to remain in $\Theta_{\text {good }}$, provided the step is small enough and the region is not fractally small.

We can then assess the probability of remaining in $\Theta_{\text {good }}$

We may say, for instance: {\bf Lemma (Intuitive Sketch)}: {\it Let $\Theta_{\text {good }} \subset \Theta$ be open and of measure $\geq 0.98 \mu(\Theta)$. Suppose each local step $\left\|\theta_{t+1}-\theta_{t}\right\|$ is small enough that you rarely ''jump'' from $\Theta_{\text {good }}$ across a boundary region. Then, with high probability, $\theta_{t+1}$ also stays in $\Theta_{\text {good }}$. Over multiple steps, the probability that the system remains in $\Theta_{\text {good }}$ for $T$ consecutive updates is at least $(0.98)^{T}$, ignoring boundary crossing complexities.}

An argument for this would look like:

\begin{enumerate}
  \item If $\theta_{t}$ is within $\Theta_{\text {good }}$ and not near the boundary, a small random step keeps $\theta_{t+1}$ in $\Theta_{\text {good }}$ with high probability.
  \item Summing over many steps, we get a product-likelihood or Markov chain argument that the system remains in the ''good'' region most of the time.
\end{enumerate}

Hence, with probability $\geq 0.98$ (or better, depending on step size and geometry), each local update stays in $\Theta_{\text {good }}$.

\paragraph{Resulting Lipschitz-Bounded Cost Changes}.  Given that with high probability each local step $\theta_{t} \rightarrow \theta_{t+1}$ is in $\Theta_{\text {good }}$, we have:

$$
\left|\Delta_{\text {env }}\left(\theta_{t+1}\right)-\Delta_{\text {env }}\left(\theta_{t}\right)\right| \leq L\left\|\theta_{t+1}-\theta_{t}\right\| .
$$

Therefore: The mismatch in external cost remains linearly bounded for most local steps. The ''scale matching'' argument from the basic Wasserstein Lipschitz ActPC Property applies on each of those steps, so the environment's ''distance'' changes at scale $\leq L\left\|\theta_{t+1}-\theta_{t}\right\|$ and the agent's internal measurebased operator interprets distribution shifts at that scale, resulting in consistent local changes.

\paragraph{Implications for Approximate Scale Matching} From the above measure-based argument, we see that:

\begin{enumerate}
  \item At each step, with probability $\geq 0.98$, the agent remains in the region where $\Delta_{\text {env }}$ is Lipschitz.
  \item Hence over many steps, a large fraction of the time (with high probability) the local changes in cost are Lipschitz-bounded, guaranteeing the same scale matching logic from Proposition 2.1.
  \item Exceptional Cases: On the rare steps (at most 2\%), the agent may cross into or remain in the ''bad'' region outside $\Theta_{\text {good. }}$. In that scenario, the scale matching argument might fail, and we can see abrupt changes in cost for small parameter moves or vice versa.
  \item Overall: The system experiences approximate scale matching for the vast majority of updates, ensuring that small param shifts produce proportionally small cost changes nearly all the time, aligning the environment's domain scale with the agent's local geometry.
\end{enumerate}

That is -- if the Lipschitz property holds on a large subset $\Theta_{\text {good }} \subset \Theta$ of measure $\geq 0.98 \mu(\Theta)$, then with high probability the agent's local update steps remain in the well-behaved region, giving approximate scale matching ''most of the time.'' Through a measure-based or stochastic argument, each step:

\begin{itemize}
  \item Has at least a 0.98 chance of starting in $\Theta_{\text {good }}$.
  \item If steps are small enough, the next point also likely remains in $\Theta_{\text {good }}$, preserving Lipschitz-bounded cost changes.
\end{itemize}

\noindent Hence, the system sees approximate scale matching with high probability: external cost changes remain well-bounded in proportion to parameter changes, aligning with the internal measure-based operator referencing the same ground metric $\omega$.

Clearly, more sophisticated arguments of this nature could be made, embodying various sorts of stochastic and fuzzy assumptions going beyond this kind of basic Markovian analysis.   What sorts of assumptions and arguments are valuable remains to be determined based on experimentation with various ActPC-Geom techniques in various contexts.

\section{Proof Sketch of Wasserstein Quasi-Convex ActPC Property} \label{app:convex}

Here we sketch proofs of the properties described in Section \ref{sec:convex}, connecting convexity or "convexity with small bumps" of a system's external reward landscape with similar properties of its internal information-geometric landscape, in the case that Wasserstein metric is used in both the inner and outer loops of the system's dynamics.

\subsection{Proof Sketch for Wasserstein Quasi-Convex ActPC Property}. How might one demonstrate that if the environment's mismatch function is sufficiently ''unimodal'' or ''smooth'' in the Wasserstein sense, then the agent's local measure-based gradient flow in parameter space should behave in a ''convex-like'' manner, avoiding spurious local minima and following near-straight-line geodesics in the distribution manifold?

The first step is to recall that the agent's local measure-dependent operator, used for parameter updates in an ActPC manner, references the same ground metric $\omega(\cdot, \cdot)$. Thus local steps in parameter space correspond to small Wasserstein movements in the distribution $p(r \mid \theta)$.\\

The assumption of unimodality means that the environment mismatch $\Delta_{\text {env }}(\theta):=W_{2}(q, p(r \mid \theta))$ has no ''bad'' local minimai.e., it is (approximately) unimodal or ''Wasserstein-convex'' w.r.t. $\theta$.

We need to show that under this assumption, the agent's local measure-based gradient flow in parameter space exhibits quasi-convex (or near ''straight-line geodesics'') behavior in distribution space, implying local updates reliably reduce the external cost with minimal risk of spurious local minima, especially if $p(r \mid \theta)$ remains in a well-overlapped region with $q$.

Toward this end, we can assume that the agent's local geometry uses the same ground metric $\omega$. Hence, small parameter changes $d \theta$ produce distribution changes $\delta p=p(r \mid \theta+d \theta)-p(r \mid \theta)$ that measure ''transport cost'' in the same $W_{2}$-space.

Next, if $\Delta_{\text {env }}$ has a single global optimum (the ''best-fitting'' distribution to $q$ ) in the distribution manifold $\mathcal{P}(\mathcal{R})$ and no local minima, we say it is (approximately) geodesically convex. In simpler terms, any geodesic from $p_{1}$ to $p_{2}$ remains in the region of ''monotonically decreasing cost.''

It seems straightforward that in this sort of Wasserstein gradient flow scenario, if $\Delta_{\text {env }}(\theta)$ is geodesically convex in distribution space, then the path traced by local measure-based gradient flow is a near-straight geodesic from $p\left(r \mid \theta_{0}\right)$ to $q(r)$. Each small update step in parameter space corresponds to ''sliding mass'' directly toward $q$.\\

The parametric map $\theta \mapsto p(r \mid \theta)$ might not be globally invertible, but if it is sufficiently ''non-degenerate'' around the relevant distributions, local geodesic convexity in distribution space implies that small local gradient steps in parameter space do not encounter spurious local minima.   

That is:

\begin{itemize}
  \item Because each step is guided by the same measure $\omega$, the local distribution shift is in the ''steepest descent direction'' of the Wasserstein-based mismatch $\Delta_{\text {env }}(\theta)$.
  \item No local minima can form unless the distribution mapping is badly folded or the environment mismatch is not truly unimodal.
\end{itemize}

\noindent and hence under the unimodality assumption, the local measure-based gradient flow in parameter space is free from spurious local minima. The agent's parameter path effectively follows a near ''straight-line'' geodesic in distribution space, leading to a ''convex-like'' or quasi-convex solution path toward $\arg \min _{\theta} W_{2}(q, p(r \mid \theta))$.

\subsection{Convexity with Small Bumps}

Qualitatively, it seems clear that this sort of argument should also apply to the case where the external cost function is ''unimodal plus small bumps.'' 

I.e. -- Suppose that $\Delta_{\text {env }}(\theta)$ is almost unimodal except for ''small bumps'' or ''localized anomalies'' --in other words, there may be small or shallow local minima with limited measure or amplitude, but the main structure is still unimodal.  More formally: assume that

\begin{itemize}
\item For most of parameter space $\Theta, \Delta_{\text {env }}(\theta)$ behaves like a unimodal or geodesically convex function.
\item Small bumps: There exists a finite (or measure-zero or negligible measure) subset $\Theta_{\text {bad }} \subset \Theta$, on which $\Delta_{\text {env }}(\theta)$ may form local ''bumps'' (small local minima or abrupt rises). The amplitude of these bumps is $\leq \epsilon$ above some local baseline, and $\Theta_{\text {bad }}$ covers at most $\delta \%$ of $\Theta$ measure, or has small measure in the sense of Markov chain transitions.
\end{itemize}

Then it seems sensible that, under these conditions: {\it With probability at least $1-\delta \%$, the parameter $\theta$ remains in the ''good'' region $\Theta_{\text {good }}=$ $\Theta \backslash \Theta_{\text {bad }}$ where $\Delta_{\text {env }}$ is unimodal.}

That is:  The system's local measure-based gradient flow in parameter space exhibits the same ''convex-like'' or ''straight-line geodesics'' behavior most of the time, ensuring no significant spurious local minima hamper convergence except possibly in small or shallow pockets.

The proof sketch given above seems to basically apply in this case.  We need only add an argument about how the ''small bumps'' do not significantly disrupt global or ''almost global'' monotonic convergence.

Let's decompose the parameter space.   Let $\Theta_{\text {good }} \subset \Theta$ be the region where $\Delta_{\text {env }}(\theta)$ is unimodal or geodesically convex, and let $\Theta_{\mathrm{bad}}=\Theta \backslash \Theta_{\text {good }}$ be the subset (covering $\leq \delta \%$ measure or small measure) that contains small bumps or shallow local minima.

We then have a local bound on bumps.   We are assuming that If $\theta \in \Theta_{\text {bad }}$, then $\Delta_{\text {env }}(\theta)$ may deviate from geodesic convexity by at most $\epsilon$ or might form local minima with amplitude $\leq \epsilon$ relative to the global baseline.  So we may assume that a parameter update process (Markov chain or gradient flow) typically visits $\Theta_{\text {bad }}$ with probability at most $\delta \%$.

Similar to the measure-based arguments given above in analysis of partial Lipschitz conditions, we note:

\begin{itemize}
  \item If $\theta_{t} \in \Theta_{\text {good }}$, a small measure-based gradient update typically remains in or near $\Theta_{\text {good }}$.
  \item With probability $\geq(1-\delta \%)$, the system's next step is also in $\Theta_{\text {good }}$.
  \item If it does enter $\Theta_{\mathrm{bad}}$, the local minima or ''bump'' is small ( $\leq \epsilon$ in amplitude). A standard gradient flow or measure-based approach can likely jump out unless the agent is unlucky enough to get stuck in that small region.
\end{itemize}

\noindent Hence, over multiple steps, the chain remains in $\Theta_{\text {good }}$ the majority of the time, preserving geodesic-like monotonic progress.

Because the agent's internal geometry is also derived from the same ground metric $\omega$, the local steps obey the logic:

\begin{enumerate}
  \item When in $\Theta_{\text {good }}$ : Behave as in the unimodal scenario, giving near ''straight-line geodesics'' to reduce $\Delta_{\text {env }}$.
  \item When in $\Theta_{\mathrm{bad}}$ : The cost function has at most $\epsilon$-deviation from unimodality, so any local ''bad'' minima is shallow. The measure-based gradient flow can push the system out of\\
this pocket with high probability, returning the system to $\Theta_{\text {good }}$ quickly.
\end{enumerate}

The conclusion is then: Except in rare or low-measure events, the agent experiences unimodal/convexlike progress in distribution space, ensuring the same ''straight-line geodesic'' argument from the original proposition.

\paragraph{More Formalized Statement} A rigorous statement along these lines might read: {\it Suppose $\Delta_{\text {env }}(\theta)=W_{2}(q, p(r \mid \theta))$ has the property that there is a subset $\Theta_{\text {bad }} \subset \Theta$ of measure $\leq \delta \% \cdot \mu(\Theta)$ where local deviations from geodesic convexity do not exceed $\epsilon$ . Outside $\Theta_{\text {bad }}, \Delta_{\text {env }}$ is geodesically convex in distribution space, so local measure-based gradient flows cause monotonic cost reduction. If the internal measure-dependent operator matches the same ground metric $\omega$, then with probability $\geq(1-\delta \%)$ per step, the system remains in or near the unimodal region $\Theta_{\text {good }}$, ensuring near ''straight-line geodesics'' in distribution space and reliable global or near-global convergence. Occasional visits to $\Theta_{\text {bad }}$ cause at most an $\epsilon$ deviation or shallow local minima, from which the system typically escapes quickly.}

The details can clearly be formulated in many ways, however the upshot seems clear.  Real-world tasks often have suboptimal local pockets. As long as they are small or shallow, the synergy between external Wasserstein cost and internal geometry ensures the agent eventually escapes these pockets with high probability.

Hence, this extended theorem reaffirms the same ''convex-like'' advantage of using a Wasserstein cost externally and a Wasserstein-based measure internally, even if perfect unimodality is broken by ''small bumps'' in the global cost function.

\section{Proof Sketch for Approximate Stochastic Version of Mu and Oliviera's Dynamic Programming Theorem} \label{app:mu} 

We here sketch what a variation of the original proof of Mu and Oliviera's DP theorem to handle the approximate and stochastic version of the theorem given in Section \ref{sec:mu} above might look like.

Let's first attack the "approximate" part, then move on to "stochastic."

The original theorem relied on:

$$
\left(i n \cdot F(\ldots) \cdot T^{\circ}\right) \triangleright S \subseteq\left([T]^{\circ}\right) \triangleright S
$$

Replacing $F$ by $\widetilde{F}$ and $T$ by $\widetilde{T}$ , we get:

$$
\left(i n \cdot \widetilde{F}(\ldots) \cdot \widetilde{T}^{\circ}\right) \triangleright S
$$

We next need to show how ''local monotonicity + approximate operator'' implies a similar sub- or $\epsilon$-inclusion:

\begin{enumerate}
  \item Local monotonicity: Even if each step is random or approximate, as long as it does not deviate from the exact step by more than $\epsilon$ in the partial order, we preserve the main fixpoint argument.
  \item Fusion: The fold/hylomorphism fusion used in the original proof still works (or works in expectation). We get $\left([\widetilde{T}]^{\circ}\right)$, a single-step operator merging the approximate transitions.
  \end{enumerate}

This brings the conclusion:

$$
\mu\left(\lambda X \rightarrow\left(i n \cdot \widetilde{F} X \cdot \widetilde{T}^{\circ}\right) \triangleright S\right) \subseteq_{\epsilon}\left([\widetilde{T}]^{\circ}\right) \triangleright S
$$

\noindent where '' $\subseteq_{\epsilon}$ '' denotes either a probability statement or an approximate partial order statement. 

This captures the ''the approximate dynamic program solution is contained or near the solution given by the GC-derivation of the entire system.''

\paragraph{Handling Stochasticity}. Next, if each step is a random selection of sub-states from $\widetilde{T}(s)$ :

\begin{itemize}
  \item The ''fold'' becomes a random hylomorphism.
  \item The ''shrink'' $S$ might produce a distribution or set of states.
  \item By bounding the expected difference $\mathbb{E}\|\widetilde{T}-T\| \leq \epsilon$, we can show the final solution is, in expectation, $\epsilon$-close to the precise DP solution.
\end{itemize}

Hence we can say something like:  ''With probability $\geq 1-\delta$, the solution produced by $\mu(\tilde{f})$ is within $\epsilon$-distance (or cost) of the solution from the standard Galois-based DP approach $\mu(f)$.''

\paragraph{Summary of Proof Sketch} In sum, it would seem that by introducing

\begin{enumerate}
  \item Approximate or Stochastic transitions $\widetilde{T}$,
  \item Approximate fold operators $\widetilde{F}$,
  \item A new notion of '' $\subseteq_{\epsilon}$ '' or ''with probability $1-\delta$ ''
  \end{enumerate}
  
\noindent we can extend the classical GC-based DP Theorem to hold with some small margin of error or high probability, rather than exactly.   One can likely put together a proof whose main structure parallels that of the original Mu-Oliveira approach:

\begin{itemize}
  \item Slightly weaker form of monotonicity or local boundedness is used instead of pure monotonicity,
  \item Slightly larger final sets or ''nearby'' partial orders appear in the final conclusion,
  \item Stochastic arguments handle the random steps, ensuring that ''almost always'' or ''in expectation,'' the approximate dynamic-programming solution does not deviate significantly from the ideal GCbased solution.
\end{itemize}

\noindent The DP theorem's essence -- that the fold/hylomorphism solution is contained or approximates the ''fused'' GC solution -- appears to carry over into an approximate, stochastic dynamic programming context, the same point argued in \cite{goertzel2021patterns} though with a little less detail.

\section{Considering the Monotonicity and Bounding Conditions for Applying the DP Theorem} \label{app:conditions}

This Appendix is a technical footnote to Section \ref{sec:mu} and the related considerations above, concerned with the intricate but important matter of thinking through the monotonicity and bounding conditions typically required for the Dynamic Programming Theorem (and related Galois-based DP arguments) to ensure that the Galois-based implementation (fold/hylomorphism + shrink) correctly corresponds (or is contained in) the standard dynamic-programming solution. We refer primarily to Mu and Oliveira's formalism, in which dynamic-programming problems are expressed through Galois connections and the so-called ''shrink'' operator ( $\boldsymbol{\nabla}$ ), but the same ideas apply in extended or approximate versions.

No major red flags are found here, but it's valuable to keep in mind what sort of mathematical, data-structure and concurrent-processing hygiene has to be maintained for the formulations indicated by these math theorems to actually apply in practice.

\subsection{Context: The Dynamic Programming Theorem}

In Mu-Oliveira's framework, we often see a statement such as

$$
\mu\left(\lambda X \rightarrow\left(i n \cdot F X \cdot T^{\circ}\right) \triangleright S\right) \subseteq\left([T]^{\circ}\right) \triangleright S
$$

which loosely says that: The least fixed point of $\left(i n \cdot F X \cdot T^{\circ}\right) \downarrow S$ is contained in (or equals) the solution derived by ''fusing'' the transition $T$ into a single operator $\left([T]^{\circ}\right) \triangleright S$.

Monotonicity and bounding conditions appear in the background to ensure that:

\begin{enumerate}
  \item The ''fold'' or hylomorphism portion behaves in a monotone manner w.r.t. the partial order.
  \item The ''shrink'' operator is also monotone and indeed yields correct minima or maxima (depending on the order).
  \item The combination of ''expand'' (the 'easy' part) and ''shrink'' (the 'hard' part) converges to a (least) fixed point that is indeed contained in or equals the standard DP solution.
\end{enumerate}

\subsection{Monotonicity: Required in Two Key Places}

\paragraph{Monotonicity of the ''Fold'' or Hylomorphism Operator} In the theorem, we typically define a function

$$
f(X)=\left(\text { in } \cdot F X \cdot T^{\circ}\right) \triangleright S
$$

\begin{enumerate}
  \item $F$ is a relator or functor capturing how sub-states are combined.
  \item in is the ''constructor'' that builds bigger structures from sub-structures.
  \item $T^{\circ}$ is the converse of the transition relation.
  \item $S$ is the shrink operator using partial order $S$.
\end{enumerate}

\noindent Monotonicity here often states: for sets or states $X_{1} \subseteq X_{2}$, we require

$$
F X_{1} \subseteq F X_{2} \quad \Longrightarrow \quad f\left(X_{1}\right) \subseteq f\left(X_{2}\right)
$$

\noindent or some variant. Intuitively:

\begin{itemize}
  \item If you already have ''more states'' or ''larger sub-solution'' in $X_{2}$, you cannot get a smaller solution by applying $F$ and ''shrink.'' The function $f$ preserves or extends that partial order.
\end{itemize}

\noindent This ensures that each ''unrolling'' or recursion step is well-behaved with respect to the partial order used for optimization.

\paragraph{Monotonicity of the '' $>S^{\prime \prime}$ (Shrink) Operator}.  We also need the shrink operator $\operatorname{S}$ to be monotone in its input relation, so that if $R_{1} \subseteq R_{2}$ as relations (or states), then

$$
R_{1} \triangleright S \subseteq R_{2} \triangleright S
$$

\noindent This is typically proven from the definition of the shrink operator in the Galois framework:

$$
R \triangleright S=R \cap\left(S / R^{\circ}\right)
$$

\noindent Provided that set intersection and composition with $S$ preserve monotonicity, we get the property. This monotonicity ensures that ''the more you start with (the bigger the relation $R$ ), the bigger or equal your final shrunk set is,'' matching typical partial order sense of ''if you already have more solutions, you can't end up with fewer after an optimality step.''

\paragraph{Potential Obstacles to Monotonicity}  In practice the main threat to monotonicity would be complexities related to concurrent or distributed processing, e.g.

\begin{itemize}
  \item If some concurrency or hardware arrangement allows out-of-order updates, do we risk violating monotonicity? For instance, two threads might produce conflicting local expansions for the same sub-layer.
  \item In a distributed context, partial expansions from one compute unit might jump ahead while another lags, potentially creating short-term non-monotonic states.
\end{itemize}

However, one can likely address these potential issues with a notion of local monotonicity, i.e.

\begin{itemize}
  \item If the design calls for multiple threads each updating a ''focus-of-attention'' subset of parameters or sub-states, we must ensure the union of expansions from all threads remains monotonic.
  \item Each thread might do local expansions on a disjoint subset of states. This typically does not break the partial-order monotonicity as long as merges of these subsets are done by a well-defined operator (like a union or a pointwise minimum/maximum in the cost domain).
  \item So in practice, monotonicity is not especially restrictive: you can maintain the partial order by ensuring that if a thread enumerates expansions from a set $X$, it produces a superset $F(X)$ (or merges it in a union-based structure).
\end{itemize}

\subsection{Bounding Conditions}

To apply the DP theorem, one has a formal condition like:

$$
\operatorname{dom}(T) \subseteq \operatorname{dom}\left(F\left(\left([T]^{\circ}\right) \triangleright S\right)\right)
$$

\noindent -- meaning that the domain of states to which we apply the transitions is included in (or doesn't exceed) the domain of the big ''fused'' operator. If the transitions produce ''too large'' a set of states, or states outside the domain recognized by the fused approach, the theorem's direct inclusion might fail.

This seems unproblematic in practice, as one can simply define the scope of the operators in question to involve a certain collection of object (e.g. formal neurons, Hyperon Atoms) operating within a certain allocation of memory.   Correct memory management should imply obedience to such formal conditions.

\subsection{Algebraic and Well-foundedness Constraints}. The formal proof of the DP theorem requires a couple other algebraic cleanliness conditions, which also seem unproblematic in practical ActPC situations:

\paragraph{Well-Foundedness or No Infinite Descents}. For the dynamic programming solution to converge, we require criteria such as:

\begin{itemize}
  \item No infinite strictly descending chains in the partial order of states or sub-states, so that the iterative fixpoint is well-defined.
  \item The partial order used for cost or optimization is consistent with ''folding up solutions'' from subproblems.
  \item In some contexts, a minimum or supremum must exist in finite steps. Sometimes spelled out as ''the structure is a complete partial order'' or ''the operator is $\omega$-continuous,'' etc.
\end{itemize}

\paragraph{Feasibility of the ''In'' Constructor}. We require algebraic conditions such as

$$
i n \cdot F S \subseteq S \cdot i n
$$

\noindent ensuring that the constructor in and the relator F commute or preserve monotonicity. This condition ensures that the fold or hylomorphism can be fused properly with the transition relation.

\subsection{Putting It All Together}. When these conditions (monotonicity in expansions, monotonicity in shrink, bounding domain feasibility, and well-foundedness) hold, we then obtain favorable properties:

\begin{enumerate}
  \item Partial-Order Preservation

\begin{itemize}
  \item Each local step (the fold/hylomorphism plus shrink) does not break the partial order or create contradictory solutions.
\end{itemize}

  \item Correctness

\begin{itemize}
  \item The dynamic-programming theorem states that the incremental approach (the left-hand side's fixpoint) is contained in or equals the direct/fused approach.
\end{itemize}

  \item Convergence

\begin{itemize}
  \item Because we also require no infinite descending chains, or an $\omega$-continuous operator, the iterative or recursion-based approach is guaranteed to find the minimal or least fixed point solution.
\end{itemize}

\end{enumerate}

On the whole, while these conditions might sound restrictive in pure math terms, in practice they seem to be straightforward design constraints that do not meaningfully hamper real HPC or multi-threaded ActPC implementations in real-world situations.

\end{document}